\documentclass{article}

\RequirePackage{silence}
\PassOptionsToPackage{numbers,compress}{natbib}
\usepackage[preprint]{neurips_2026}

\usepackage[utf8]{inputenc}
\usepackage[T1]{fontenc}
\usepackage{hyperref}
\usepackage{url}
\usepackage{booktabs}
\usepackage{amsfonts}
\usepackage{amsmath,amssymb}
\usepackage{nicefrac}
\usepackage{microtype}
\usepackage{xcolor}
\usepackage{colortbl}
\usepackage{multicol}
\usepackage{xfp}
\definecolor{cellLo}{HTML}{EDEDED}
\definecolor{cellHi}{HTML}{7CBCF7}
\definecolor{cellOurs}{HTML}{F7B77C}
\definecolor{cellOracle}{HTML}{7CD99B}
\definecolor{linkColor}{HTML}{2B7CD3}
\definecolor{phRed}{HTML}{D7261D}   
\definecolor{phGreen}{HTML}{0F9D0F} 
\definecolor{phBlue}{HTML}{2353D8}  
\definecolor{phPink}{HTML}{E0218A}  
\newcommand{\pht}[2]{\textcolor{#1}{$\boldsymbol{t{=}#2}$}}
\newcommand{\vc}[1]{%
  \ifdim #1pt>0.5pt%
    \cellcolor{cellHi!\fpeval{round(100*#1*#1*#1*#1)}!white}%
  \else%
    \cellcolor{cellHi!\fpeval{round(60*#1)}!white}%
  \fi
  #1}
\newcommand{\vcx}[2]{%
  \ifdim #2pt>0.5pt%
    \cellcolor{#1!\fpeval{round(max(12,100*#2*#2*#2*#2))}!white}%
  \else%
    \cellcolor{#1!\fpeval{round(max(12,60*#2))}!white}%
  \fi
  #2}
\newcommand{\vco}[1]{\vcx{cellOurs}{#1}}   
\newcommand{\vcg}[1]{\vcx{cellOracle}{#1}} 

\hypersetup{
    colorlinks=true,
    linkcolor=linkColor,
    citecolor=linkColor,
    urlcolor=linkColor
}
\usepackage{graphicx}
\usepackage{algorithm}
\usepackage{algorithmicx}
\usepackage{algpseudocode}
\usepackage{multirow}
\usepackage{xspace}
\usepackage{array}
\usepackage{tabularx}
\usepackage{adjustbox}
\usepackage{enumitem}
\usepackage{caption}
\usepackage{wrapfig}

\newcommand{\muvla}{$\mu$VLA\xspace}
\newcommand{\openvlaoft}{OpenVLA-OFT\xspace}
\newcommand{\openvla}{OpenVLA\xspace}
\newcommand{\mikasa}{MIKASA-Robo\xspace}

\newboolean{showcomments}
\setboolean{showcomments}{true}

\ifthenelse{\boolean{showcomments}}
  {\newcommand{\nb}[3]{
  {\color{#2}\small\fbox{\bfseries\sffamily\scriptsize#1}}
  {\color{#2}\sffamily\small$\triangleright~$\textit{\small #3}$~\triangleleft$}
  }
  }
  {\newcommand{\nb}[3]{}
  }

\makeatletter
\renewcommand{\@bottomtitlebar}{%
  \vskip 0.29in
  \vskip -\parskip
  \hrule height 1\p@
  \vskip -0.2in%
}
\makeatother

\title{\muvla: On Recurrent Memory for Partially Observable Manipulation in VLA Models}

\author{
  \textbf{Egor Cherepanov}$^{1,2}$ \quad
  \textbf{Nikita Kachaev}$^{1}$ \quad
  \textbf{Daniil Zelezetsky}$^{2}$ \\[1pt]
  \textbf{Aydar Bulatov}$^{1,2}$ \quad
  \textbf{Artem Pshenitsyn}$^{1,2}$ \quad
  \textbf{Yuri Kuratov}$^{1,2}$ \\[1pt]
  \textbf{Alexey Skrynnik}$^{1,2}$ \quad
  \textbf{Aleksandr I. Panov}$^{1,2}$ \quad
  \textbf{Alexey K. Kovalev}$^{1,2}$ \\[2pt]
  $^1$CogAI Lab, Moscow, Russia \qquad
  $^2$MIRAI, Moscow, Russia \\[2pt]
  \href{https://avanturist322.github.io/mu-vla/}{\textbf{\textcolor[HTML]{3497f3}{avanturist322.github.io/mu-vla}}}
}

\begin{document}

\maketitle

\textbf{}
\vspace{-4.5em}
\begin{center}
    \includegraphics[width=0.85\linewidth]{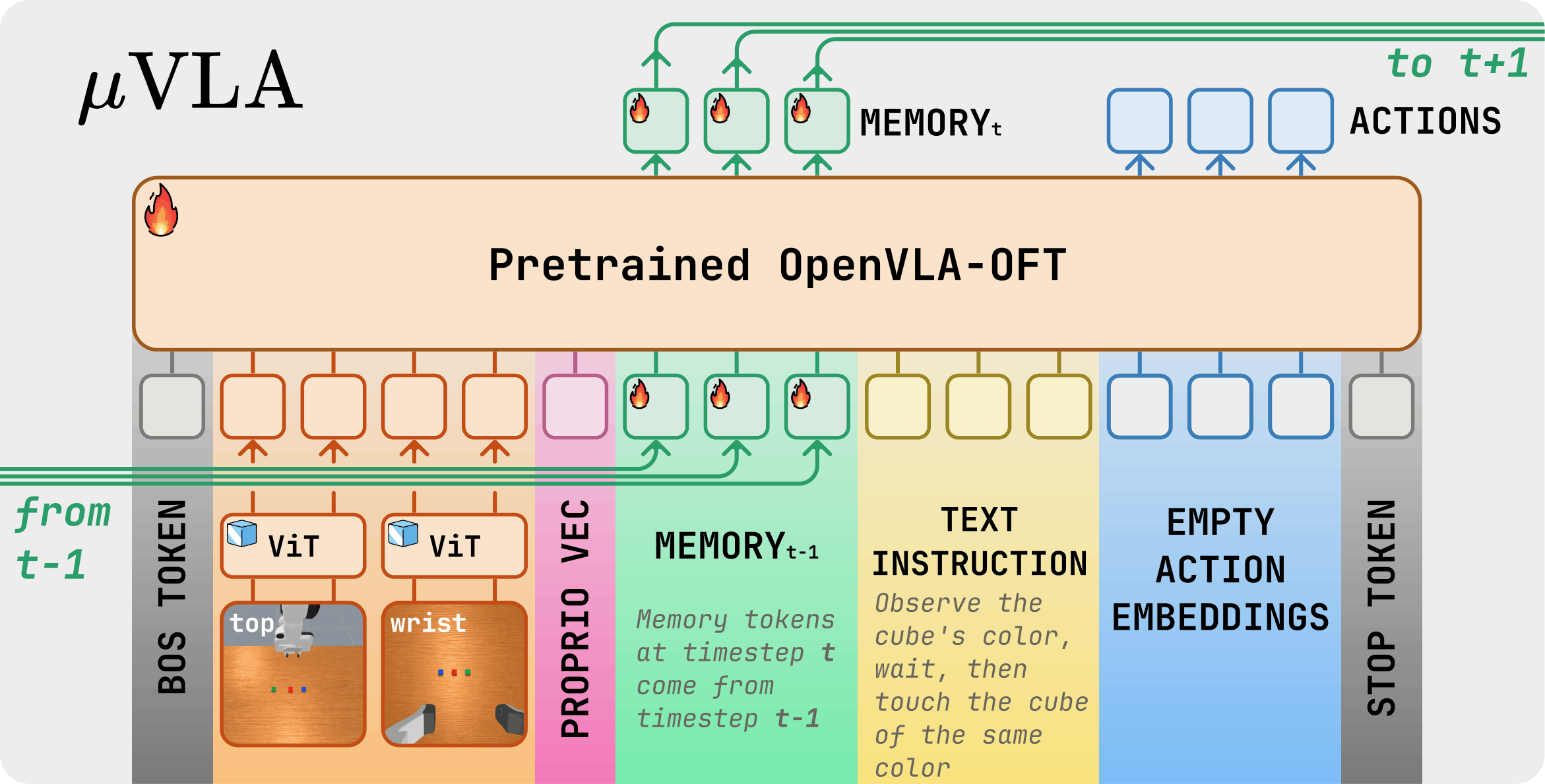}
    \vspace{-0.5em}
    \captionof{figure}{
        \textbf{\muvla: recurrent memory inside a pretrained VLA backbone.} At each timestep $t$, the model consumes visual observations, proprioception, language, and memory tokens from the previous step, then predicts actions and updated memory tokens passed recurrently to $t+1$.
    }
    \label{fig:visual_abstract}
    \vspace{-0.5em}
\end{center}

\begin{abstract}
\vspace{-1.2em}
Vision-language-action (VLA) models predict chunks of future actions from the current observation, an assumption that fails under partial observability, where decisions depend on information no longer visible. Existing memory-augmented VLAs simultaneously introduce recurrence, retrieval, compression modules, auxiliary objectives, hierarchical memory, or task-specific architectural changes, so the contribution of recurrence itself remains entangled with surrounding machinery. We present a controlled isolation study of recurrence in a strong pretrained VLA backbone. Our formulation augments the transformer with a small set of learnable memory tokens carried across timesteps and updated through self-attention, trained end to end with truncated backpropagation through time, with no auxiliary losses and no architectural changes. We instantiate this as $\mu$VLA, a family of OpenVLA-OFT variants parameterized by memory width $m$, TBPTT length $K$, and the memory update rule (cross-step gradients or a detached EMA), so that recurrence is the only varying factor. On MIKASA-Robo, $\mu$VLA improves average success rate on five training tasks from $0.42$ to $0.84$ at the strongest setting and reaches $0.23$ on held-out tasks with the same memory structure versus $0.07$ for the memoryless baseline. On tasks requiring different memory structure, performance remains near baseline. On LIBERO, the strongest recurrent variant achieves $96.2\%$ average success, indicating no regression under full observability. We interpret these results as a calibration of the capability envelope of minimal in-backbone recurrence, identifying the regime in which it is sufficient and the regime where additional memory structure is required. Demos and videos can be found in \textbf{\url{https://avanturist322.github.io/mu-vla/}}.
\end{abstract}

\begin{figure}[t]
  \vspace{-0.5em}
  \centering
  \includegraphics[width=1\linewidth]{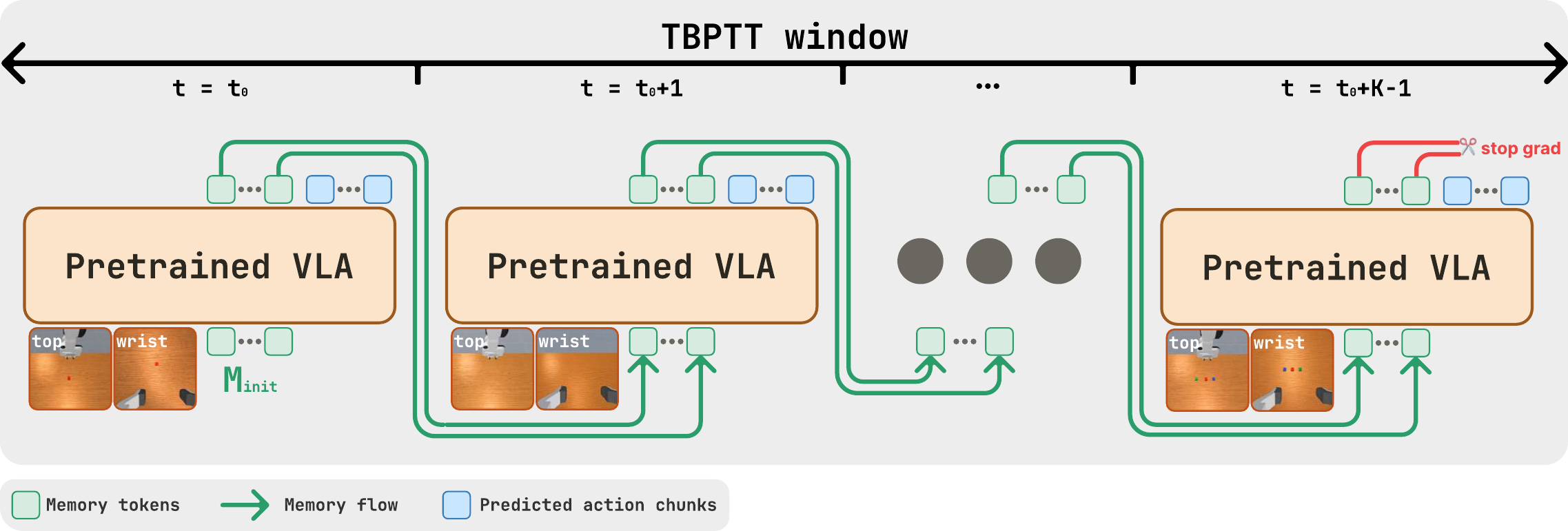}
  \caption{\textbf{TBPTT training of \muvla.} The pretrained VLA is unrolled over a window of $K$ steps. Memory tokens are initialized from $\boldsymbol{M}^{\mathrm{init}}$ at $t_0$ and recurrently passed between successive steps. Gradients flow through the unrolled pathway and are detached at the window boundary.}
  \label{fig:overview}
  \vspace{-0.5em}
\end{figure}
\vspace{-0.5em}
\section{Introduction}
\vspace{-0.5em}

Vision-language-action (VLA) models cast manipulation as next-token prediction over multimodal sequences~\citep{rt2, kim2024openvla, openvlaoft, pi0, pi05, octo, gr00t, smolvla, xvla}, predicting a chunk of actions from the current observation. This factorization scales but assumes the current observation is a sufficient statistic for control. Under partial observability, where decision-relevant information is transient, occluded, or set earlier in the episode, a Markovian policy cannot recover it. Existing VLAs incorporate history through finite context windows~\citep{contextvla, longvla, li2025cronusvla}, KV-cache reuse~\citep{tempofit, kvefficient}, external memory and retrieval~\citep{hamlet, memer, noteself, mapvla}, or recurrence~\citep{rememvla, avavla, recdepthvla, vpwem}, but typically combine the chosen mechanism with auxiliary losses, hierarchical state, or task-specific architectural changes, so the contribution of recurrence itself remains entangled with surrounding machinery.

We treat recurrence as a controlled variable inside a strong pretrained VLA. Our formulation augments the backbone token sequence with a small bank of learnable memory tokens carried across timesteps and updated through the standard self-attention forward pass, trained end to end with truncated backpropagation through time (TBPTT) on temporally ordered episodes, with no auxiliary losses and no architectural additions. We instantiate this as \muvla, a family of OpenVLA-OFT~\citep{openvlaoft} variants parameterized by memory width $m$, TBPTT length $K$, and the memory update rule (cross-step gradients or a detached EMA), so that family members differ only in their recurrent state. \muvla\ should be interpreted as a controlled intervention study rather than a new foundation-model architecture: we calibrate what minimal in-backbone recurrence can do, since gains reported by memory-augmented VLAs that simultaneously introduce recurrence, retrieval, hierarchical state, and auxiliary objectives cannot otherwise be attributed to recurrence per se. We therefore primarily compare within the \muvla\ family on \mikasa\footnote{Throughout we use MIKASA-Robo-VLA, a version of the MIKASA-Robo benchmark adapted specifically for training and evaluating VLA models: \url{https://mikasarobo.github.io/}.}~\citep{mikasarobo} and use LIBERO~\citep{liu2023libero} as a fully observable control, including representative memory-augmented VLAs (CronusVLA~\citep{li2025cronusvla}, MemoryVLA~\citep{memoryvla}) only to contextualize the resulting performance regime.

\textbf{Contributions.} (i) We present, to the best of our knowledge, the first controlled isolation study of recurrence in a pretrained VLA backbone, varying recurrence as a single experimental axis with no retrieval, compression, hierarchical memory, or auxiliary losses entangled with the recurrent state. (ii) We introduce \muvla (Fig.~\ref{fig:visual_abstract}), a parameterized family of recurrent fine-tunes of \openvlaoft\ over $(m, K, \text{write rule})$ sharing dataloader, backbone, optimizer, and inference. (iii) We calibrate the capability envelope of minimal in-backbone recurrence on \mikasa\ and LIBERO, identifying three regimes (in-distribution partial observability $0.42\!\to\!0.84$, held-out tasks with the same memory structure $0.07\!\to\!0.23$, and tasks with new memory structure where recurrence stays near baseline), and probe the trained state with representation dynamics, attention rollouts, causal noise and freeze-first interventions, chunked-inference and phase-length sweeps, and OOD cue-identity tests.

\vspace{-0.5em}
\section{Related Work}
\vspace{-0.5em}
\paragraph{Vision-language-action models.}
VLA models formulate robot control as sequence modeling over multimodal tokens~\citep{rt2, kim2024openvla, openvlaoft, pi0, pi05, octo, gr00t, smolvla, xvla}. The model consumes the current observation, language instruction, and proprioception, and predicts future actions. Variants differ in action parameterization (discrete tokens or continuous actions), pretraining scale, and embodiment handling, but share a common assumption: the current observation, possibly augmented with a short context window, is sufficient for control. This assumption is effective in fully observable settings, but fails under partial observability, where relevant information is absent from the current input and cannot be recovered by a Markovian policy.

\begin{wrapfigure}{r}{0.6\textwidth}
\vspace{-2.5em}
\centering
\includegraphics[width=\linewidth]{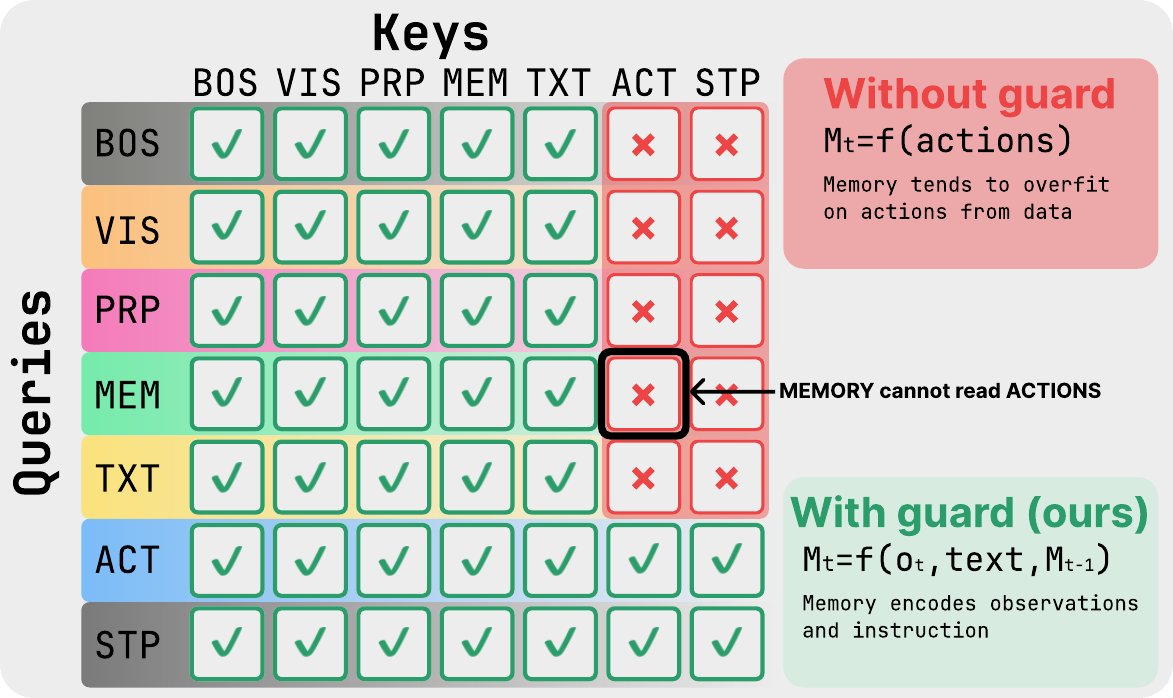}
\vspace{-1.5em}
\caption{\textbf{Attention mask with the memory-action guard.}
Memory tokens attend only to observations, proprioception, language,
and previous memory state, but cannot read action tokens. This prevents
the recurrent state from trivially copying demonstrated actions and
encourages encoding of task-relevant observations instead.}
\label{fig:attention}
\vspace{-1.0em}
\end{wrapfigure}

\paragraph{Incorporating history in VLA models.}
Prior work incorporates past information in three main ways. A first line extends the input with finite history, for example by concatenating multiple frames or reusing cached activations and KV states~\citep{contextvla, longvla, li2025cronusvla, tempofit, kvefficient}. This increases context but remains bounded and does not provide a mechanism for deciding what should persist. A second line introduces external memory and retrieval systems~\citep{hamlet, memer, mapvla, noteself, torne2026mem}, where past observations are stored and selectively accessed during inference. These approaches add flexibility but also additional components and a separate storage policy. A third line returns to recurrence, propagating latent state across timesteps~\citep{rememvla, avavla, vpwem, memoryvla}. These methods allow the model to learn what to retain, but typically combine recurrence with auxiliary objectives, dedicated memory modules, or architectural modifications.

\paragraph{Position.}
Across these directions, recurrence is rarely evaluated in isolation: reported gains typically arise from systems that simultaneously introduce retrieval, hierarchical memory, auxiliary objectives, compression modules, or task-specific architectural changes, making it difficult to isolate the contribution of recurrence itself. In this work, we therefore study recurrence as a controlled design variable inside a fixed pretrained VLA backbone. We consider a minimal recurrent formulation consisting of a small set of learnable memory tokens propagated across environment steps through the backbone self-attention, trained end-to-end with TBPTT, without auxiliary supervision or additional modules. The design is closely related to recurrent memory transformers~\citep{rmt,ratm,cherepanov2026elmur}, but adapted to multimodal VLA control with an attention mask that prevents trivial action copying and an inference protocol aligned with training-time memory updates. Our goal is not to propose a new foundation-model architecture, but to characterize the capabilities and limits of minimal in-backbone recurrence under partial observability. We evaluate on \mikasa~\citep{mikasarobo}, which isolates memory-intensive manipulation by controlling the underlying dependency structure of each task, and on LIBERO~\citep{liu2023libero} as a fully observable control suite. This setup allows us to attribute changes in behavior specifically to the recurrent state and identify the regime in which minimal recurrence is sufficient, as well as where additional memory structure becomes necessary.

\vspace{-0.5em}
\section{Preliminaries}
\label{sec:prelim}
\vspace{-0.5em}

\paragraph{Imitation training of VLAs.}
We model manipulation as a POMDP $\langle \mathcal{S}, \mathcal{A}, \mathcal{O}, T, \Omega,
\mu_0 \rangle$. In the standard control loop, at each step the agent observes
$o_t \in \mathcal{O}$ and selects an action $a_t \in \mathcal{A}$ conditioned
on an instruction $\ell$. Since the state is only partially observed, the optimal
policy depends on history, $\pi^\star(a_t \mid o_{1:t}, \ell)$, and cannot in
general be recovered by a memoryless policy $\pi(a_t \mid o_t, \ell)$.

\vspace{-0.5em}
\paragraph{Standard VLA training.}
VLA policies typically depart from the one-step control loop by predicting an
open-loop chunk of $H$ future actions from a single model query,
$\pi_\theta(\boldsymbol{a}_{t:t+H-1} \mid o_t, \ell)$, and executing the entire
chunk before querying the model again. Training follows the same factorization:
a timestep $t$ is sampled from a demonstration episode
$(o_{1:T}, a_{1:T}, \ell) \sim \mathcal{D}$, and the model is optimized to
match the corresponding future action chunk,
\begin{equation}
\mathcal{L}_{\mathrm{VLA}}(\theta) =
\mathbb{E}_{(o_{1:T}, a_{1:T}, \ell) \sim \mathcal{D},\; t \sim \mathcal{U}(1,T)}
\left[
\sum_{h=0}^{H-1} w_{t,h}
\left\|
\hat a_{t+h}^{(t)} - a_{t+h}
\right\|_1
\right],
\label{eq:vla_chunk_loss}
\end{equation}
where $(\hat a_t^{(t)}, \ldots, \hat a_{t+H-1}^{(t)}) =
\pi_\theta(o_t,\ell)$ and $w_{t,h}$ masks actions beyond the episode
boundary. This training regime treats sampled timesteps as independent
examples and does not maintain any persistent state across steps.

\vspace{-0.5em}
\paragraph{Recurrent VLA training.}
For recurrent VLAs, the i.i.d.\ training setup is no longer sufficient,
since the policy depends on a latent state that evolves over time.
Training is therefore performed on temporally ordered episodes, with a
recurrent state $\boldsymbol{m}_t$ propagated across steps. At each
timestep, the model predicts an $H$-step action chunk conditioned on
the current observation and previous memory,$(\hat a_t^{(t)}, \ldots, \hat a_{t+H-1}^{(t)}) =
\pi_\theta(o_t, \boldsymbol{m}_{t-1}, \ell)$, 
and updates its state as $\boldsymbol{m}_t = f_\theta(o_t, \boldsymbol{m}_{t-1}, \ell)$.

The supervision signal matches the standard VLA objective, but is now
applied along full trajectories rather than independent timesteps:
\begin{equation}
\mathcal{L}_{\mathrm{rec}}(\theta) =
\mathbb{E}_{(o_{1:T}, a_{1:T}, \ell) \sim \mathcal{D}}
\left[
\sum_{t=1}^{T} \sum_{h=0}^{H-1} w_{t,h}
\left\|
\hat a_{t+h}^{(t)} - a_{t+h}
\right\|_1
\right].
\label{eq:rec_chunk_loss}
\end{equation}

\vspace{-0.5em}
The recurrent state $\boldsymbol{m}_t$ is not provided by the
dataset and is instead rolled forward by the model while consuming the
ground-truth observation sequence. Training thus follows a
teacher-forcing-like regime: observations come from the dataset, but the
latent state is fully model-generated. This forces the model to learn how
to write information into memory that remains useful for future
predictions.

At inference time, we use receding-horizon control. At every step, the
model predicts an action chunk
$\pi_\theta(\boldsymbol{a}_{t:t+H-1} \mid o_t, \boldsymbol{m}_{t-1}, \ell)$,
but executes only the first action $\hat a_t^{(t)}$, then re-queries the
model at the next step. This keeps the memory-update cadence aligned
with the environment step and is critical in dynamic settings, where
decision-relevant events may occur and disappear within a single chunk.

\vspace{-0.5em}
\paragraph{The OpenVLA-OFT backbone.}
We build on \openvlaoft~\citep{openvlaoft}, a fine-tuned variant of OpenVLA~\citep{kim2024openvla} that pairs a dual SigLIP/DINOv2~\citep{siglip, dinov2} encoder with a Llama-2-7B backbone~\citep{llama2}, fuses vision, language, and proprioception into one bidirectionally-attended sequence, and predicts a chunk of $H{=}8$ continuous actions with an L1 regression head. 

\vspace{-0.5em}
\paragraph{The MIKASA-Robo benchmark.}
\mikasa~\citep{mikasarobo} is a large tabletop manipulation benchmark designed to require memory under partial observability. Tasks cover three regimes: cue-recall (a short-lived visual signal at the start of the episode defines the goal, e.g., \texttt{RememberColor}), occlusion (object identity tracked through a partially observable shuffle, e.g., \texttt{ShellGame}), and sequential or predictive memory (\texttt{TakeItBack}, \texttt{Intercept}). Because the latent dependency structure is documented at the benchmark level, \mikasa\ isolates memory dependence as the primary varying factor and admits causal interventions on the recurrent channel.

\vspace{-0.5em}
\section{$\mu$VLA: A Controlled Family of Recurrent VLAs}
\label{sec:method}
\vspace{-0.5em}

We design \muvla\ (Fig.~\ref{fig:visual_abstract}) as a controlled recurrent model in which a single bank of $m$ learnable memory tokens is carried across timesteps inside the backbone self-attention and trained end to end against the action loss alone, with no auxiliary signals and no architectural additions. The model is closely related to recurrent memory transformers~\citep{rmt, ratm}, but applied at the level of environment steps in a multimodal VLA backbone. Three pitfalls must be addressed for the design to work in practice: a degenerate self-referential write through the action region, a step-shuffling dataloader that destroys temporal order, and a train-test mismatch between per-step memory updates and open-loop action chunking. We close them with a single change to the attention mask, a round-robin episodic dataloader, and receding-horizon inference. A schematic and full mask, dataloader, and TBPTT details are in Appendix~\ref{app:method} and Figure~\ref{fig:overview}.

\vspace{-0.5em}
\paragraph{Memory tokens and recurrence.}
\muvla\ inserts $m$ learnable memory tokens between \texttt{PROPRIO} and \texttt{TEXT} in the \openvlaoft\ input sequence:
$[\texttt{BOS}]\,[\text{VIS}]\,[\texttt{PROPRIO}]\,[\boldsymbol{M}_t]\,[\text{TEXT}]\,[\text{ACT}]\,[\texttt{STOP}]$,
with $\boldsymbol{M}_t\in\mathbb{R}^{m\times d}$ and $d$ the backbone hidden dimension. At the first step of each episode $\boldsymbol{M}_0\triangleq\boldsymbol{M}^{\mathrm{init}}$ is a shared learnable parameter; for $t\!\geq\!1$ memory is read from the previous step's hidden states at the memory positions: $\boldsymbol{M}_t \;=\; h_\theta(o_t,\ell;\,\boldsymbol{M}_{t-1})\bigl[\text{mem positions}\bigr]$ (Figure~\ref{fig:overview}),
so a single forward pass simultaneously reads memory and writes the recurrent state for the next step. The memory width $m$ is a hyperparameter; placement and positional-embedding details are in Appendix~\ref{app:method:placement}.

\vspace{-0.5em}
\paragraph{Attention-mask guard.}
\openvlaoft\ uses bidirectional self-attention over the entire input context, including the action region. Memory update therefore admits a degenerate self-referential solution $\boldsymbol{M}_t^{(i)}=\phi(\text{ACTION}_t^{(i)})$ that copies the predicted action chunk into memory. We zero the context-to-action block of the mask so that memory tokens (and the rest of the prefix) never attend to the action region, while the action region itself continues to read the full context (Figure~\ref{fig:attention}). The full mask and an information-theoretic argument are in Appendix~\ref{app:method:mask}.

\vspace{-0.5em}
\paragraph{Round-robin episodic dataloader.}
The standard \openvlaoft\ pipeline shuffles individual $(o_t, a_{t:t+H})$ pairs across episodes, which destroys the temporal order a recurrent state needs. We replace it with a round-robin dataset that maintains $B$ independent streams (one per batch slot); each stream walks a single episode step by step and then samples a new one, with a per-slot \texttt{is\_first} flag used by the recurrent loop to reset only those streams that started a new episode via $\boldsymbol{M}^{\mathrm{init}}$. Multi-environment mixtures and DDP seeding are in Appendix~\ref{app:method:dataloader}.

\vspace{-0.5em}
\paragraph{TBPTT and EMA write rules.}
We initialize \muvla\ from the released \openvla~\citep{kim2024openvla} checkpoint, pretrained on Open~X-Embodiment~\citep{openx}, and fine-tune end to end with LoRA~\citep{lora} adapters of rank $32$ on the backbone and action head, together with the memory-position embeddings and $\boldsymbol{M}^{\mathrm{init}}$. We compare two recurrent write rules. \emph{TBPTT} accumulates the L1 chunk loss over $K$ consecutive steps without detaching the memory chain, then takes one backward pass through the $K$-step recurrent graph; memory is detached only at the truncation boundary. \emph{EMA} replaces the end-to-end write with a detached low-pass filter,
\begin{equation}
  \boldsymbol{M}_{t+1} \;=\; \alpha\,\boldsymbol{M}'_t.\textsc{detach}() \;+\; (1-\alpha)\,\boldsymbol{M}_t.\textsc{detach}(),
  \label{eq:ema}
\end{equation}
where $\boldsymbol{M}'_t$ is the read-out at the memory positions and both operands are detached, so backward is local to a single step. The two variants share the same backbone, mask, dataloader, memory width, and inference protocol, and differ only in whether the recurrent write is learned end to end (TBPTT, length $K$) or applied as a detached exponential average (EMA, factor $\alpha$). Algorithm and hyperparameters are in Appendices~\ref{app:method:tbptt} and~\ref{app:hparams}.

\begin{table*}[t]
\caption{
  \textbf{Per-environment SR on all 23 \mikasa-VLA environments}
  (100 deterministic episodes; mean reported).
  Rows are grouped by the memory-semantic split of RQ3.
  Columns include the $\pi_{0.5}$ memoryless baseline; three
  OpenVLA-OFT reference points: (R1) the original dataloader recipe,
  (R2) the episodic dataloader without recurrence, and
  (R3) the episodic dataloader with access to the first frame.
  The \textbf{(+1st)} column appends the first observation of the episode
  to the model context at every timestep, and therefore acts as an oracle-style upper
  bound for tasks whose latent cue is fully visible in the initial observation.
  Remaining columns are \muvla variants at $m{=}64$
  varying the recurrent write ($K{=}1,2,8$, EMA, EMA without the
  action-copy guard), plus an $m{=}1$ bandwidth probe and a
  single-task \texttt{RememberColor5} control. $\dagger$ - episodic dataloader.
  Our best memory configuration ($m{=}64$, $K{=}2$;
  \colorbox{cellOurs!25}{orange}) and the oracle column
  (\colorbox{cellOracle!25}{green}) are highlighted.
}
\label{tab:main}
\centering
\footnotesize
\setlength{\tabcolsep}{3pt}
\resizebox{\textwidth}{!}{%
\begin{tabular}{lcccccccccc|c}
\toprule
\textbf{Environment}
  & \textbf{$\pi_{0.5}$}
  & \textbf{OpenVLA}
  & \textbf{OpenVLA}
  & \multicolumn{7}{c}{\textbf{\muvla family (Ours)}}
  & \cellcolor{cellOracle!25}\textbf{OpenVLA} \\
\cmidrule(lr){5-11}
  & 
  & \textbf{-OFT}
  & \textbf{-OFT}$^\dagger$
  & $\boldsymbol{m{=}1}$
  & $\boldsymbol{m{=}64}$
  & \cellcolor{cellOurs!25}$\boldsymbol{m{=}64}$
  & $\boldsymbol{m{=}64}$
  & $\boldsymbol{m{=}64}$
  & $\boldsymbol{m{=}64}$
  & $\boldsymbol{m{=}64}$
  & \cellcolor{cellOracle!25}\textbf{-OFT} \\
  & & & & $\boldsymbol{K{=}8}$ & $\boldsymbol{K{=}8}$ & \cellcolor{cellOurs!25}$\boldsymbol{K{=}2}$ & $\boldsymbol{K{=}1}$ & \textbf{EMA} & \textbf{EMA} & $\boldsymbol{K{=}8}$ & \cellcolor{cellOracle!25}\footnotesize \textbf{(+1st obs.)} \\
  & & & & & & \cellcolor{cellOurs!25}\footnotesize \textbf{best} & & & \footnotesize \textbf{full mask} & \footnotesize \textbf{single task} & \cellcolor{cellOracle!25}\footnotesize \textbf{oracle}\\
\midrule
\multicolumn{12}{l}{\emph{Training tasks (in-distribution)}} \\
\midrule
\texttt{ShellGamePush}                & \vc{0.86} & \vc{0.83} & \vc{0.90} & \vc{0.91} & \vc{0.83} & \vco{0.95} & \vc{0.93} & \vc{0.77} & \vc{0.94} & \vc{0.13} & \vcg{0.99} \\
\texttt{InterceptMedium}              & \vc{0.40} & \vc{0.36} & \vc{0.39} & \vc{0.49} & \vc{0.55} & \vco{0.47} & \vc{0.44} & \vc{0.55} & \vc{0.56} & \vc{0.01} & \vcg{0.45} \\
\texttt{TakeItBack}                   & \vc{0.85} & \vc{0.83} & \vc{0.87} & \vc{0.97} & \vc{0.98} & \vco{0.99} & \vc{0.99} & \vc{0.99} & \vc{0.99} & \vc{0.00} & \vcg{0.94} \\
\texttt{RememberColor5}               & \vc{0.12} & \vc{0.04} & \vc{0.09} & \vc{0.24} & \vc{0.35} & \vco{0.93} & \vc{0.40} & \vc{0.44} & \vc{0.25} & \vc{0.16} & \vcg{0.96} \\
\texttt{RememberShapeAndColor3x3}     & \vc{0.10} & \vc{0.03} & \vc{0.13} & \vc{0.08} & \vc{0.12} & \vco{0.86} & \vc{0.09} & \vc{0.09} & \vc{0.10} & \vc{0.05} & \vcg{0.91} \\
\textbf{Average (5 envs)}             & \vc{0.46} & \vc{0.42} & \vc{0.48} & \vc{0.54} & \vc{0.57} & \vco{0.84} & \vc{0.57} & \vc{0.57} & \vc{0.57} & \vc{0.07} & \vcg{0.85} \\
\midrule
\multicolumn{12}{l}{\emph{Held-out, matched memory semantics}} \\
\midrule
\texttt{ShellGameTouch}               & \vc{0.00} & \vc{0.00} & \vc{0.00} & \vc{0.00} & \vc{0.00} & \vco{0.00} & \vc{0.00} & \vc{0.00} & \vc{0.00} & \vc{0.07} & \vcg{0.00} \\
\texttt{ShellGamePick}                & \vc{0.00} & \vc{0.00} & \vc{0.00} & \vc{0.00} & \vc{0.00} & \vco{0.00} & \vc{0.01} & \vc{0.00} & \vc{0.01} & \vc{0.00} & \vcg{0.01} \\
\texttt{InterceptSlow}                & \vc{0.05} & \vc{0.05} & \vc{0.04} & \vc{0.05} & \vc{0.06} & \vco{0.07} & \vc{0.06} & \vc{0.05} & \vc{0.06} & \vc{0.02} & \vcg{0.04} \\
\texttt{InterceptFast}                & \vc{0.10} & \vc{0.00} & \vc{0.33} & \vc{0.21} & \vc{0.28} & \vco{0.27} & \vc{0.19} & \vc{0.28} & \vc{0.35} & \vc{0.00} & \vcg{0.31} \\
\texttt{InterceptGrabSlow}            & \vc{0.00} & \vc{0.00} & \vc{0.00} & \vc{0.00} & \vc{0.00} & \vco{0.00} & \vc{0.00} & \vc{0.00} & \vc{0.00} & \vc{0.00} & \vcg{0.00} \\
\texttt{InterceptGrabMedium}          & \vc{0.00} & \vc{0.00} & \vc{0.00} & \vc{0.00} & \vc{0.00} & \vco{0.00} & \vc{0.00} & \vc{0.00} & \vc{0.00} & \vc{0.00} & \vcg{0.00} \\
\texttt{InterceptGrabFast}            & \vc{0.00} & \vc{0.00} & \vc{0.00} & \vc{0.00} & \vc{0.00} & \vco{0.00} & \vc{0.00} & \vc{0.00} & \vc{0.00} & \vc{0.00} & \vcg{0.00} \\
\texttt{RememberColor3}               & \vc{0.07} & \vc{0.00} & \vc{0.19} & \vc{0.30} & \vc{0.41} & \vco{0.92} & \vc{0.38} & \vc{0.37} & \vc{0.28} & \vc{0.30} & \vcg{0.91} \\
\texttt{RememberColor9}               & \vc{0.05} & \vc{0.03} & \vc{0.11} & \vc{0.08} & \vc{0.11} & \vco{0.41} & \vc{0.09} & \vc{0.11} & \vc{0.11} & \vc{0.05} & \vcg{0.47} \\
\texttt{RememberShapeAndColor3x2}     & \vc{0.07} & \vc{0.03} & \vc{0.09} & \vc{0.06} & \vc{0.11} & \vco{0.59} & \vc{0.11} & \vc{0.12} & \vc{0.04} & \vc{0.11} & \vcg{0.62} \\
\texttt{RememberShapeAndColor5x3}     & \vc{0.04} & \vc{0.01} & \vc{0.06} & \vc{0.12} & \vc{0.04} & \vco{0.28} & \vc{0.15} & \vc{0.08} & \vc{0.07} & \vc{0.09} & \vcg{0.29} \\
\textbf{Average (11 envs)}            & \vc{0.03} & \vc{0.01} & \vc{0.07} & \vc{0.07} & \vc{0.09} & \vco{0.23} & \vc{0.09} & \vc{0.09} & \vc{0.08} & \vc{0.06} & \vcg{0.24} \\
\midrule
\multicolumn{12}{l}{\emph{Held-out, novel memory semantics}} \\
\midrule
\texttt{RememberShape3}               & \vc{0.05} & \vc{0.00} & \vc{0.08} & \vc{0.21} & \vc{0.27} & \vco{0.35} & \vc{0.35} & \vc{0.28} & \vc{0.22} & \vc{0.29} & \vcg{0.46} \\
\texttt{RememberShape5}               & \vc{0.04} & \vc{0.00} & \vc{0.11} & \vc{0.20} & \vc{0.21} & \vco{0.46} & \vc{0.20} & \vc{0.20} & \vc{0.20} & \vc{0.16} & \vcg{0.40} \\
\texttt{RememberShape9}               & \vc{0.00} & \vc{0.00} & \vc{0.11} & \vc{0.14} & \vc{0.11} & \vco{0.30} & \vc{0.09} & \vc{0.11} & \vc{0.13} & \vc{0.06} & \vcg{0.29} \\
\texttt{RotateLenientPos}             & \vc{0.01} & \vc{0.06} & \vc{0.04} & \vc{0.00} & \vc{0.01} & \vco{0.00} & \vc{0.00} & \vc{0.01} & \vc{0.01} & \vc{0.02} & \vcg{0.02} \\
\texttt{RotateLenientPosNeg}          & \vc{0.00} & \vc{0.03} & \vc{0.08} & \vc{0.01} & \vc{0.03} & \vco{0.02} & \vc{0.06} & \vc{0.00} & \vc{0.10} & \vc{0.02} & \vcg{0.07} \\
\texttt{RotateStrictPos}              & \vc{0.00} & \vc{0.00} & \vc{0.04} & \vc{0.03} & \vc{0.02} & \vco{0.00} & \vc{0.04} & \vc{0.00} & \vc{0.02} & \vc{0.02} & \vcg{0.05} \\
\texttt{RotateStrictPosNeg}           & \vc{0.00} & \vc{0.00} & \vc{0.03} & \vc{0.01} & \vc{0.01} & \vco{0.00} & \vc{0.06} & \vc{0.00} & \vc{0.03} & \vc{0.04} & \vcg{0.04} \\
\textbf{Average (7 envs)}             & \vc{0.00} & \vc{0.01} & \vc{0.07} & \vc{0.09} & \vc{0.09} & \vco{0.16} & \vc{0.11} & \vc{0.09} & \vc{0.10} & \vc{0.09} & \vcg{0.19} \\
\midrule
\textbf{Avg.\ over 23 environments}   & \vc{0.10} & \vc{0.10} & \vc{0.16} & \vc{0.18} & \vc{0.20} & \vco{0.34} & \vc{0.20} & \vc{0.19} & \vc{0.19} & \vc{0.07} & \vcg{0.36} \\
\bottomrule
\end{tabular}}
\vspace{-2em}
\end{table*}

\vspace{-0.5em}
\paragraph{Receding-horizon inference.}
The dataloader updates memory at every step, while standard \openvlaoft\ deploys with open-loop $H$-step action chunking~\citep{act}: under chunked deployment memory would update only once every $H$ steps, a factor-of-$H$ mismatch with training. Beyond this mismatch, task-relevant cues can appear and disappear within a single chunk, so a policy that updates memory only once per chunk may miss them entirely. We therefore use receding-horizon control~\citep{chi2023diffusion}: at every environment step we re-query the model and execute only the first action of the predicted chunk. Compute cost is reported in Appendix~\ref{app:cost}.

\vspace{-0.5em}
\section{Experiments}
\label{sec:exp}
\vspace{-0.5em}

Our experiments target four research questions (RQs) about adding
in-context recurrence to a transformer VLA:
\textbf{(RQ1)} Can recurrent memory be added at the fine-tuning stage
to a memoryless VLA pretrained on memoryless tasks, without retraining
the backbone from scratch?
\textbf{(RQ2)} Does end-to-end fine-tuning of the recurrence benefit
from cross-step gradient flow --- TBPTT versus a per-step detach
($K{=}1$) versus a learning-free EMA write --- and if so, what
truncation length $K$ is right?
\textbf{(RQ3)} How does the trained memory behave at inference time
on (a) held-out tasks that share the memory semantics of a training
task (e.g., trained on \texttt{RememberColor5}, evaluated on
\texttt{RememberColor9}) versus (b) held-out tasks whose memory
semantics conceptually differs from the training mixture (e.g., the \texttt{Rotate}
family)?
\textbf{(RQ4)} Does adding the recurrent channel introduce
performance degradation on Markovian (MDP) tasks where memory is not
required (e.g., the fully observable LIBERO suite)?

\vspace{-0.5em}
\paragraph{Suites and training mixture.}
We train and evaluate \muvla on two manipulation suites trained as
multi-task mixtures.
\mikasa~\citep{mikasarobo} is the partially observable side: we train
jointly on a five-task mixture chosen to cover the benchmark's three
memory categories (Section~\ref{sec:prelim}) --- \texttt{RememberColor5}
and \texttt{RememberShapeAndColor3x3} for cue-recall,
\texttt{ShellGamePush} for tracking through occlusion, and
\texttt{TakeItBack} together with \texttt{InterceptMedium} for
sequential and predictive memory.
The five tasks were fixed before any model evaluation; multi-task
training on this mixture lets us probe two generalization axes on the
remaining 18 \mikasa environments evaluated here.
We split the 18 held-out environments along the memory-semantic axis
relevant to RQ3.
\emph{Matched memory semantics} (11 environments) covers held-out tasks
whose memory requirement is the same as one of the training tasks but
the difficulty level or action primitive differs: the remaining
\texttt{ShellGame} variants (touch and pick), the remaining
\texttt{Intercept} difficulties (slow and fast) and the
\texttt{InterceptGrab} variants, and the larger and smaller
\texttt{RememberColor} and \texttt{RememberShapeAndColor} sizes.
\emph{Novel memory semantics} (7 environments) covers held-out tasks
whose memory requirement is not represented in the training mixture:
the \texttt{RememberShape} family (shape memory rather than color or
shape{+}color), and the \texttt{Rotate} family (the agent must internally track progress).

\vspace{-0.5em}
\paragraph{Backbone and training recipe.}
All conditions fine-tune the released \openvla checkpoint with
LoRA~\citep{lora}; the full recipe (optimizer, schedules, image
inputs, action chunking, dataloader differences across conditions)
is given in Appendix~\ref{app:method:finetune}.

\vspace{-0.5em}
\paragraph{Members of the family and reference points.}
We report ten conditions on \mikasa.
Three are memoryless reference points, included as calibration rather
than as baselines we beat:
\textbf{(R1)} \emph{OpenVLA-OFT}, the memoryless \openvlaoft with the
original training recipe and open-loop action chunking;
\textbf{(R2)} \emph{OpenVLA-OFT$^\dagger$ (episodic)}, the same memoryless model finetuned
with our episodic dataloader and $m{=}0$ memory tokens, which isolates
the dataloader change from the effect of the recurrence;
\textbf{(R3)} \emph{OpenVLA-OFT$^\dagger$ (episodic) + 1st observation}, identical to (R2) except
that the very first top-down frame of the episode is appended as a
third image input.
The first frame Markovianizes any task whose memory cue is contained
in the initial observation, and is a strong non-recurrent oracle reference
for the family.
The remaining seven span the \muvla family.
At $m{=}64$ we vary the form of the recurrent write: TBPTT at
truncation lengths $K{=}1, 2, 8$, an EMA write
($\boldsymbol{M}_{t+1} = \alpha\,\boldsymbol{M}'_t + (1-\alpha)\,\boldsymbol{M}_t$,
$\alpha=0.1$, both detached from the graph), and an EMA variant
that drops the action-copy guard of
Appendix~\ref{app:method:mask} and uses a full attention mask
instead.
We also include $m{=}1$ at $K{=}8$ as a bandwidth probe and an
$m{=}64$, $K{=}8$ control trained on \texttt{RememberColor5} alone, to
test whether the multi-task mixture interferes with the recurrent
state on any individual task.
On LIBERO we report the published \openvlaoft numbers
\citep{openvlaoft}, $m{=}64$/$K{=}8$, and the $m{=}64$/EMA configuration.
The question on the Markovian side is whether the recurrence regresses
the host model, not whether it sets a new state of the art.

\begin{wraptable}{r}{0.52\textwidth}
    \vspace{-1.5em}
    \centering
    \caption{
        \textbf{Performance comparison on LIBERO~\citep{liu2023libero}}. Success rates (\%) are reported for the four LIBERO suites and their average.
    $^\ast$ denotes memory-augmented VLA models.
    }
    \vspace{-0.5em}
    \label{tab:libero}
    \footnotesize
    \setlength{\tabcolsep}{3pt}
    \renewcommand{\arraystretch}{0.92}
    \resizebox{0.52\textwidth}{!}{
    \begin{tabular}{l|cccc|c}
        \toprule
        Method & Spatial & Object & Goal & Long-10 & Avg. \\
        \midrule
        Diffusion Policy~\citep{chi2023diffusion} & 78.3 & 92.5 & 68.3 & 50.5 & 72.4 \\
        Octo~\citep{octo}                         & 78.9 & 85.7 & 84.6 & 51.1 & 75.1 \\
        OpenVLA~\citep{kim2024openvla}            & 84.7 & 88.4 & 79.2 & 53.7 & 75.9 \\
        SpatialVLA~\citep{qu2025spatialvla}       & 88.2 & 89.9 & 78.6 & 55.5 & 71.7 \\
        UniACT~\citep{zheng2025universal}         & 77.0 & 87.0 & 77.0 & 70.0 & 76.8 \\
        $\pi_0$~\citep{pi0}            & 96.8 & 98.8 & 95.8 & 85.2 & 94.2 \\
        $\pi_0$-FAST~\citep{pertsch2025fast}     & 96.4 & 96.8 & 88.6 & 60.2 & 85.0 \\
        CogACT~\citep{li2024cogact}               & 97.2 & 98.0 & 90.2 & 88.8 & 93.2 \\
        \openvlaoft~\citep{li2024cogact}               & 97.6 & 98.4 & 97.9 & 94.5 & 97.1 \\
        CronusVLA$^\ast$~\citep{li2025cronusvla}         & 90.1 & 94.7 & 91.3 & 68.7 & 86.2 \\
        MemoryVLA$^\ast$~\citep{memoryvla}               & 98.4 & 98.4 & 96.4 & 93.4 & 96.5 \\
        \midrule
        \rowcolor{linkColor!20} \muvla~($m{=}64, K{=}8$) & 93.0 & \textbf{99.4} & 96.6 & \textbf{95.8} & 96.2 \\
        \rowcolor{linkColor!20} \muvla~($m{=}64$, EMA)     & 70.8 & 64.4 & \phantom{0}6.6 & 37.2 & 44.8 \\
        \bottomrule
    \end{tabular}}
    \vspace{-10pt}
\end{wraptable}

\vspace{-0.5em}
\paragraph{Comparison policy.}
The goal of the \mikasa\ evaluation is not to maximize benchmark
performance through an increasingly sophisticated memory system, but
to isolate the contribution of recurrence itself inside a fixed VLA
backbone. The primary comparisons therefore run within the $\mu$VLA
family and against the three memoryless reference points (R1, R2, R3)
on a single fixed training mixture, training recipe, and inference
protocol, with the recurrent state as the only varying factor. To
contextualize the resulting performance regime against the broader
literature, we additionally list the published numbers of two memory-augmented VLAs (CronusVLA~\citep{li2025cronusvla} and
MemoryVLA~\citep{memoryvla}) on the LIBERO suites where their numbers
are available (Table~\ref{tab:libero}).

\vspace{-0.5em}
\paragraph{Evaluation.}
Each method is evaluated with 100 deterministic episodes per
environment using receding-horizon inference
(Section~\ref{sec:method}); the full protocol is given in
Appendix~\ref{app:eval}.

\vspace{-0.5em}
\subsection{Results on \mikasa}
\label{sec:exp:mikasa}
\vspace{-0.5em}
Table~\ref{tab:main} reports per-environment SR on
all 23 \mikasa-VLA environments for all ten conditions, grouped into
the three blocks of RQ3: training tasks (in-distribution), held-out
with matched memory semantics, and held-out with novel memory
semantics. We summarize the headline numbers below; we defer
interpretation to Section~\ref{sec:discussion}.

\vspace{-0.5em}
\paragraph{Headline numbers per RQ.}
On the five training tasks (RQ1), recurrent fine-tuning lifts average
SR from $0.42$ (\openvlaoft) and $0.48$
(\openvlaoft$^\dagger$ episodic) to $0.84$ at $K{=}2$.
Within the \muvla family (RQ2), the $K{=}2$ truncation outperforms
both $K{=}1$ ($0.57$) and $K{=}8$ ($0.57$) on the training average
and on cue-recall tasks (e.g., \texttt{RememberColor5}: $0.93$ at
$K{=}2$ vs.\ $0.35$/$0.40$ at $K{=}1$/$K{=}8$); the EMA write tracks
the $K{=}8$ regime. On held-out evaluation (RQ3), the
matched-semantics environments follow the training trend (best $0.24$
for the first-frame oracle reference and $0.23$ at $K{=}2$, against
$0.07$ for \openvlaoft$^\dagger$ episodic), while the
novel-semantics environments remain close to the memoryless references
and the spread among recurrent variants narrows. For RQ4, LIBERO
(Section~\ref{sec:exp:libero}) shows that on a fully observable
manipulation suite the recurrent variants stay within a few
percentage points of the memoryless \openvlaoft baseline.

\vspace{-0.5em}
\subsection{Recurrence is benign on a Markovian suite (LIBERO)}
\label{sec:exp:libero}
\vspace{-0.5em}
LIBERO serves as the fully observable control suite in our evaluation.
Unlike \mikasa, the current observation in LIBERO exposes the task-relevant
objects, layout, and goals. The purpose of this evaluation is therefore
not to compare recurrent carriers, but to test whether
the strongest recurrent configuration regresses the host VLA under full
observability. The relevant comparison is \openvlaoft~\citep{openvlaoft}
versus $\mu$VLA at $K{=}8$; the EMA row in
Table~\ref{tab:libero} is included only as a recurrence-training ablation. Table~\ref{tab:libero} confirms the no-regression result. $\mu$VLA at
$K{=}8$ reaches $96.2\%$ average success across the four LIBERO suites,
matching or exceeding the strongest non-recurrent VLA baselines.
The recurrent channel therefore does not harm the host policy on a fully
observable benchmark. Gains concentrate on Object and Long-10, where
tracking object-specific or progress-related context remains useful even
without partial observability, while Spatial and Goal remain close to
\openvlaoft. The EMA write falls behind on three suites,
consistent with the \mikasa\ results.
Receding-horizon inference is not incidental to these numbers:
evaluating the same $\mu$VLA ($m{=}64$, $K{=}8$) checkpoint with
open-loop $H$-step chunked execution instead collapses Long-10 from
$95.8$ to $5.4$ and Goal from $96.6$ to $35.8$. This reflects the
train-test mismatch discussed in Section~\ref{sec:method}: under
chunked execution the recurrent state is updated only once per chunk,
a regime the model was never trained in.

\vspace{-0.5em}
\subsection{Memory diagnostics and inference-time ablations}
\label{sec:exp:diagnostics}
\vspace{-0.5em}
We probe \emph{how} the recurrent channel is used through seven inference-time diagnostics, run on three already-trained \muvla\ checkpoints (TBPTT $K{=}8$, EMA, TBPTT $K{=}2$) and on the five in-distribution \mikasa\ training tasks. Full per-environment outputs and extended discussion are in Appendix~\ref{app:diagnostics}.

\begin{wrapfigure}{r}{0.6\textwidth}
\vspace{-1.5em}
\centering
\includegraphics[width=\linewidth]{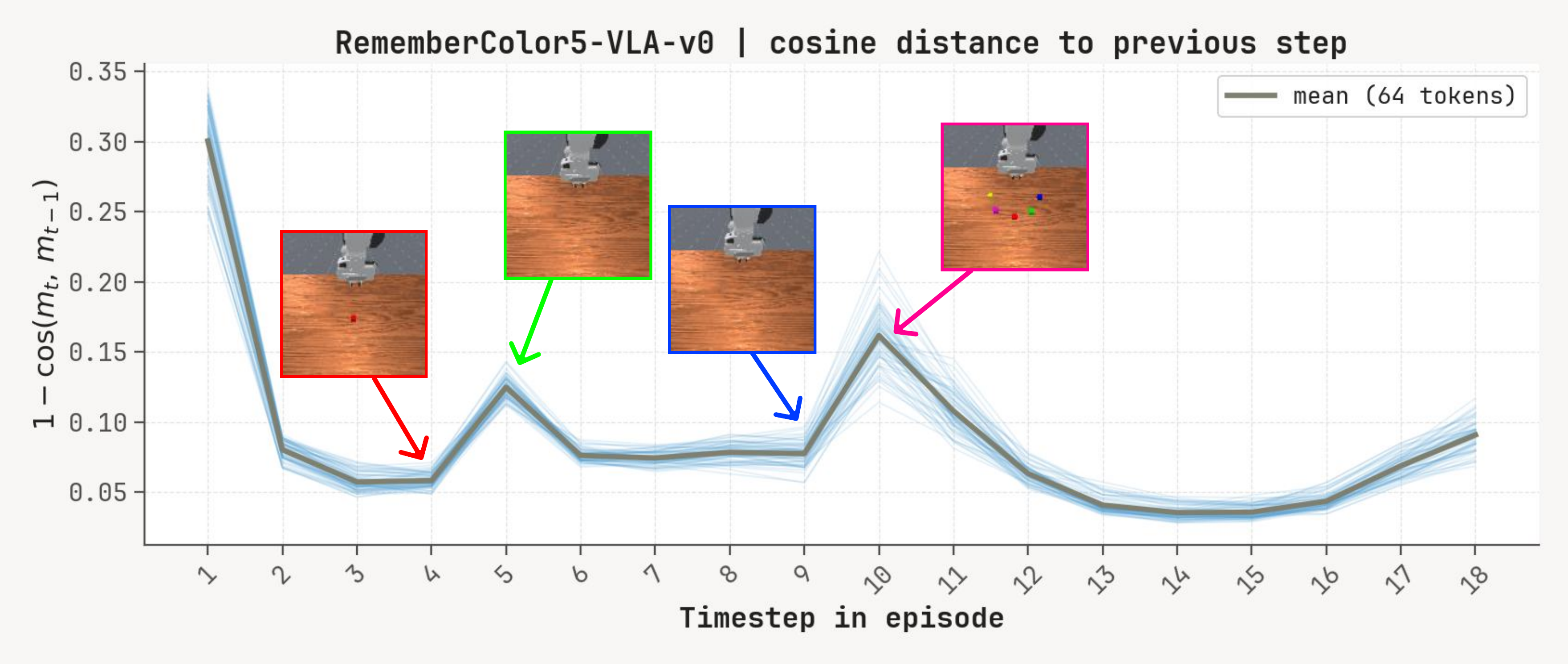}
\caption{\textbf{Cosine distance $M_t$ vs. $M_{t-1}$ on \texttt{RememberColor5} ($K{=}2$, $m{=}64$).} Pale blue lines: individual memory tokens; dark grey: mean over all $64$. The episode contains two phase transitions: the cue cube is visible until \pht{phRed}{4}; at \pht{phGreen}{5} it disappears and the agent observes an empty table until \pht{phBlue}{9}, while at \pht{phPink}{10} the candidate cubes appear and the task switches from memory maintenance to retrieval and action execution. Colors match the frames of the inset snapshots. Both transitions produce clear spikes in memory change.}
\label{fig:cosine_dynamics}
\vspace{-1.0em}
\end{wrapfigure}

\paragraph{Representation dynamics and attention rollouts.}
\label{sec:exp:diagnostics:01}
On the strongest carrier ($K{=}2$, $m{=}64$), the recurrent state exhibits clear phase-conditioned dynamics (Figure~\ref{fig:cosine_dynamics}). In \texttt{RememberColor5}, the agent observes the cue cube until \pht{phRed}{4}; at \pht{phGreen}{5} the cue disappears and the environment remains empty until \pht{phBlue}{9}. At \pht{phPink}{10}, the candidate cubes appear and the task switches from memory maintenance to retrieval and action execution. Both transitions produce distinct spikes in $1{-}\cos(\boldsymbol{M}_t,\boldsymbol{M}_{t-1})$, indicating that the recurrent state reorganizes sharply when the latent phase changes. Similar transition-aligned effects appear in other environments: \texttt{TakeItBack} produces a second spike at the return-phase transition, while \texttt{InterceptMedium} sustains elevated dynamics around the catch event. Per-token heatmaps confirm that this structure is concentrated in a small subset of memory tokens rather than distributed uniformly across all $64$.

\begin{wrapfigure}{r}{0.42\textwidth}
\vspace{-1.2em}
\centering
\includegraphics[width=\linewidth]{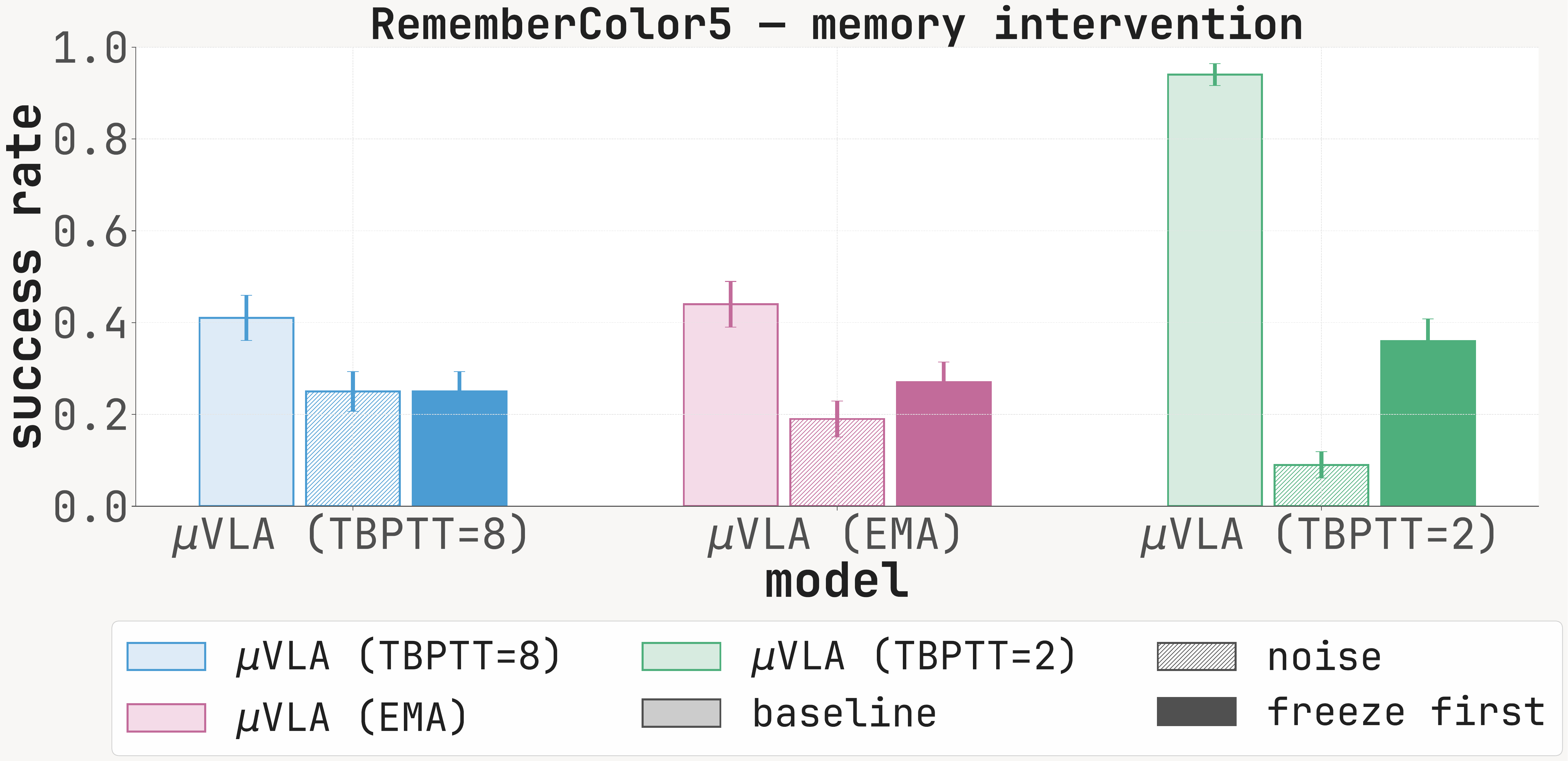}
\caption{\textbf{Memory intervention on \texttt{RememberColor5}.} SR at $100$ episodes under baseline, \texttt{noise}, and \texttt{freeze\_first}.}
\label{fig:intervention}
\vspace{-1.0em}
\end{wrapfigure}

\begin{figure*}[t]
\vspace{-0.5em}
\centering
\begin{minipage}{0.98\textwidth}\centering
  \includegraphics[width=\linewidth]{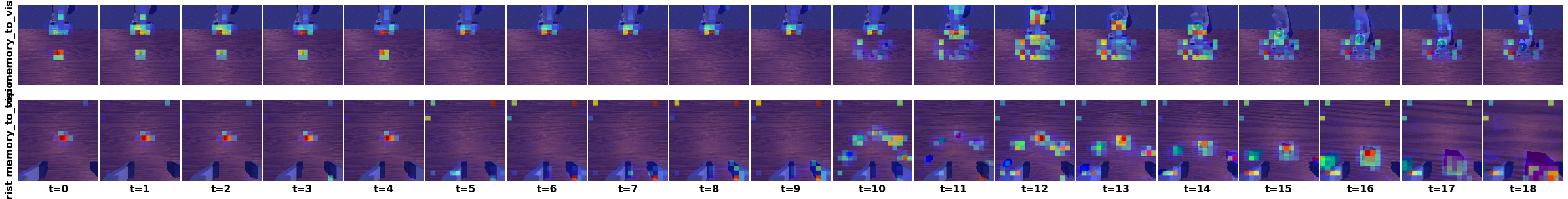}\\[-2pt]
  \footnotesize (c) memory$\to$vision attention.
\end{minipage}\\[0.3em]
\vspace{-0.5em}
\caption{\textbf{Memory$\to$vision attention rollout on \texttt{RememberColor5}
($K{=}2$, $m{=}64$).} The cue panel is visible early in the episode and
then masked. Memory attention concentrates on the cue grid while it is
visible and becomes diffuse after occlusion, matching the phase
transition in Figure~\ref{fig:mem_dynamics_remsac}.}
\label{fig:attn_rollout_rc5_1}
\vspace{-1em}
\end{figure*}

Composing per-layer attention with rollout (mean over heads, residual factor $0.5$) on three query groups (action$\to$vision, memory$\to$vision, vision$\to$vision) yields overlays (Appendix~\ref{app:attn-rollouts}) in which action attention tracks the gripper and target object, while memory attention concentrates on the cue region before masking and diffuses afterward (Figure~\ref{fig:attn_rollout_rc5_1}). Interestingly, the sharp transition peaks in memory dynamics suggest that recurrent-state change itself could potentially serve as a learned key-frame signal, allowing the policy to identify semantically important events without relying on an external VLM-based selector as in \citep{memer}. Per-environment overlays and four-panel dynamics cards for the remaining tasks are provided in Appendices~\ref{app:attn-rollouts} and~\ref{app:mem-dynamics}.

\vspace{-0.5em}
\paragraph{Causal intervention on the memory channel.}
\label{sec:exp:diagnostics:06}
We re-run the $100$-episode \mikasa\ validation under two interventions: \texttt{noise} replaces $\boldsymbol{M}_t$ with i.i.d.\ Gaussian noise before every forward; \texttt{freeze\_first} locks the state to $\boldsymbol{M}_1$ (Figure~\ref{fig:intervention}). \texttt{noise} drops SR sharply in every (carrier, env) cell with a non-trivial baseline (\texttt{RememberColor5}, $K{=}2$: $0.94\!\to\!0.09$; \texttt{TakeItBack}, $K{=}2$: $0.99\!\to\!0.21$), confirming the channel is functionally read at inference. \texttt{freeze\_first} preserves SR on first-frame-cue tasks (\texttt{RC5}, $K{=}2$: $0.36$) but collapses on dynamics-grounded tasks (\texttt{InterceptMedium}, $K{=}2$: $0.07$). Per-env panels in Appendix~\ref{app:intervention-full}.

\vspace{-0.5em}
\paragraph{Robustness to chunked inference.}
\label{sec:exp:diagnostics:07}
Sweeping the open-loop chunk length over $\{1,2,4,8,16\}$ with memory active versus zeroed shows the receding-horizon regime (chunk${=}1$) is where the with-memory minus no-memory gap is largest on cue-recall (\texttt{RememberColor5}, $K{=}2$: $0.94$ vs $0.00$; \texttt{TakeItBack}, $K{=}2$: $0.99$ vs $0.00$). At chunk${\geq}2$, SR collapses on cue-recall tasks for every carrier, while \texttt{ShellGamePush} and \texttt{InterceptMedium} partially recover at chunk${=}8$ where a stereotyped open-loop motion suffices, a statement about training-inference cadence alignment rather than memory dependence. Per-env panels in Appendix.

\vspace{-0.5em}
\paragraph{Length and OOD-cue generalization.}
\label{sec:exp:diagnostics:08}
\label{sec:exp:diagnostics:09}
On \texttt{RememberColor5-Phase}$N$ variants ($N\!\in\!\{3,5,10,20,50,100\}$ per phase, training $N\!\in\!\{1,\dots,5\}$), $K{=}2$ leads in-distribution ($N{=}3$: $0.93$) but loses ${\sim}75\%$ of its recall by $N{=}20$; $K{=}8$ is flat at $0.31$--$0.38$ across all $N$; EMA collapses below $0.15$ at $N\!\geq\!20$. On color-swap variants ($1$--$5$ in-distribution colors replaced by an OOD palette), $K{=}2$ is the most robust at every level ($0.48$ at five swaps) and no carrier falls to chance, indicating a partially abstract cue encoding. Curves and inference-cost are in Appendix~\ref{app:diagnostics} (Figure~\ref{fig:rc5_generalization}) and Appendix~\ref{app:cost}.

\vspace{-1em}
\section{Discussion}
\label{sec:discussion}
\vspace{-1em}

We summarize the four RQs and defer full analysis to Appendix~\ref{app:discussion-extended}.

\vspace{-1em}
\paragraph{RQ1: Recurrence can be added at fine-tuning.} Adding $m{=}64$ recurrent memory tokens to a memoryless \openvlaoft\ checkpoint raises the training-mixture average from $0.42$ (OFT-RLDS) and $0.48$ (OFT-epis.) to $0.84$ at $K{=}2$, with the lift concentrated on cue-recall (\texttt{RememberColor5}: $0.04/0.09\!\to\!0.93$). The episodic dataloader contributes $+6$\,pp ($0.42\!\to\!0.48$), memory bandwidth $+9$\,pp at $K{=}8$, and TBPTT length is the dominant lever for cue-recall.

\vspace{-1em}
\paragraph{RQ2: Cross-step gradients matter; the right $K$ is short.} TBPTT length traces a U-shape: $K{=}2$ reaches $0.93$ on \texttt{RememberColor5} against $0.35/0.40$ at $K{=}1/K{=}8$, average climbing from $0.57$ to $0.84$. Chunk-length and phase-length sweeps plus the noise intervention place $K{=}2$ at a sweet spot in the credit-assignment versus signal-resolution trade-off. A detached EMA write tracks $K{=}8$ on the training average but trails $K{=}2$ on cue-recall.

\vspace{-1em}
\paragraph{RQ3: Behavior under memory-semantic shift.} The family transfers on matched semantics ($K{=}2$: $0.23$ vs $0.07$ for OFT-epis.); the first-frame reference matches $K{=}2$ on cue-recall but trails when the cue is not in the first frame (\texttt{TakeItBack}: $0.99$ vs $0.94$). On novel semantics the recurrent family stays close to the memoryless references, calibrating the capability envelope at fixed $m$ and training distribution.

\vspace{-0.5em}
\paragraph{RQ4: No penalty on Markovian tasks.} On LIBERO the recurrent variants stay competitive with the memoryless \openvlaoft baseline across all four task families, proving the method versatility.

\vspace{-0.5em}
\section{Conclusion}
\vspace{-0.5em}
We studied recurrence as an isolated design variable inside a modern
VLA backbone. Across controlled experiments on \mikasa, minimal
in-backbone recurrence substantially improves performance on partially
observable manipulation tasks that are difficult or impossible for a
memoryless policy, while preserving performance on fully observable
control suites such as LIBERO. The results suggest that even a small
recurrent state qualitatively changes the operating regime of VLA
models, but also reveal clear limits: gains transfer only partially
across unseen memory semantics, and long-horizon dependencies remain
fragile under minimal recurrence alone.
More broadly, our findings argue that recurrence itself is already a
strong and understudied ingredient in VLA systems, even before adding
retrieval, hierarchical memory, or external planning modules. We hope
the controlled setup introduced here helps future work disentangle
which capabilities arise from recurrence itself and which require more
structured memory mechanisms.



\bibliographystyle{plainnat}
\bibliography{bib}


\newpage
\appendix

\section{Limitations and Scope}
\label{app:limitations}

Our recurrent formulation is trained with TBPTT, so gradients do not span entire episodes and the method
is not intended as a solution for arbitrarily long-horizon memory.
However, the results show that even short truncation windows are
sufficient to learn stable recurrent updates that support cue-recall,
phase tracking, and short-term latent-state maintenance across
partially observable manipulation tasks. In this sense, the recurrent
state can already function as a compact online state estimator that
maintains task-relevant information beyond the current observation and
potentially supplies structured signals for higher-level long-horizon
memory systems.

A second scope condition is that recurrence is introduced only at the
fine-tuning stage, with a fixed pretrained backbone and LoRA rank
$32$. The reported results should therefore be interpreted as a
measurement of what minimal in-backbone recurrence alone contributes
under a constrained adaptation budget, rather than as the upper limit
of recurrent VLA performance. Scaling the recurrent state, extending
the adaptation capacity, or combining recurrence with more structured
memory hierarchies are likely directions for closing the remaining gap
on harder held-out memory regimes.

\section{Extended Discussion}
\label{app:discussion-extended}

This appendix expands the four-paragraph summary in Section~\ref{sec:discussion}.

\subsection{RQ1: Recurrence can be added at fine-tuning}

The main result is that adding recurrent memory tokens during
fine-tuning, on top of a memoryless \openvlaoft\ checkpoint, raises the
training-mixture average SR from $0.42$ (\openvlaoft) and $0.48$
(\openvlaoft$^\dagger$ episodic) to $0.84$ at $m{=}64$, $K{=}2$.
The gain is concentrated on cue-recall tasks:
\texttt{RememberColor5} improves from $0.04$/$0.09$ to $0.93$, and
\texttt{RememberShapeAndColor3x3} from $0.03$/$0.13$ to $0.86$.
This shows that recurrence can be introduced into a pretrained
transformer VLA through fine-tuning alone, without modifying the
backbone architecture.

Part of the improvement comes from the training procedure itself.
Switching from shuffled single-timestep training to the episodic
dataloader of Appendix~\ref{app:method:dataloader}, while keeping
$m{=}0$, already increases the training average from $0.42$ to $0.48$
and improves transfer on matched-semantics environments
(\texttt{InterceptFast}: $0.00\!\to\!0.33$,
\texttt{RememberColor3}: $0.00\!\to\!0.19$).
The correct reference for recurrence is therefore
\openvlaoft$^\dagger$ episodic rather than the original shuffled
training setup.

Memory bandwidth contributes less than cross-step optimization.
At $K{=}8$, increasing memory width from $m{=}0$ to $m{=}1$ raises the
training average from $0.48$ to $0.54$, while increasing from
$m{=}1$ to $m{=}64$ gives only another $+3$\,pp. Even a single
recurrent token carries useful information, but the dominant factor for
cue-recall is the recurrent optimization regime itself, which we study
in RQ2. Section~\ref{sec:exp:diagnostics:06} further confirms that the
policy actively reads the recurrent state at inference:
replacing $\boldsymbol{M}_t$ with i.i.d.\ Gaussian noise sharply drops
SR on all cue-recall tasks.

\subsection{RQ2: Cross-step gradients matter, and the right $K$ is short}

Holding $m{=}64$ fixed, the TBPTT length $K$ produces a U-shaped
performance curve. On \texttt{RememberColor5}, $K{=}2$ reaches $0.93$
versus $0.35$ at $K{=}1$ and $0.40$ at $K{=}8$; on
\texttt{RememberShapeAndColor3x3}, $0.86$ versus $0.12$ and $0.09$.
The training-mixture average rises from $0.57$ at both $K{=}1$ and
$K{=}8$ to $0.84$ at $K{=}2$.

We interpret this as the result of two competing effects.
At $K{=}1$, cross-step gradients are removed entirely, so the recurrent
write receives no future supervision signal indicating what information
should persist across timesteps. The recurrent state therefore behaves
mostly as a passive encoding of the current observation.
At $K{=}8$, the write must propagate gradients through long recurrent
chains across repeated forward passes of a large pretrained backbone,
which appears to produce a more diffuse and lower-resolution recurrent
representation.

Several diagnostics support this interpretation.
First, the $K{=}8$ representation is flatter under chunk-length sweeps
(Section~\ref{sec:exp:diagnostics:07}): the performance gap between
memory-enabled and memory-disabled inference decreases more slowly as
the chunk length grows. Second, under phase-length shifts,
$K{=}8$ remains relatively stable across delays, while $K{=}2$
achieves stronger in-distribution recall but degrades faster outside
the training horizon. Third, Gaussian-noise interventions sharply
degrade both $K{=}2$ and $K{=}8$
(\texttt{RememberColor5}: $0.94\!\to\!0.09$ at $K{=}2$),
showing that both variants actively use the recurrent state rather than
collapsing to a memoryless solution.

The EMA write reaches nearly the same average SR as $K{=}8$, but falls
far behind $K{=}2$ on cue-recall
(\texttt{RememberColor5}: $0.44$ versus $0.93$;
\texttt{RememberShapeAndColor3x3}: $0.09$ versus $0.86$),
indicating that detached low-pass updates are insufficient for precise
memory retention.

\subsection{RQ3: Behavior under memory-semantic shift}

Held-out behavior separates clearly along the memory-semantic split of
Table~\ref{tab:main}. On matched-semantics environments, whose memory
requirements are already represented in the training mixture,
the recurrent family follows the same trend as the training tasks:
$m{=}64$, $K{=}2$ reaches $0.23$ average SR versus $0.07$ for
\openvlaoft$^\dagger$ episodic and $0.01$ for the original
\openvlaoft setup. The largest gains appear on
\texttt{RememberColor3} ($0.19\!\to\!0.92$) and
\texttt{RememberShapeAndColor3x2} ($0.09\!\to\!0.59$).

The first-frame oracle reference remains competitive on this block and
matches or exceeds recurrent variants on several cue-recall tasks
(\texttt{RememberColor5}: $0.96$ versus $0.93$;
\texttt{RememberShapeAndColor3x3}: $0.91$ versus $0.86$).
We interpret this as a positive result rather than a weakness:
on these environments the relevant latent information is fully visible
in the initial observation, and the recurrent state learns to preserve
precisely this information. Where the first frame is insufficient,
however, recurrence becomes necessary.
On \texttt{TakeItBack}, recurrent variants reach $0.99$ versus $0.94$
for the first-frame oracle and $0.87$ for
\openvlaoft$^\dagger$ episodic.

The mask ablation directly supports the action-copy argument of
Appendix~\ref{app:method:mask}. Removing the memory-action guard
degrades cue-recall
(\texttt{RememberColor5}: $0.44\!\to\!0.25$;
\texttt{RememberColor3}: $0.37\!\to\!0.28$),
while improving motor-heavy environments
(\texttt{ShellGamePush}: $0.77\!\to\!0.94$;
\texttt{InterceptFast}: $0.28\!\to\!0.35$).
This matches the predicted degeneracy:
without the guard, the recurrent state shifts toward encoding action
trajectories instead of latent cue identity.

On novel-semantics environments, whose memory requirements are absent
from the training mixture, the spread among recurrent variants narrows
substantially. The first-frame oracle performs best ($0.19$),
$m{=}64$, $K{=}2$ follows at $0.16$, and the remaining recurrent
variants cluster around $0.09$--$0.11$.
Nevertheless, some transfer remains:
\texttt{RememberShape5} reaches $0.46$ at $K{=}2$ despite the model
never being trained on shape-only memory tasks.

Sections~\ref{sec:exp:diagnostics:08}
and~\ref{sec:exp:diagnostics:09} further isolate two memory-semantic
distribution shifts on \texttt{RememberColor5}. The $K{=}2$ variant
achieves the strongest in-distribution recall but loses most of its SR
once the cue-to-action delay exceeds the training range, whereas
$K{=}8$ trades lower absolute SR for greater phase-length invariance.
We interpret this not as a failure of recurrence, but as a calibration
of the capability envelope of minimal in-backbone recurrent state at
fixed memory width and training distribution.

\subsection{RQ4: No penalty on Markovian tasks}

On tasks where the current observation already exposes everything
needed for control, recurrence should be inert: the recurrent state
has nothing useful to carry, and the only question is whether it
hurts. LIBERO (Section~\ref{sec:exp:libero}) confirms that it does
not: \muvla\ at $m{=}64$, $K{=}8$ reaches $96.2\%$ average SR across
the four suites, matching or exceeding the memoryless \openvlaoft\
baseline, with no consistent ordering between the recurrent and
memoryless models on any suite.

We additionally test whether multi-task training harms individual
memory tasks. The same configuration ($m{=}64$, $K{=}8$) trained only
on \texttt{RememberColor5} reaches $0.16$ SR,
whereas the multi-task model reaches $0.35$ SR. The result provides no evidence that the recurrent state is degraded by the multi-task mixture.

\paragraph{Scope of the demonstration source.}
\mikasa\ demonstrations are generated by scripted oracle policies with
privileged simulator access. A natural concern is that the recurrent
policy could exploit fixed trajectory structure instead of reading the
latent cue itself. Two probes argue against this interpretation.
First, replacing $\boldsymbol{M}_t$ with i.i.d.\ Gaussian noise during
inference sharply degrades cue-recall performance
(\texttt{RememberColor5}: $0.94\!\to\!0.09$ at $K{=}2$;
\texttt{TakeItBack}: $0.99\!\to\!0.21$),
showing that the policy actively depends on the recurrent state rather
than on a memorized trajectory template.
Second, the OOD cue-identity sweep replaces in-distribution colors
with unseen palettes. A literal cue-lookup strategy would collapse to
chance under this shift, yet $K{=}2$ retains $0.48$ SR even when all
colors are replaced, indicating that the recurrent representation is at
least partially abstract rather than a direct RGB memorization.

\section{Recurrent and History-Aware Policy Comparison}
\label{app:recurrent-comparison}

Table~\ref{tab:recurrent-comparison} compares \muvla with the closest
recurrent or history-aware robot policy designs. The table is not a
performance ranking: the methods differ in backbone, data, action
parameterization, training recipe, and evaluation protocol. We instead
separate the axes that matter for this paper: where history is stored,
whether gradients pass through the temporal memory update, whether
training uses losses beyond the action objective, whether the method
explicitly blocks action-token leakage into memory, and whether the
recurrence operates across environment steps or only inside a single
model call. VPWEM is included as an adjacent history-aware visuomotor
policy rather than a VLA-backbone method.

\begin{table}[ht!]
\caption{
  \textbf{\muvla compared with recurrent and history-aware robot policy designs.}
  \emph{History mechanism} indicates where past information is represented.
  \emph{Temporal grad.} indicates whether gradients flow through the
  cross-step memory update. \emph{Extra loss} denotes supervision or
  regularization beyond the action loss. \emph{Action-copy guard}
  indicates whether the design explicitly prevents action tokens from
  being written into recurrent memory. \emph{Cadence} distinguishes
  environment-step memory from within-call iterative refinement.
}
\label{tab:recurrent-comparison}
\centering
\setlength{\tabcolsep}{3pt}
\begin{adjustbox}{width=\textwidth}
\begin{tabular}{l l c c c l}
\toprule
\textbf{Method}
& \textbf{History mechanism}
& \textbf{Temporal grad.}
& \textbf{Extra loss}
& \textbf{Action-copy guard}
& \textbf{Cadence} \\
\midrule
ReMem-VLA~\citep{rememvla}
& Frame- and chunk-level recurrent queries
& no; EMA update
& yes
& ---
& frame / chunk \\

AVA-VLA~\citep{avavla}
& Recurrent belief state for visual attention
& yes; TBPTT
& yes
& ---
& per step \\

Recurrent-Depth VLA~\citep{recdepthvla}
& Weight-tied latent action refinement
& ---
& ---
& ---
& within call \\

VPWEM~\citep{vpwem}
& Working memory and episodic compressor
& ---
& ---
& ---
& per step \\

\textbf{\muvla (ours)}
& Backbone self-attention with $m$ memory tokens
& yes; TBPTT
& no
& yes; attention mask
& per step, receding horizon \\
\bottomrule
\end{tabular}
\end{adjustbox}
\end{table}

\section{Method Details}
\label{app:method}

This appendix collects the technical details deferred from
Section~\ref{sec:method}: the placement of memory tokens
(Section~\ref{app:method:placement}), the full attention mask and the
information-theoretic argument for the context-to-action zero
(Section~\ref{app:method:mask}), the episodic dataloader mechanics
(Section~\ref{app:method:dataloader}), the fine-tuning protocol
(Section~\ref{app:method:finetune}), and the TBPTT loop with the reset
operator (Section~\ref{app:method:tbptt}).

\subsection{Memory token placement}
\label{app:method:placement}

We place memory tokens at the end of the multimodal prefix, after the
vision patches and the proprioception token but before the language
instruction. This is a natural location for a recurrent state in a VLA
backbone: at this point the multimodal observation has already been
laid out as a sequence of tokens, and what follows is the
task-conditional language and action region. From an attention
mechanics standpoint the choice is also flexible. The \openvlaoft
input context uses bidirectional self-attention within the context
block, so any context token can attend to any other context token
regardless of their relative order. The placement of memory tokens
\emph{inside} the input context therefore does not change which
tokens read from or write to the recurrent state, and memory does not need to sit at the very end
of the input sequence as it would in a strictly causal language
model. We use this freedom to keep the leading per-step inputs at
their original positional embeddings, so that only the language and
action tokens shift by a constant offset of $m$.

\subsection{Attention mask}
\label{app:method:mask}

\begin{table}[ht!]
\caption{
  \textbf{Per-step attention mask used by \muvla.}
  Rows are queries, columns are keys; a $1$ entry means the query
  group attends to the key group.
  The context-to-action block (top-right $5\!\times\!2$) is zero, so
  memory tokens, like the rest of the input context, never see the action
  region. This is the only departure from the \openvlaoft attention
  pattern.
}
\label{tab:mask}
\centering
\small
\setlength{\tabcolsep}{6pt}
\begin{tabular}{lccccccc}
\toprule
Query $\backslash$ Key
  & \texttt{BOS} & \texttt{VISION} & \texttt{PROPRIO}
  & \texttt{MEM} & \texttt{TEXT}
  & \texttt{ACTION} & \texttt{STOP} \\
\midrule
\texttt{BOS}            & 1 & 1 & 1 & 1 & 1 & 0 & 0 \\
\texttt{VISION\_PATCHES} & 1 & 1 & 1 & 1 & 1 & 0 & 0 \\
\texttt{PROPRIO}        & 1 & 1 & 1 & 1 & 1 & 0 & 0 \\
\texttt{MEM($m$)}       & 1 & 1 & 1 & 1 & 1 & 0 & 0 \\
\texttt{TEXT}           & 1 & 1 & 1 & 1 & 1 & 0 & 0 \\
\texttt{ACTION\_TOKENS} & 1 & 1 & 1 & 1 & 1 & 1 & 1 \\
\texttt{STOP}           & 1 & 1 & 1 & 1 & 1 & 1 & 1 \\
\bottomrule
\end{tabular}
\end{table}

The action-copy block is load-bearing. Without it, the recurrence admits a degenerate solution
$\boldsymbol{M}_t^{(i)} = \phi\bigl(\text{ACTION}_t^{(i)}\bigr)$, in
which the model writes a learnable projection of the predicted action
chunk into memory. This satisfies the recurrence exactly, but stores
nothing about the environment. The action tokens at step $t$ are
themselves a deterministic function of $(o_t, \ell,
\boldsymbol{M}_{t-1})$ produced in the same forward pass, so the
recurrence becomes self-referential and the latent state collects no
new bits across timesteps. Zeroing the context-to-action block forces
the write to be conditioned on the actual environment cues
(\texttt{VISION}, \texttt{PROPRIO}) and on the instruction
(\texttt{TEXT}). The argument is information-theoretic, so it remains
valid even under self-rollout fine-tuning (DAgger or RL
post-training), where the predicted action does affect the next
observation.

\subsection{Episodic dataloader details}
\label{app:method:dataloader}

The dataset maintains $B$ independent \emph{streams}, one per batch
element. Each stream walks a single episode step by step from $t{=}0$
to its final step, then samples a new episode uniformly at random from
the training pool and continues with that episode's first step. The
dataset yields steps in a fixed round-robin order, taking one step
from stream $0$, then one from stream $1$, and so on up to stream
$B{-}1$. The underlying PyTorch \texttt{DataLoader} collates these $B$
yields into a tensor of shape $(B, \cdot)$ in which slot $b$ always
belongs to stream $b$. Two consecutive batches advance every stream
by exactly one step, so the recurrent state stored in slot $b$ at
batch $\tau$ is the correct input for the model's prediction at batch
$\tau{+}1$. The order is deterministic given the random seed. For
distributed training, each rank offsets its seed by its rank index,
so the effective batch diversity is $B \times W$ for $W$ data-parallel
ranks.

Every yielded step carries a boolean \texttt{is\_first} flag that is
true exactly when a new episode opens on that stream. Episode
boundaries occur at arbitrary positions within a batch and within the
TBPTT window of Section~\ref{app:method:tbptt}, because the streams
are independent. The recurrent loop reads the flag per batch element
and re-seeds the memory of just those streams that started a new
episode, using the learnable initial state
$\boldsymbol{M}^{\text{init}}$ via the reset operator
(Eq.~\eqref{eq:resetop}).

When the training mixture spans multiple environments, as in the
five-task \mikasa mixture used in our main results, all environments
contribute to the same shared episode pool. Action statistics
($q_{01}/q_{99}$) are computed jointly across the pool, and a stream
moves between environments at episode boundaries. The language
instruction changes with the new episode, and \texttt{is\_first}
signals the recurrent reset for that stream.

\subsection{Fine-tuning protocol}
\label{app:method:finetune}

All conditions start from the released \openvlaoft checkpoint and are
fine-tuned with LoRA~\citep{lora} of rank 32, AdamW, two camera views
($224{\times}224$, top-down + wrist), proprioception, and action chunk
size $H{=}8$.
\muvla runs use a cosine learning-rate schedule
($\alpha{=}5{\times}10^{-4}$, 2000 warmup steps, minimum ratio 0.1) and
the episodic dataloader of Appendix~\ref{app:method:dataloader};
the memoryless baseline that mirrors the \openvlaoft recipe uses its
original constant-then-decay schedule and the standard RLDS
dataloader.
Episodes are RLDS-formatted and identical across suites.

Each \mikasa task ships its own per-task language instruction, which
is presented through the \texttt{TEXT} group of the input context and
switches at episode boundaries together with the recurrent reset.
The vision encoder, language tokenizer, and base transformer weights
are inherited from the \openvlaoft checkpoint and are not updated
outside of the LoRA adapters. Remaining numerical hyperparameters
(learning rate, schedule, optimizer, batch size, number of GPUs) are
listed in Appendix~\ref{app:hparams}.

Per-condition training wall-clock and peak memory are reported alongside the inference-cost summary in Section~\ref{sec:exp:diagnostics}.

\begin{figure}[t]
  \centering
  \includegraphics[width=\linewidth]{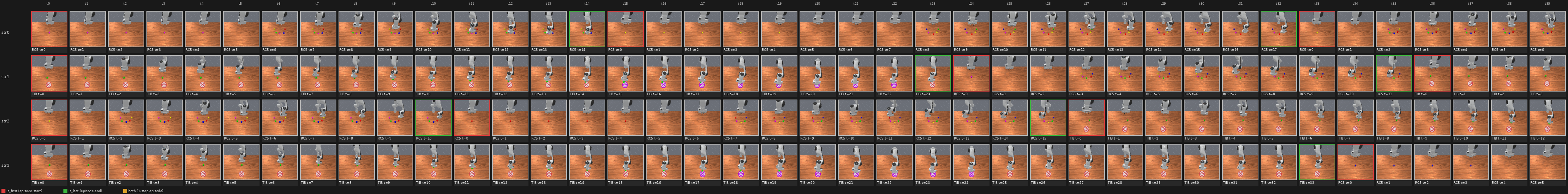}
  \caption{
    \textbf{Round-robin episodic dataloader.}
    Each batch slot $b \in \{0, \ldots, B{-}1\}$ is assigned an
    independent stream that walks one episode step by step before
    sampling a new episode. Steps are emitted in fixed round-robin
    order so that slot $b$ in two consecutive batches always belongs
    to the same stream. The recurrent state for slot $b$ accumulated
    at batch $\tau$ is therefore the correct input at batch $\tau{+}1$.
    An \texttt{is\_first} flag marks episode boundaries and triggers a
    per-stream memory reset via $\boldsymbol{M}^{\text{init}}$.
    In contrast, the standard \openvlaoft pipeline interleaves steps
    from all episodes in a shuffle buffer, destroying the temporal
    order a recurrent state requires.
  }
  \label{fig:round_robin_dataloader}
\end{figure}

\subsection{TBPTT loop}
\label{app:method:tbptt}

TBPTT unrolls each batch stream for $K$ consecutive environment steps
without detaching the recurrent memory state. We accumulate the L1
chunk loss over the unroll and then take a single backward pass through
the resulting $K$-step recurrent graph. The memory tensor is detached
only after the optimizer step, at the truncation boundary. Thus, a loss
at step $t{+}k$ can propagate gradients through up to $k$ memory
updates within the current window. If an episode boundary occurs inside
the window, only the corresponding batch streams are reset to the
learnable initial memory $\boldsymbol{M}^{\mathrm{init}}$; other streams
continue from their current recurrent state.

We implement per-stream resets with the operator
\begin{equation}
  \textsc{ApplyReset}
  \!\left(
    \boldsymbol{M},
    \texttt{is\_first},
    \boldsymbol{M}^{\mathrm{init}}
  \right)_{b}
  \;\triangleq\;
  \begin{cases}
    \boldsymbol{M}^{\mathrm{init}} & \text{if } \texttt{is\_first}[b], \\
    \boldsymbol{M}_{b} & \text{otherwise,}
  \end{cases}
  \label{eq:resetop}
\end{equation}
where $b$ indexes the batch stream. The reset branch keeps
$\boldsymbol{M}^{\mathrm{init}}$ in the compute graph, so streams that
start a new episode inside the unroll contribute gradients to the
initial-memory parameter. Algorithm~\ref{alg:tbptt} summarizes the loop.

\begin{algorithm}[ht!]
\caption{TBPTT training loop for \muvla.
$\boldsymbol{M}\in\mathbb{R}^{B\times m\times d}$ stores the recurrent
memory for all $B$ batch streams. The dataloader emits a per-stream
episode-boundary flag $\texttt{is\_first}\in\{0,1\}^{B}$.
\textsc{ApplyReset} is defined in Eq.~\eqref{eq:resetop}.}
\label{alg:tbptt}
\begin{algorithmic}[1]
\Require Episodic dataloader $\mathcal{D}$ with $B$ streams;
model $f_{\theta}$; learnable initial memory
$\boldsymbol{M}^{\mathrm{init}}$; TBPTT length $K$; optimizer
\State $\boldsymbol{M} \leftarrow \boldsymbol{M}^{\mathrm{init}}$
replicated across the batch dimension
\State $\mathcal{L}_{\mathrm{acc}} \leftarrow 0$;\quad $k \leftarrow 0$
\For{each batch $\mathcal{B}=(o,\ell,a,\texttt{is\_first})$ from $\mathcal{D}$}
  \State $\boldsymbol{M} \leftarrow
  \textsc{ApplyReset}
  \!\left(
    \boldsymbol{M},
    \texttt{is\_first},
    \boldsymbol{M}^{\mathrm{init}}
  \right)$
  \Comment{Reset only streams starting a new episode}
  \State $\hat{a},\,\boldsymbol{M}' \leftarrow
  f_{\theta}(o,\ell,\boldsymbol{M})$
  \Comment{$\boldsymbol{M}'$ remains in the recurrent graph}
  \State $\mathcal{L}_{\mathrm{acc}} \leftarrow
  \mathcal{L}_{\mathrm{acc}} + \frac{1}{K}\,\ell_{\mathrm{chunk}}(\hat{a},a)$
  \State $\boldsymbol{M} \leftarrow \boldsymbol{M}'$;\quad
  $k \leftarrow k+1$
  \If{$k=K$}
    \State $\mathcal{L}_{\mathrm{acc}}.\textsc{backward}()$
    \Comment{Backpropagate through the $K$-step recurrent unroll}
    \State $\textsc{optimizer.step}()$;\quad
    $\textsc{optimizer.zero\_grad}()$
    \State $\boldsymbol{M} \leftarrow \boldsymbol{M}.\textsc{detach}()$
    \Comment{Cut the recurrent graph at the truncation boundary}
    \State $\mathcal{L}_{\mathrm{acc}} \leftarrow 0$;\quad
    $k \leftarrow 0$
  \EndIf
\EndFor
\end{algorithmic}
\end{algorithm}

For the EMA variant, we use the same episodic dataloader and
per-stream reset logic, but remove cross-step gradient flow. Each step
is optimized with a local action loss, and the recurrent state is
updated by a detached exponential moving average:
\[
  \boldsymbol{M}_{t+1}
  =
  \alpha\,\boldsymbol{M}'_t.\textsc{detach}()
  +
  (1-\alpha)\,\boldsymbol{M}_t.\textsc{detach}().
\]
Thus, EMA tests whether a learning-free recurrent write is sufficient
when the model can still read a persistent state at inference time.

\begin{algorithm}[ht!]
\caption{EMA training loop for \muvla.
Unlike TBPTT, the EMA variant does not backpropagate through the
cross-step memory update. At each step, the model produces a candidate
memory $\boldsymbol{M}'$, and the next memory is a detached
exponential moving average of $\boldsymbol{M}'$ and the previous memory
$\boldsymbol{M}$.}
\label{alg:ema}
\begin{algorithmic}[1]
\Require Episodic dataloader $\mathcal{D}$ with $B$ streams;
model $f_{\theta}$; learnable initial memory
$\boldsymbol{M}^{\mathrm{init}}$; EMA factor $\alpha$; optimizer
\State $\boldsymbol{M} \leftarrow \boldsymbol{M}^{\mathrm{init}}$
replicated across the batch dimension
\For{each batch $\mathcal{B}=(o,\ell,a,\texttt{is\_first})$ from $\mathcal{D}$}
  \State $\boldsymbol{M} \leftarrow
  \textsc{ApplyReset}
  \!\left(
    \boldsymbol{M},
    \texttt{is\_first},
    \boldsymbol{M}^{\mathrm{init}}
  \right)$
  \Comment{Reset only streams starting a new episode}
  \State $\hat{a},\,\boldsymbol{M}' \leftarrow
  f_{\theta}(o,\ell,\boldsymbol{M})$
  \Comment{Forward pass with current recurrent state}
  \State $\mathcal{L} \leftarrow \ell_{\mathrm{chunk}}(\hat{a},a)$
  \State $\mathcal{L}.\textsc{backward}()$
  \Comment{Backpropagate only through the current step}
  \State $\textsc{optimizer.step}()$;\quad
  $\textsc{optimizer.zero\_grad}()$
  \State $\boldsymbol{M} \leftarrow
  \alpha\,\boldsymbol{M}'.\textsc{detach}()
  +(1-\alpha)\,\boldsymbol{M}.\textsc{detach}()$
  \Comment{Detached EMA memory update}
\EndFor
\end{algorithmic}
\end{algorithm}

\section{Evaluation Protocol}
\label{app:eval}

On \mikasa, each method is evaluated with $100$ deterministic episodes
per environment. Evaluation uses a fixed set of seeds that is held out
from training and identical across methods, so observed gaps reflect
policy differences rather than seed variation. The primary metric is
\texttt{success\_once}. On LIBERO, each suite consists of $10$
subtasks, and each subtask is evaluated with $50$ episodes; we report
the suite's standard success rate. All \muvla evaluations use
receding-horizon inference (Section~\ref{sec:method}); hyperparameters
and compute are listed in Appendices~\ref{app:hparams}
and~\ref{app:cost}.

\section{Training Hyperparameters}
\label{app:hparams}

\begin{table}[ht!]
\caption{\textbf{Training hyperparameters for \muvla.}}
\label{tab:hparams}
\centering
\small
\begin{tabular}{ll}
\toprule
\textbf{Hyperparameter} & \textbf{Value} \\
\midrule
Base model               & OpenVLA-7B (Llama-2-7B + SigLIP ViT) \\
LoRA rank                & 32 \\
LoRA $\alpha$            & 32 \\
Optimizer                & AdamW \\
Learning rate            & $5 \times 10^{-4}$ \\
LR schedule              & Cosine decay \\
Warmup steps             & 2000 \\
Minimum LR ratio         & 0.1 \\
Batch size per GPU       & 4 \\
Number of GPUs           & 8 (NVIDIA H100 80\,GB, DDP) \\
Max training steps       & 150{,}000 \\
TBPTT length $K$         & 8 \\
Memory tokens $m$        & 64 \\
Hidden dimension $d$     & 4096 \\
Action chunk size $H$    & 8 \\
Image augmentation       & Random crop, color jitter \\
Input resolution         & $224 \times 224$ \\
Number of camera views   & 2 (top-down + wrist) \\
\bottomrule
\end{tabular}
\end{table}

\section{Computational Cost}
\label{app:cost}

We report measured inference and training costs for the \muvla family
and the \openvlaoft references on the configurations used in the main
experiments (LoRA $r{=}32$, $H{=}8$, batch size $4$ per GPU,
$8{\times}$NVIDIA H100 80\,GB, DDP, and up to $150{,}000$ training
steps). Two factors explain most of the cost difference. First, all
\muvla evaluations use receding-horizon inference, i.e., one model call
per environment step, whereas \openvlaoft executes an $H{=}8$ open-loop
action chunk per call. As a result, \muvla has nearly the same per-call
forward latency as \openvlaoft, but lower closed-loop throughput.
Second, during training, the dominant cost is the TBPTT length $K$:
larger $K$ keeps a longer recurrent computation graph in memory and
therefore increases both wall-clock time and peak GPU memory.

\begin{table}[ht!]
\caption{
  \textbf{Evaluation-time cost on \texttt{RememberColor5-VLA-v0}.}
  Forward latency is measured per model call; simulation-step latency
  includes environment stepping; throughput is measured in closed-loop
  environment steps per second; peak GPU memory is total memory used by
  the evaluation process. \openvlaoft uses chunked inference with
  $H{=}8$, while \muvla uses receding-horizon inference.
}
\label{tab:cost-eval}
\centering
\begin{adjustbox}{width=\textwidth}
\begin{tabular}{lcccc}
\toprule
\textbf{Method}
  & \textbf{Fwd latency (ms)}
  & \textbf{Sim-step (ms)}
  & \textbf{Throughput (Hz)}
  & \textbf{Peak GPU (MiB)} \\
\midrule
\openvlaoft (chunked, $H{=}8$)
  & $61.33$ & $47.93$ & $20.86$ & $15{,}358.70$ \\
\muvla \mbox{exp-id$=$4} ($m{=}64$, $K{=}8$)
  & $67.16$ & $109.44$ & $\phantom{0}9.14$ & $15{,}434.04$ \\
\muvla \mbox{exp-id$=$6} ($m{=}64$, EMA)
  & $68.27$ & $109.95$ & $\phantom{0}9.09$ & $15{,}434.54$ \\
\muvla \mbox{exp-id$=$12} ($m{=}64$, $K{=}2$)
  & $66.65$ & $107.55$ & $\phantom{0}9.30$ & $15{,}434.04$ \\
\bottomrule
\end{tabular}
\end{adjustbox}
\end{table}

\begin{table}[ht!]
\caption{
  \textbf{Training-time cost across all reported runs.}
  Forward latency is the mean per training step, peak memory is the
  per-GPU maximum during training, and total training time is wall-clock
  time on $8{\times}$H100 80\,GB. The main recurrent cost is controlled
  by the TBPTT length $K$; EMA has a cost profile close to $K{=}1$.
}
\label{tab:cost-train}
\centering
\setlength{\tabcolsep}{4pt}
\begin{adjustbox}{width=\textwidth}
\begin{tabular}{lrrr}
\toprule
\textbf{Configuration}
  & \textbf{Fwd (ms)}
  & \textbf{Peak mem (GiB)}
  & \textbf{Train time} \\
\midrule
\openvlaoft, \mikasa, original setup              & $17.94$ & $37.21$ & $20$h$38$m \\
\openvlaoft, \mikasa, episodic                    & $17.66$ & $37.18$ & $22$h$41$m \\
\openvlaoft, \mikasa, episodic $+$ 1st frame      & $25.32$ & $46.17$ & $1$d$10$h$25$m \\
\midrule
\muvla, \mikasa, $m{=}1$, $K{=}8$                 & $22.74$ & $54.74$ & $10$d$8$h$35$m \\
\muvla, \mikasa, $m{=}64$, $K{=}8$                & $22.32$ & $55.63$ & $12$d$0$h$06$m \\
\muvla, \mikasa, $m{=}64$, $K{=}1$                & $19.24$ & $38.93$ & $1$d$3$h$35$m \\
\muvla, \mikasa, $m{=}64$, EMA                    & $19.28$ & $38.92$ & $1$d$2$h$35$m \\
\muvla, \mikasa, $m{=}64$, EMA, full-mask         & $19.13$ & $38.92$ & $1$d$3$h$09$m \\
\muvla, \mikasa, $m{=}64$, $K{=}8$, single-task RC5 & $22.74$ & $55.63$ & $10$d$18$h$58$m \\
\muvla, \mikasa, $m{=}64$, $K{=}2$                & $20.71$ & $27.44$ & $2$d$18$h$21$m \\
\midrule
\openvlaoft, LIBERO                               & $18.83$ & $36.92$ & $1$d$0$h$49$m \\
\muvla, LIBERO, $m{=}64$, $K{=}8$                 & $22.73$ & $55.62$ & $10$d$19$h$12$m \\
\muvla, LIBERO, $m{=}64$, EMA                     & $19.96$ & $38.73$ & $1$d$2$h$26$m \\
\bottomrule
\end{tabular}
\end{adjustbox}
\end{table}

Tables~\ref{tab:cost-eval} and~\ref{tab:cost-train} show that the cost
of recurrence is predictable and mostly separated from the cost of the
memory tokens themselves. At evaluation time, all \muvla variants have
similar forward latency, around $67$--$68$ ms per model call, only
about $5$--$7$ ms above \openvlaoft. Peak GPU memory is also essentially
unchanged at $m{=}64$. The lower throughput of \muvla comes primarily
from receding-horizon inference: the policy is queried every
environment step, whereas \openvlaoft queries once per $H{=}8$ executed
steps.

At training time, the main scaling factor is the TBPTT length $K$.
The $K{=}1$ and EMA variants train at near-baseline cost, taking about
one day and using about $39$ GiB per GPU. The $K{=}8$ variants require
substantially longer wall-clock time because they retain an eight-step
recurrent graph before each backward pass. The most important operating
point in our experiments is $K{=}2$: it trains in $2$d$18$h on
$8{\times}$H100 and does not require substantially more GPU memory than
the non-recurrent reference runs. Despite this moderate additional
training cost, it achieves the strongest MIKASA-Robo results, including
$0.84$ average SR on the training tasks and the best held-out
matched-semantics performance among recurrent variants.

\section{Existing Assets and Licenses}
\label{app:license}

Our experiments use publicly released research assets. We initialize from \openvla/\openvlaoft checkpoints and code released under the MIT License. The backbone includes DINOv2, SigLIP/Big Vision, and Llama-2 components, used under their upstream terms: Apache-2.0 for DINOv2 and SigLIP/Big Vision, and the Llama 2 Community License and Acceptable Use Policy for Llama-2. OpenVLA was pretrained on Open X-Embodiment, whose released software is Apache-2.0 and other materials are CC-BY 4.0; we do not redistribute its data. For evaluation and fine-tuning, we use MIKASA-Robo and LIBERO, both released under the MIT License. MIKASA-Robo builds on ManiSkill, whose environments use permissive licenses such as Apache-2.0 and whose assets are CC BY-NC 4.0. Any released code, checkpoints, or videos will preserve the required copyright, license, and attribution notices.

\section{Extended Diagnostic Results}
\label{app:diagnostics}

\begin{figure}[ht!]
\centering
\begin{minipage}{0.48\textwidth}\centering
  \includegraphics[width=\linewidth]{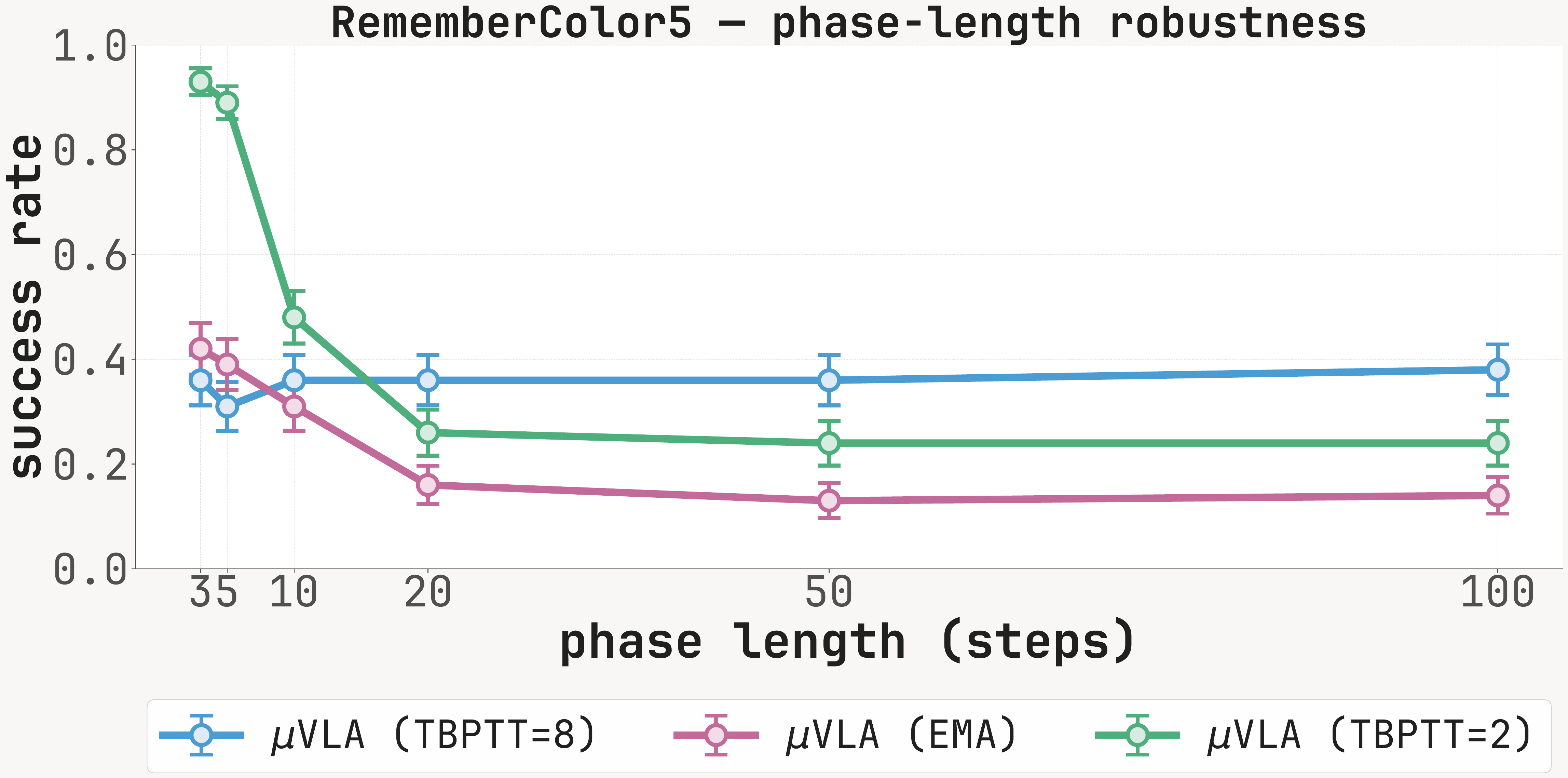}\\[-2pt]
  \footnotesize (a) Phase-length sweep
\end{minipage}\hfill
\begin{minipage}{0.48\textwidth}\centering
  \includegraphics[width=\linewidth]{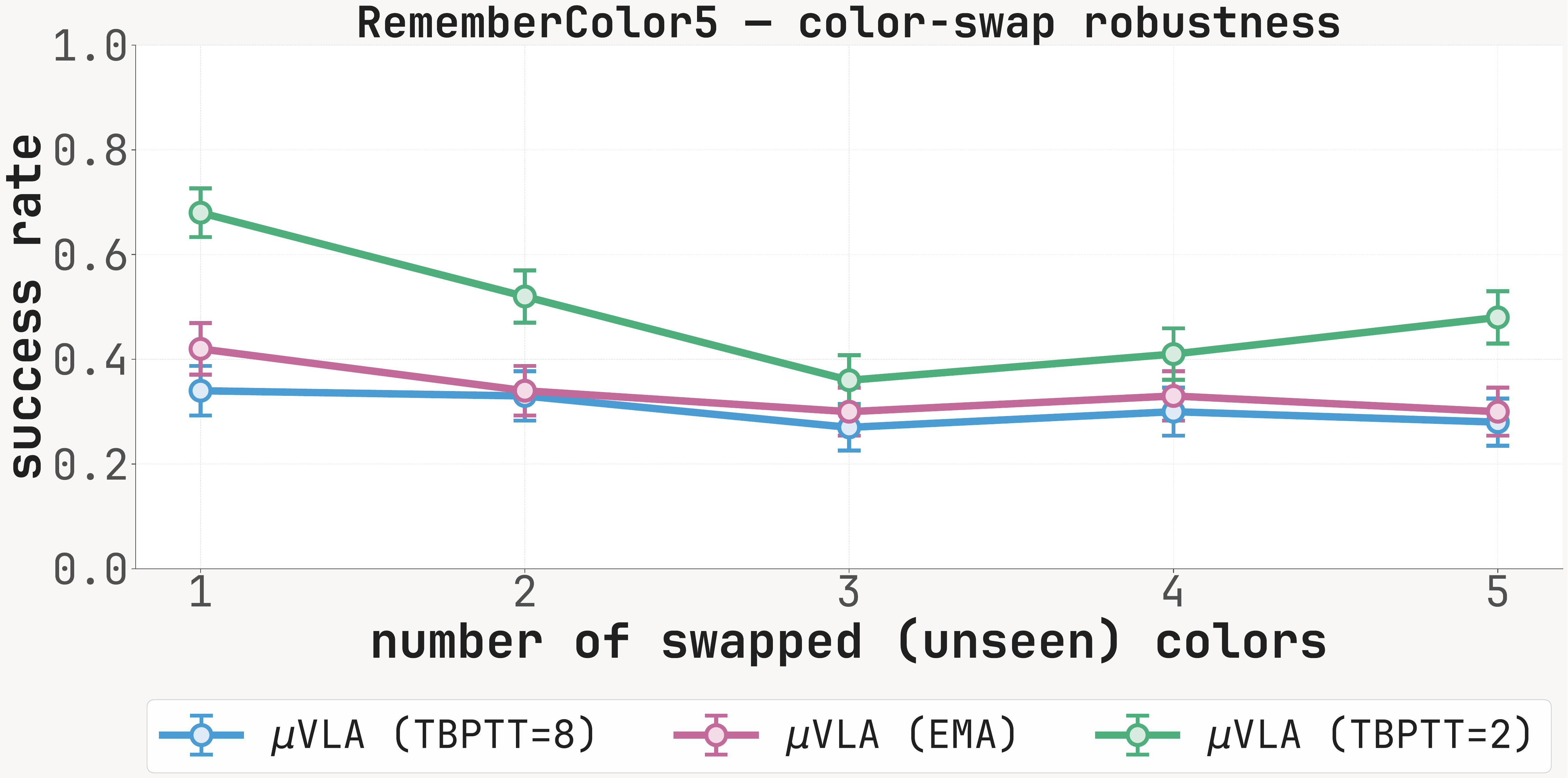}\\[-2pt]
  \footnotesize (b) Color-swap sweep
\end{minipage}
\caption{\textbf{Generalization probes on \texttt{RememberColor5}.} (a) Both task phases fixed to $N\!\in\!\{3,5,10,20,50,100\}$ steps; training covers $N\!\in\!\{1,\dots,5\}$ per phase. (b) $1$--$5$ in-distribution colors replaced with the OOD palette $\{$\texttt{Pink}, \texttt{Orange}, \texttt{Purple}, \texttt{Brown}, \texttt{White}$\}$. $K{=}2$ dominates in-distribution and remains most robust to color swap, but degrades fastest with phase length; $K{=}8$ is flat-but-low; EMA collapses with phase length.}
\label{fig:rc5_generalization}
\end{figure}

This appendix collects the per-environment diagnostic results referenced in Section~\ref{sec:exp:diagnostics}: memory representation dynamics (Appendix~\ref{app:mem-dynamics}), attention rollouts (Appendix~\ref{app:attn-rollouts}), causal memory interventions (Appendix~\ref{app:intervention-full}), chunk-length sweeps (Appendix), and inference and training cost (Appendix~\ref{app:cost}). Figure~\ref{fig:rc5_generalization} aggregates the phase-length and color-swap generalization curves for \texttt{RememberColor5}. Table~\ref{tab:mikasa_robo_comparison} contextualizes \muvla\ against representative memory-augmented VLAs on the four \mikasa\ environments where their published numbers overlap.

\begin{table}[ht!]
\caption{
  \textbf{Memoryless vs.\ memory-augmented VLAs on selected \mikasa tasks.}
  Success rates are reported per task, CronusVLA and MemoryVLA results
  are from~\citep{memoryvla}. Memory-augmented models are shaded.
}
\label{tab:mikasa_robo_comparison}
\centering
\scriptsize
\setlength{\tabcolsep}{4pt}
\renewcommand{\arraystretch}{0.95}
\begin{tabular}{lccccc}
\toprule
\textbf{Model}
& \textbf{\texttt{InterceptMedium}}
& \textbf{\texttt{RememberColor3}}
& \textbf{\texttt{RememberColor5}}
& \textbf{\texttt{RememberColor9}}
& \textbf{Avg.} \\
\midrule
SpatialVLA~\citep{qu2025spatialvla}
& 0.27 & 0.27 & 0.17 & 0.11 & 0.21 \\
\openvlaoft~\citep{openvlaoft}
& 0.14 & 0.59 & 0.16 & 0.06 & 0.24 \\
$\pi_0$~\citep{pi0}
& 0.42 & 0.35 & 0.22 & 0.15 & 0.29 \\
\midrule
\rowcolor{linkColor!15}
CronusVLA~\citep{li2025cronusvla}
& 0.05 & 0.31 & 0.13 & 0.09 & 0.15 \\
\rowcolor{linkColor!15}
MemoryVLA~\citep{memoryvla}
& 0.24 & 0.44 & 0.30 & 0.20 & 0.30 \\
\rowcolor{linkColor!15}
\textbf{\muvla}
& \textbf{0.56} & \textbf{0.92} & \textbf{0.93} & \textbf{0.41} & \textbf{0.71} \\
\bottomrule
\end{tabular}
\end{table}

\section{Per-Environment Memory Dynamics}
\label{app:mem-dynamics}

\begin{figure*}[t]
\centering

\begin{minipage}{0.98\textwidth}
\centering
  \includegraphics[width=\linewidth]{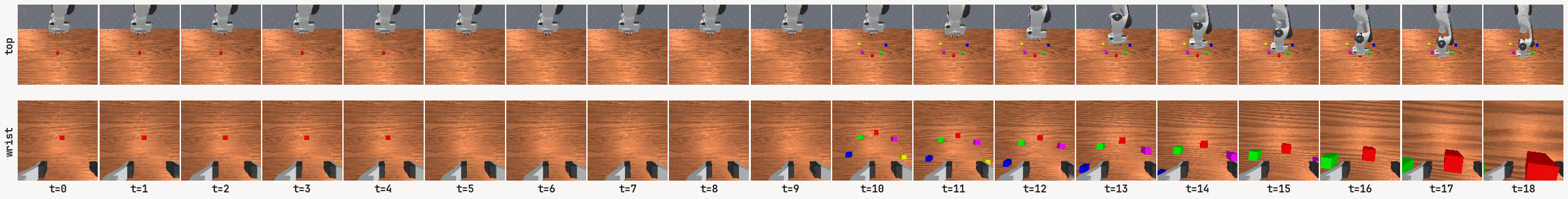}\\[-2pt]
  \footnotesize (a) Top + wrist rollout (19 steps, success).
\end{minipage}

\vspace{0.4em}

\begin{minipage}[t]{0.48\textwidth}
\vspace{0pt}
\centering
  \includegraphics[width=\linewidth]{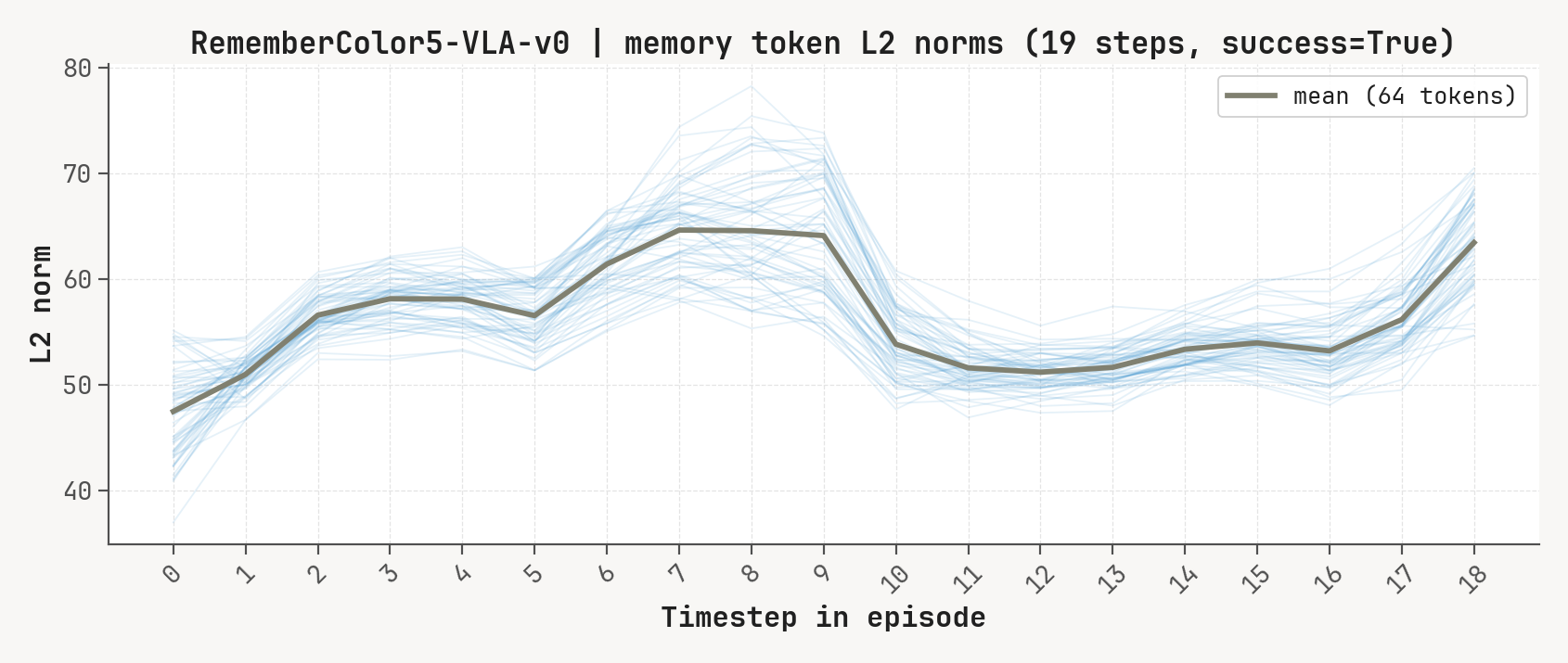}\\[-2pt]
  \footnotesize (b) Per-token L2 norms.

  \vspace{0.7em}

  \includegraphics[width=\linewidth]{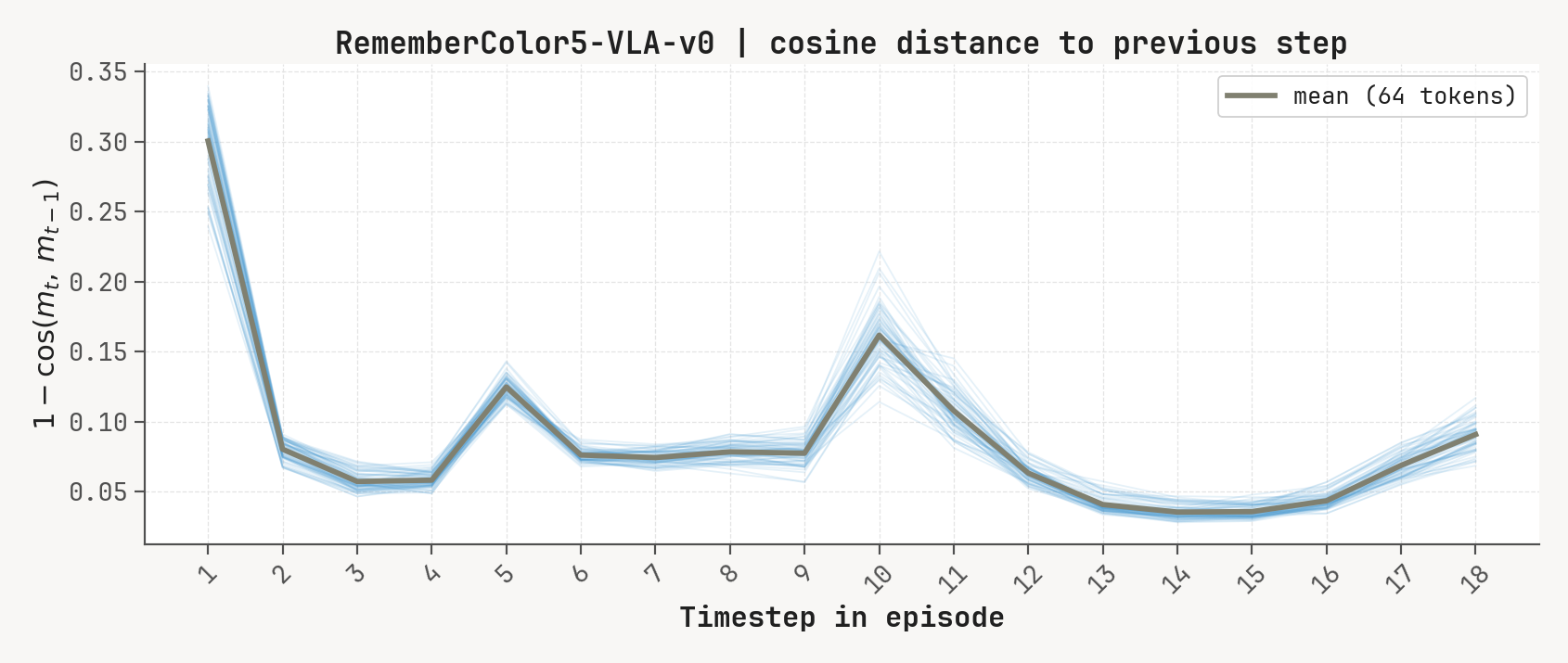}\\[-2pt]
  \footnotesize (c) $1{-}\cos(\boldsymbol{M}_t, \boldsymbol{M}_{t-1})$.
\end{minipage}
\hfill
\begin{minipage}[t]{0.48\textwidth}
\vspace{0pt}
\centering
  \includegraphics[width=0.86\linewidth]{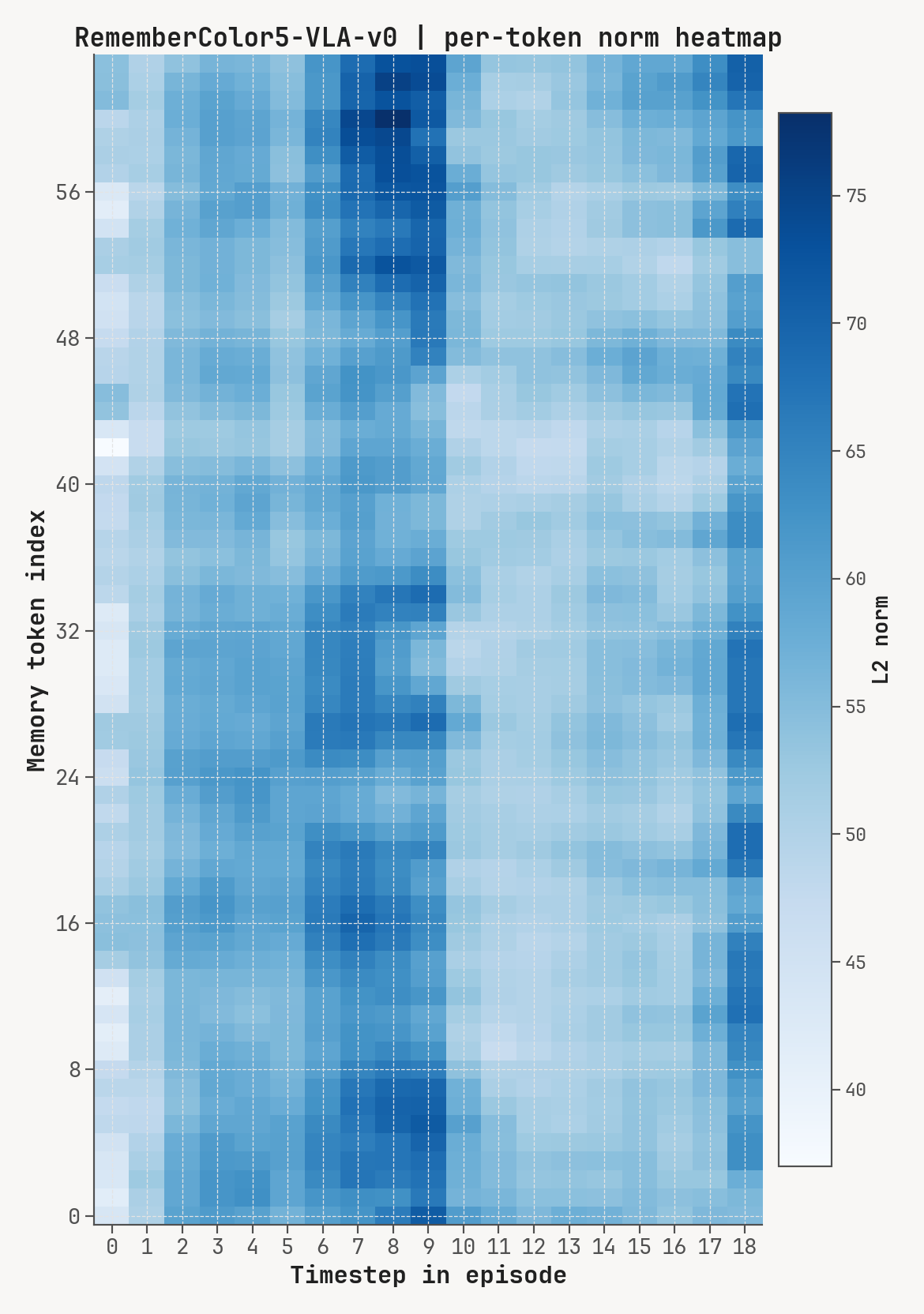}\\[-2pt]
  \footnotesize (d) Per-token norm heatmap.
\end{minipage}

\caption{\textbf{Memory dynamics on \texttt{RememberColor5}
  ($K{=}2$, $m{=}64$).}
  Companion to Figure~\ref{fig:cosine_dynamics}. The cue-window
  plateau at \pht{phGreen}{5}--\textcolor{phBlue}{$\boldsymbol{9}$}
  appears in the per-token L2 norms and
  heatmap as a sub-band of memory tokens with elevated norm, while
  the cosine peaks mark phase transitions in the episode.}
\label{fig:mem_dynamics_rc5_full}
\vspace{-1em}
\end{figure*}

\begin{figure*}[t]
\centering

\begin{minipage}{0.98\textwidth}
\centering
  \includegraphics[width=\linewidth]{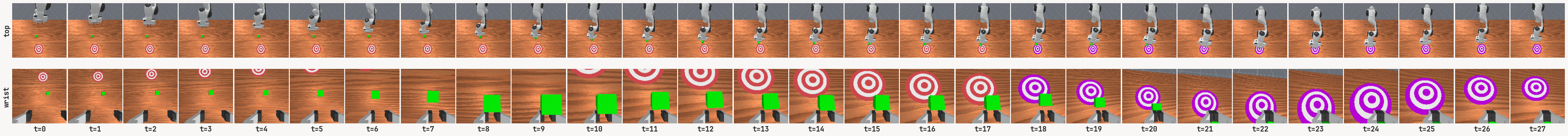}\\[-2pt]
  \footnotesize (a) Top + wrist rollout (28 steps, success).
\end{minipage}

\vspace{0.4em}

\begin{minipage}[t]{0.48\textwidth}
\vspace{0pt}
\centering
  \includegraphics[width=\linewidth]{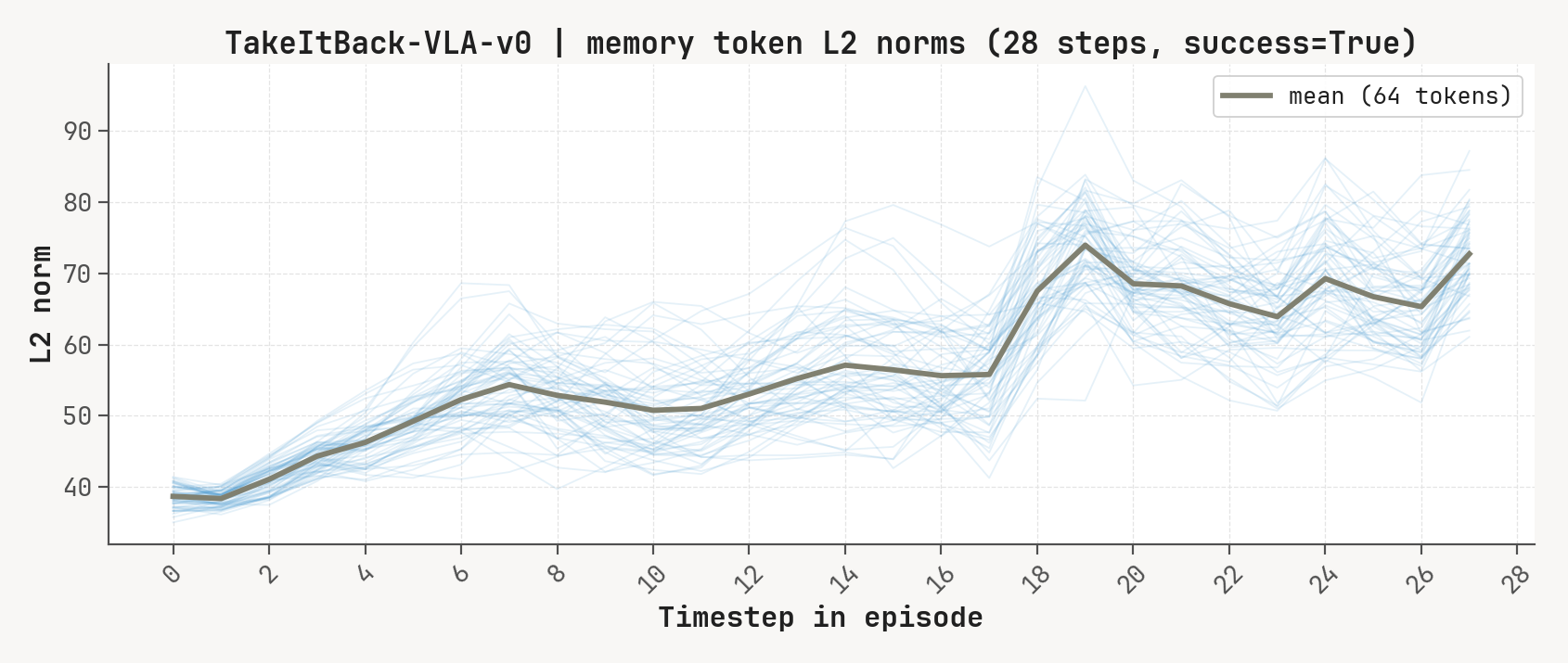}\\[-2pt]
  \footnotesize (b) Per-token L2 norms.

  \vspace{0.7em}

  \includegraphics[width=\linewidth]{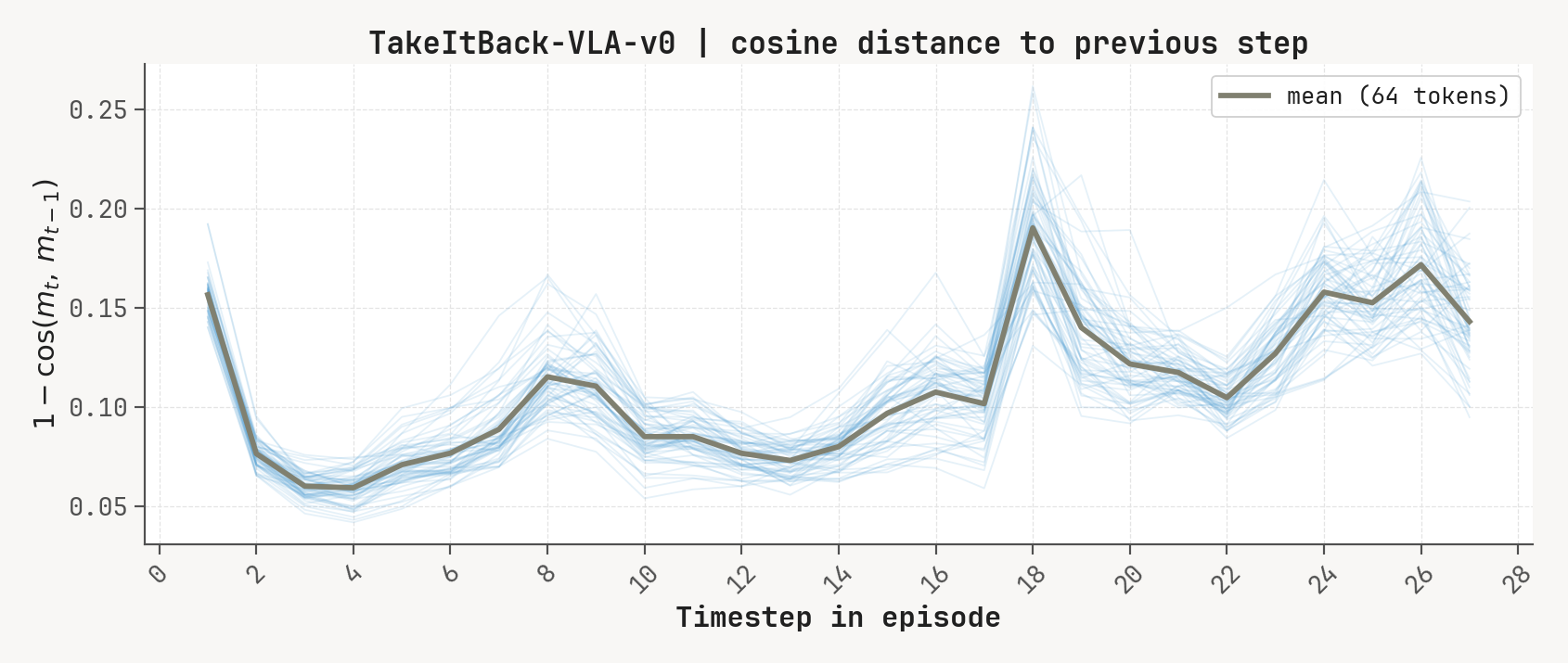}\\[-2pt]
  \footnotesize (c) $1{-}\cos(\boldsymbol{M}_t, \boldsymbol{M}_{t-1})$.
\end{minipage}
\hfill
\begin{minipage}[t]{0.48\textwidth}
\vspace{0pt}
\centering
  \includegraphics[width=0.86\linewidth]{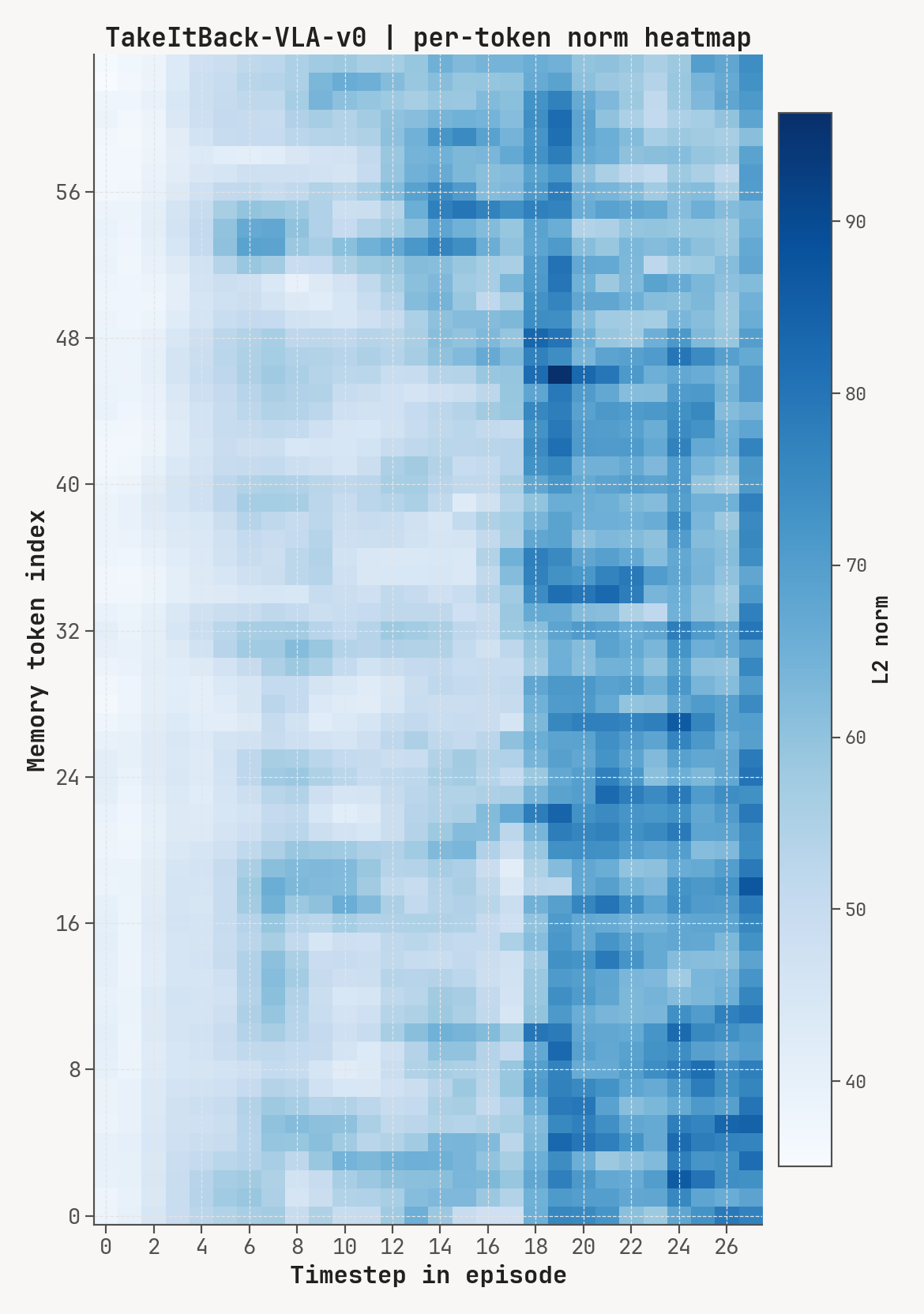}\\[-2pt]
  \footnotesize (d) Per-token norm heatmap.
\end{minipage}

\caption{\textbf{Memory dynamics on \texttt{TakeItBack}
  ($K{=}2$, $m{=}64$).}
  The cosine trace shows a two-event write pattern: the canonical
  $t{=}1$ spike ($\Delta\cos \!\approx\! 0.18$) and a second spike at
  $t{\approx}18$ when the policy switches to the return phase. The
  mean L2 norm rises from $\approx 40$ to $\approx 75$, with a sharp
  jump aligned with the same phase transition.}
\label{fig:mem_dynamics_takeitback}
\vspace{-1em}
\end{figure*}

\begin{figure*}[t]
\centering

\begin{minipage}{0.98\textwidth}
\centering
  \includegraphics[width=\linewidth]{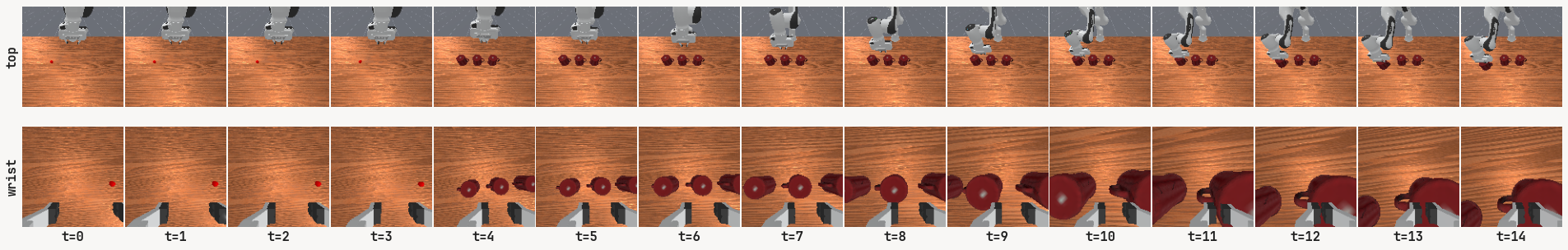}\\[-2pt]
  \footnotesize (a) Top + wrist rollout (15 steps, success).
\end{minipage}

\vspace{0.4em}

\begin{minipage}[t]{0.48\textwidth}
\vspace{0pt}
\centering
  \includegraphics[width=\linewidth]{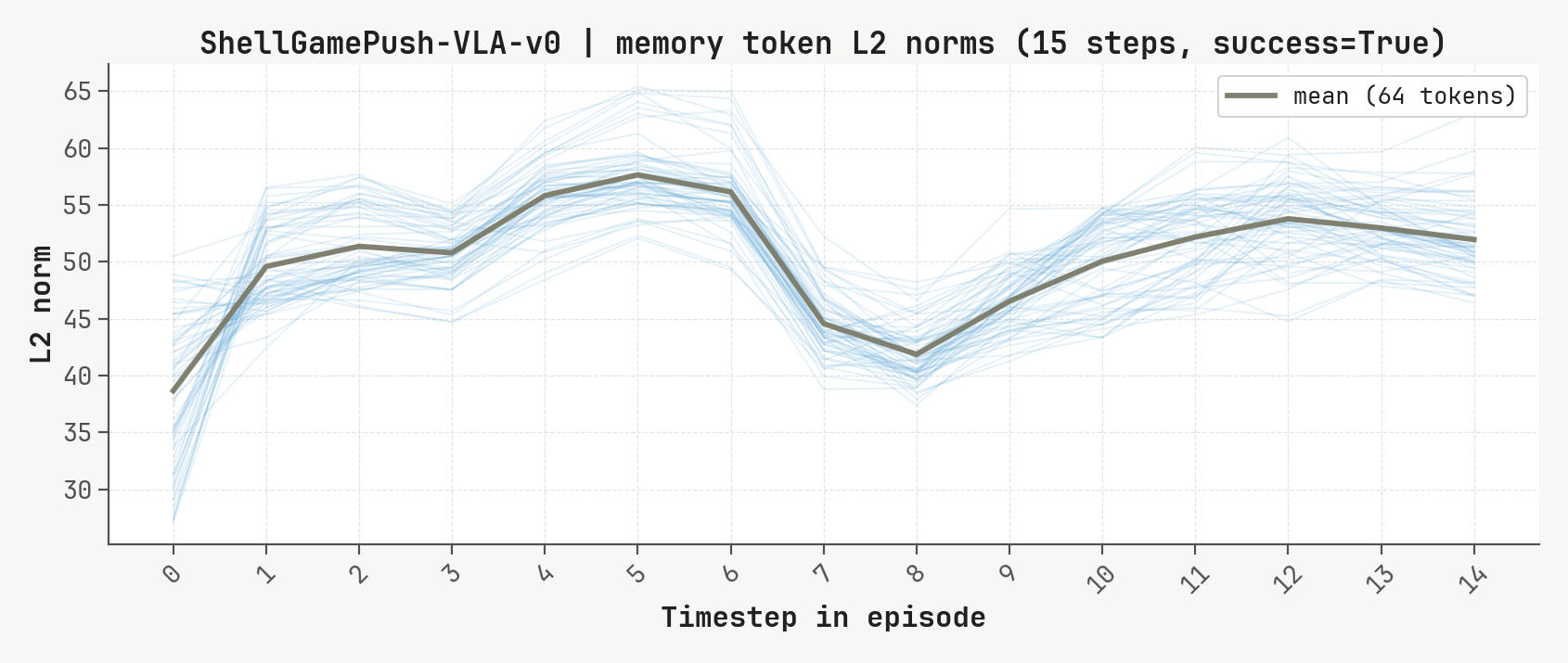}\\[-2pt]
  \footnotesize (b) Per-token L2 norms.

  \vspace{0.7em}

  \includegraphics[width=\linewidth]{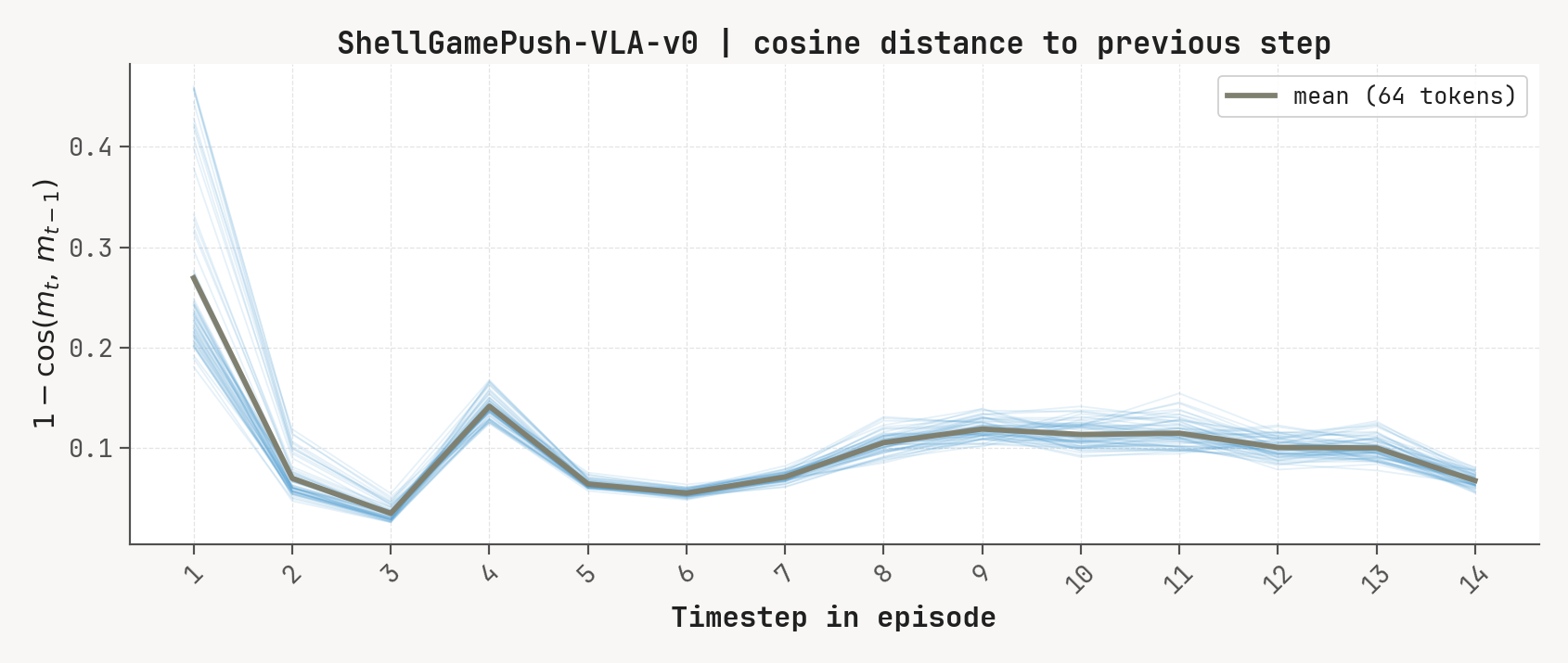}\\[-2pt]
  \footnotesize (c) $1{-}\cos(\boldsymbol{M}_t, \boldsymbol{M}_{t-1})$.
\end{minipage}
\hfill
\begin{minipage}[t]{0.48\textwidth}
\vspace{0pt}
\centering
  \includegraphics[width=0.86\linewidth]{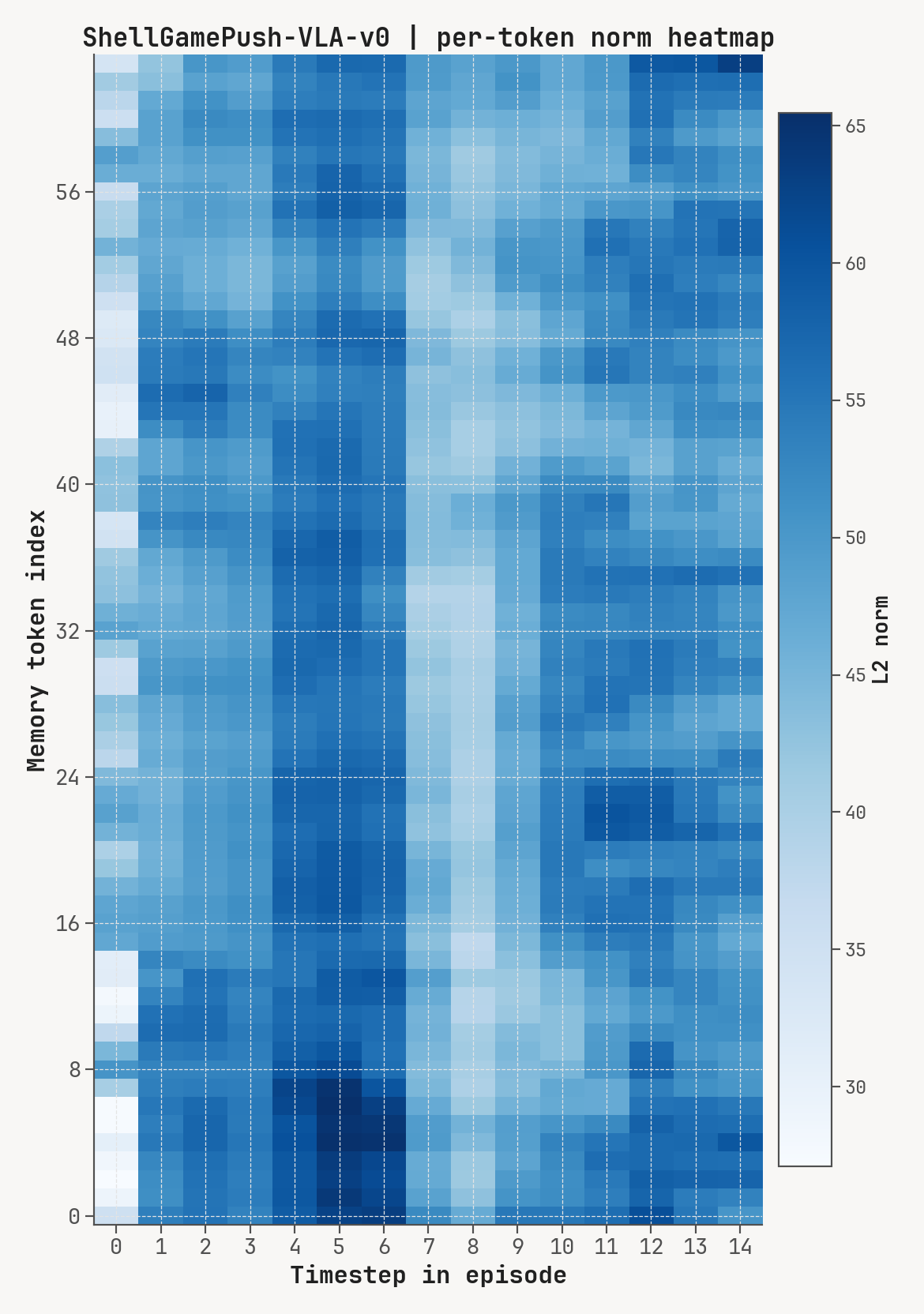}\\[-2pt]
  \footnotesize (d) Per-token norm heatmap.
\end{minipage}

\caption{\textbf{Memory dynamics on \texttt{ShellGamePush}
  ($K{=}2$, $m{=}64$).}
  A single early writing window: a large $t{=}1$ cosine spike
  ($\Delta\cos \!\approx\! 0.27$) followed by a small secondary bump
  at $t{=}4$ ($\approx 0.16$) and a quiet $\approx 0.10$ plateau
  thereafter. This is consistent with the cue being captured in the
  first frames and then carried largely unchanged through the push.}
\label{fig:mem_dynamics_shellgamepush}
\vspace{-1em}
\end{figure*}

\begin{figure*}[t]
\centering

\begin{minipage}{0.98\textwidth}
\centering
  \includegraphics[width=\linewidth]{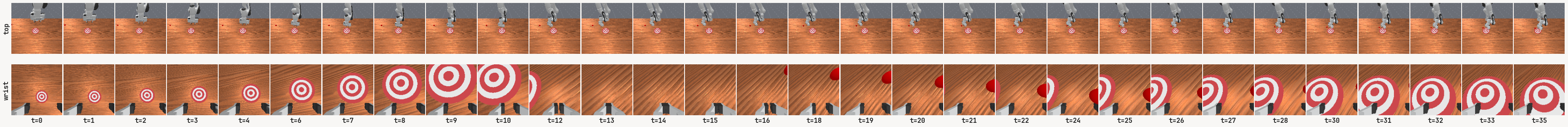}\\[-2pt]
  \footnotesize (a) Top + wrist rollout (36 steps, success).
\end{minipage}

\vspace{0.4em}

\begin{minipage}[t]{0.48\textwidth}
\vspace{0pt}
\centering
  \includegraphics[width=\linewidth]{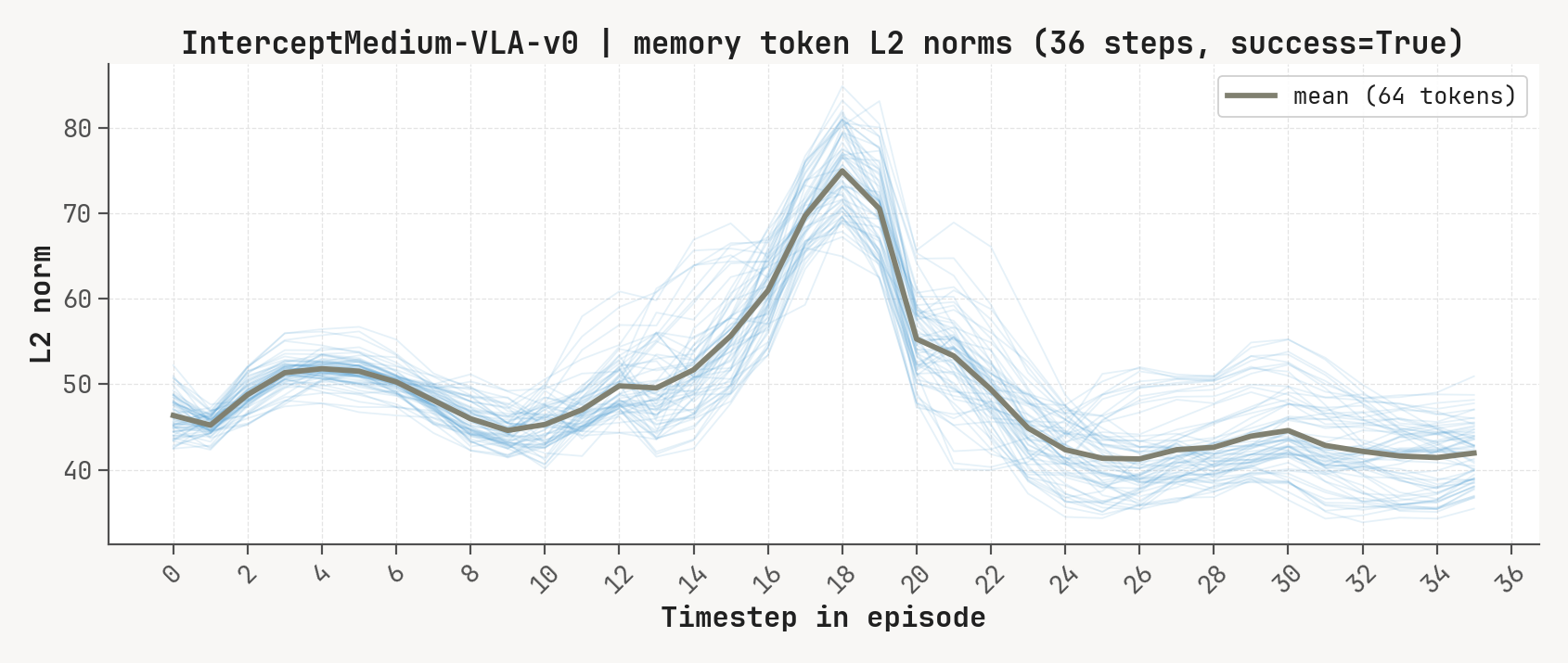}\\[-2pt]
  \footnotesize (b) Per-token L2 norms.

  \vspace{0.7em}

  \includegraphics[width=\linewidth]{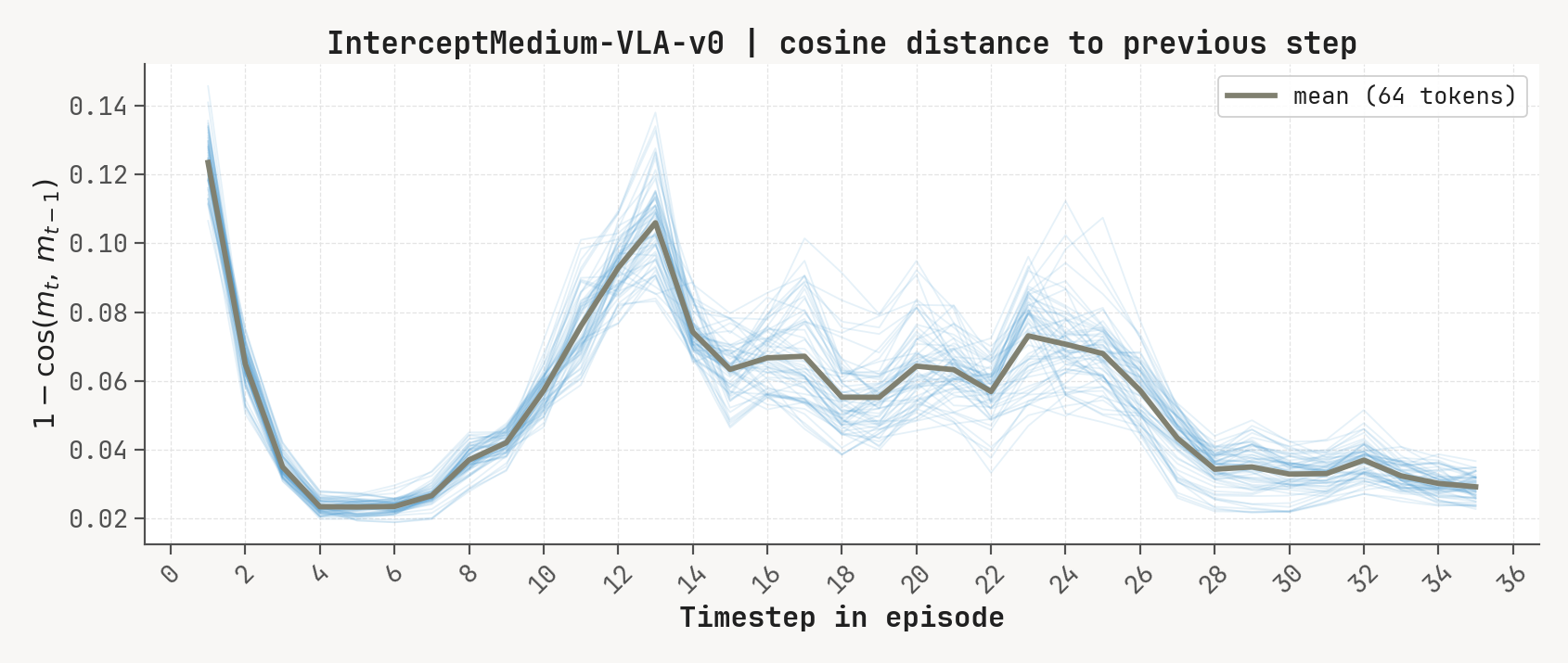}\\[-2pt]
  \footnotesize (c) $1{-}\cos(\boldsymbol{M}_t, \boldsymbol{M}_{t-1})$.
\end{minipage}
\hfill
\begin{minipage}[t]{0.48\textwidth}
\vspace{0pt}
\centering
  \includegraphics[width=0.86\linewidth]{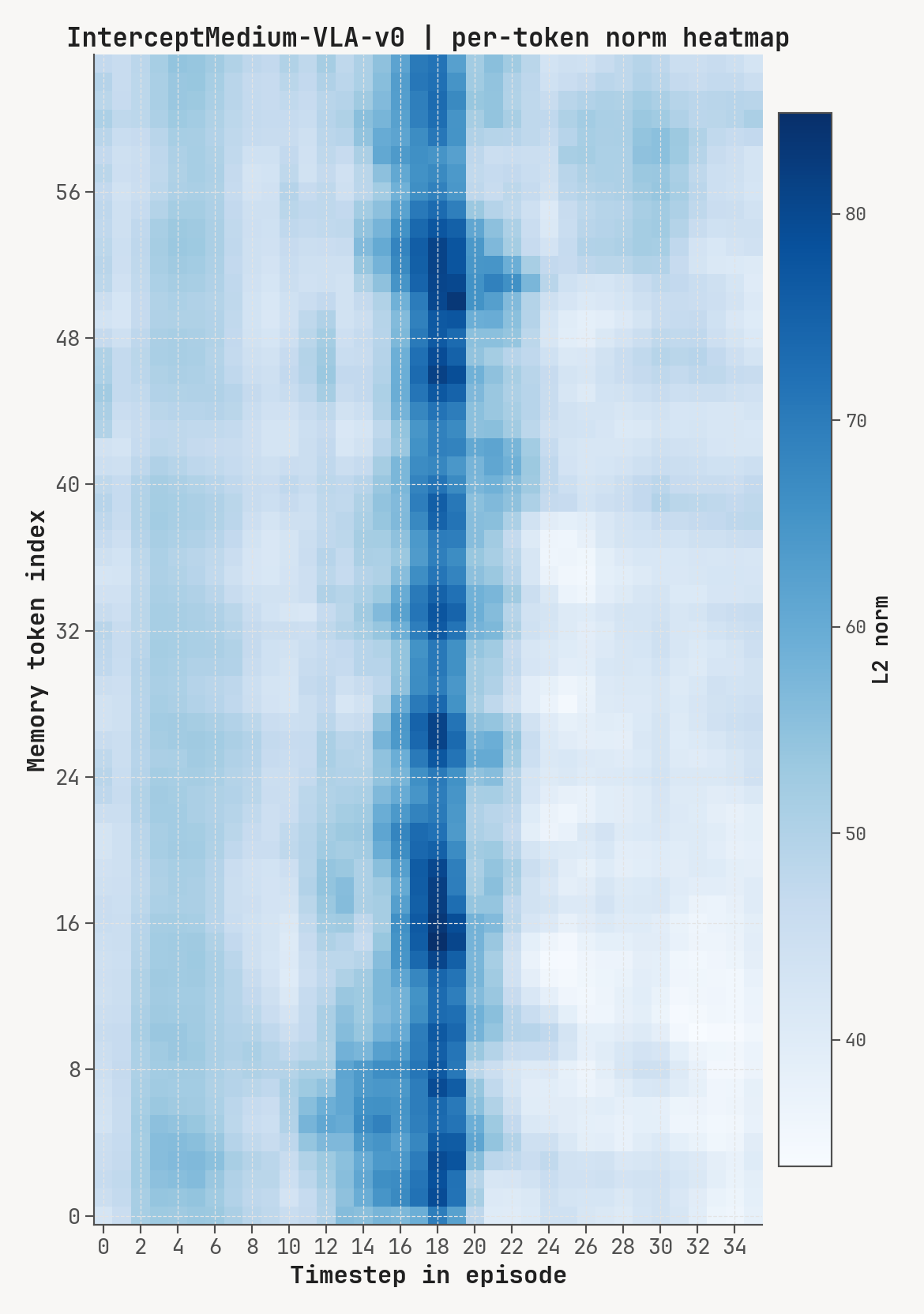}\\[-2pt]
  \footnotesize (d) Per-token norm heatmap.
\end{minipage}

\caption{\textbf{Memory dynamics on \texttt{InterceptMedium}
  ($K{=}2$, $m{=}64$).}
  Continuous re-grounding rather than a one-shot write: the
  $t{=}1$ spike ($\Delta\cos \!\approx\! 0.12$) is followed by a
  sustained mid-episode plateau at $\approx 0.06$--$0.08$. The mean
  L2 norm shows a pronounced bump at $t{=}15$--$18$ ($\approx 75$
  vs.\ $\approx 45$ baseline) aligned with the catch event.}
\label{fig:mem_dynamics_intercept}
\vspace{-1em}
\end{figure*}

\begin{figure*}[t]
\centering

\begin{minipage}{0.98\textwidth}
\centering
  \includegraphics[width=\linewidth]{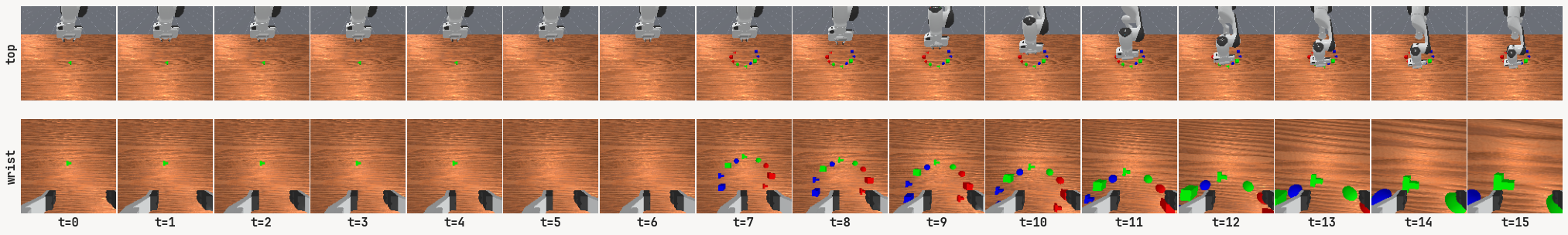}\\[-2pt]
  \footnotesize (a) Top + wrist rollout (16 steps, success).
\end{minipage}

\vspace{0.4em}

\begin{minipage}[t]{0.48\textwidth}
\vspace{0pt}
\centering
  \includegraphics[width=\linewidth]{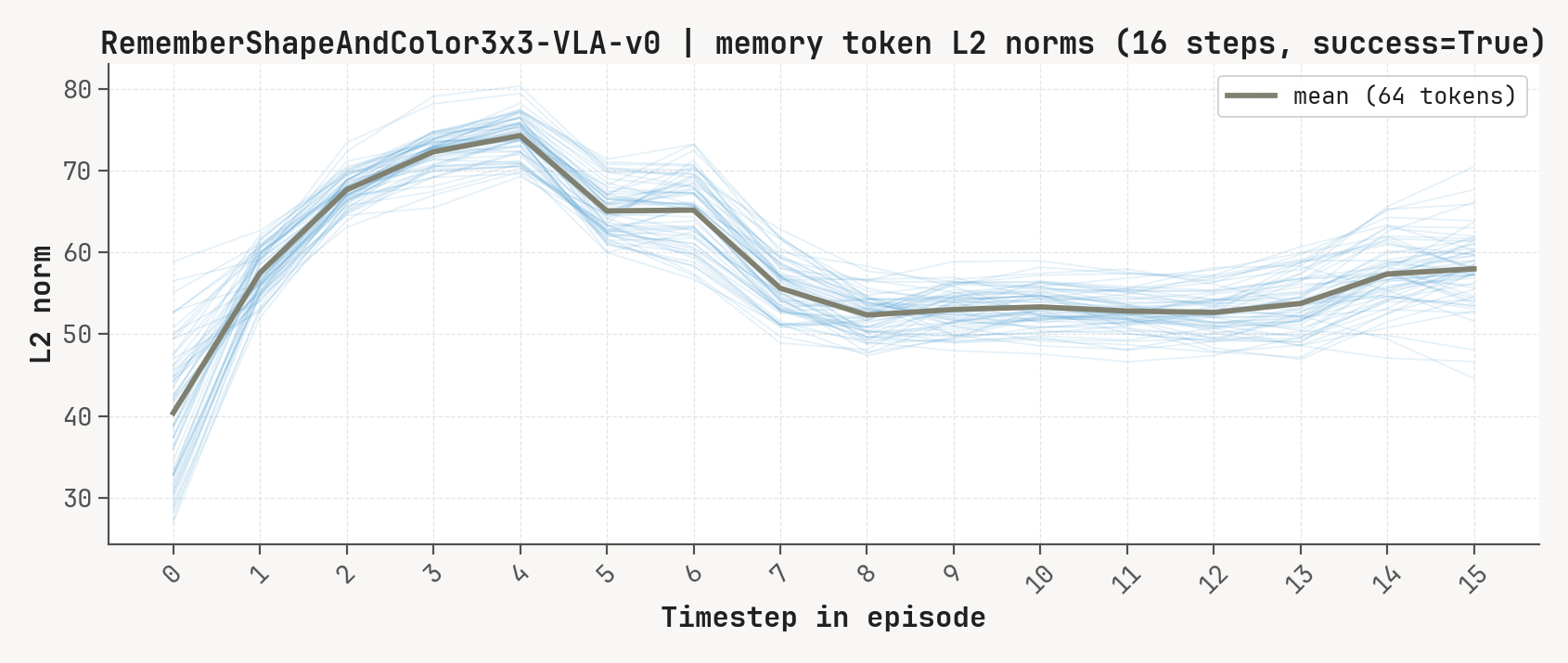}\\[-2pt]
  \footnotesize (b) Per-token L2 norms.

  \vspace{0.7em}

  \includegraphics[width=\linewidth]{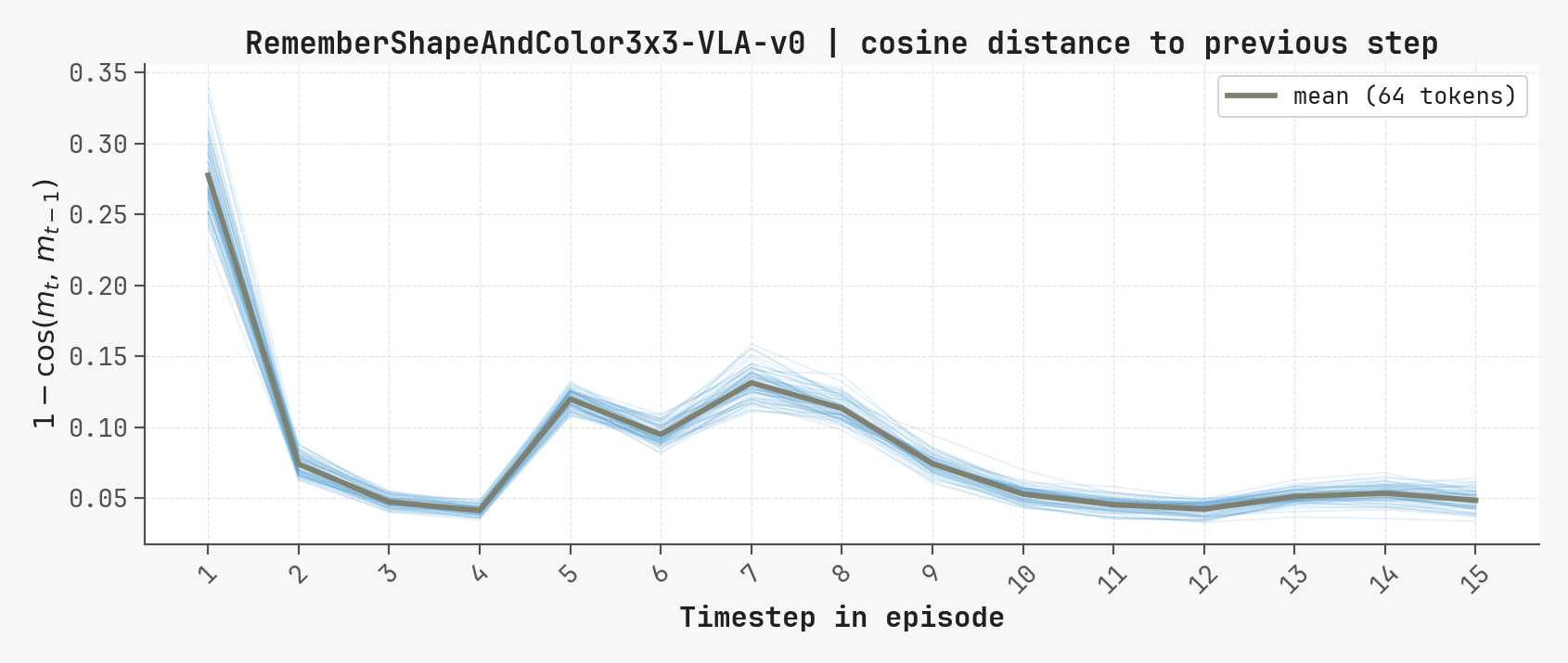}\\[-2pt]
  \footnotesize (c) $1{-}\cos(\boldsymbol{M}_t, \boldsymbol{M}_{t-1})$.
\end{minipage}
\hfill
\begin{minipage}[t]{0.48\textwidth}
\vspace{0pt}
\centering
  \includegraphics[width=0.86\linewidth]{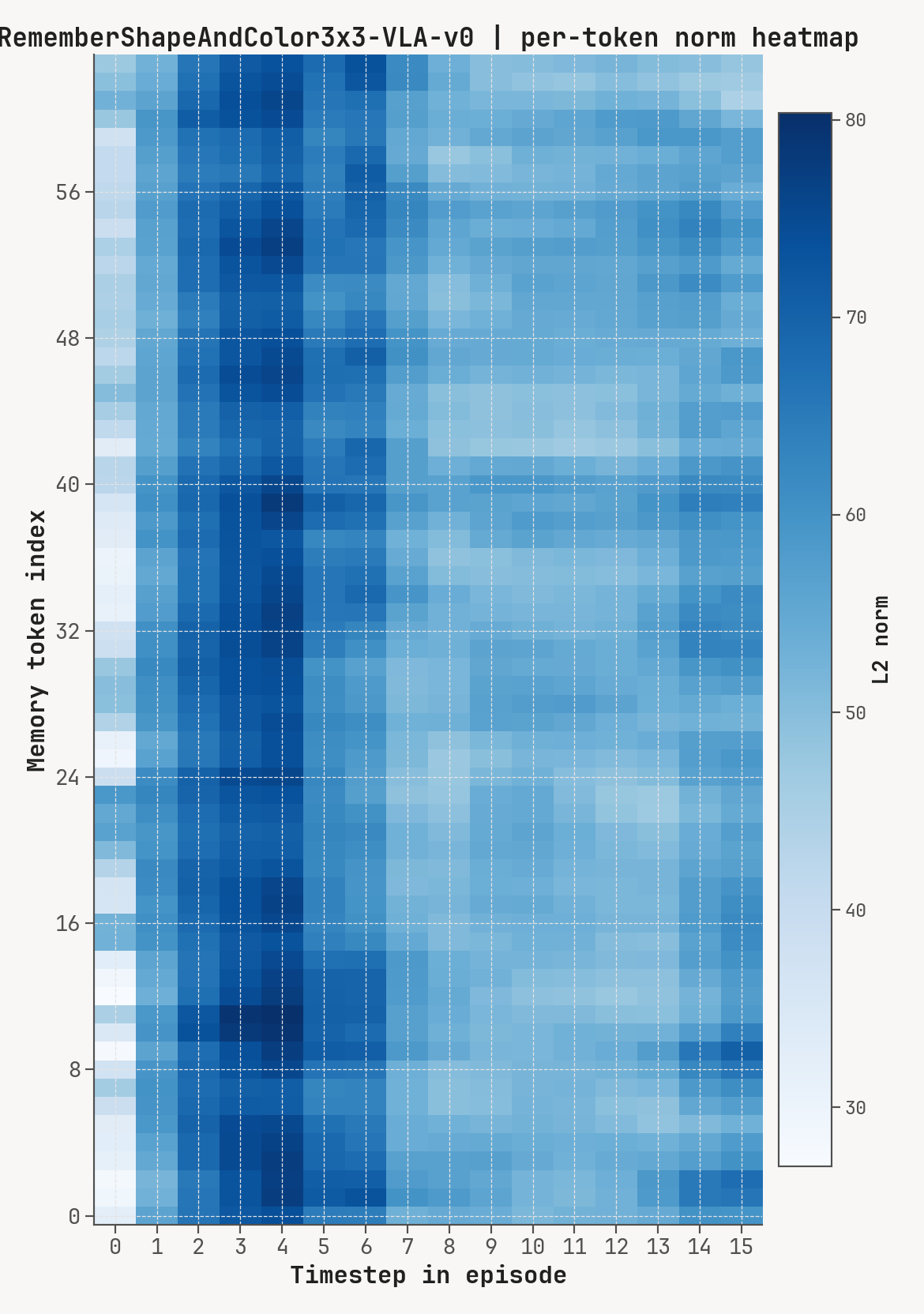}\\[-2pt]
  \footnotesize (d) Per-token norm heatmap.
\end{minipage}

\caption{\textbf{Memory dynamics on \texttt{RememberShapeAndColor3x3}
  ($K{=}2$, $m{=}64$).}
  Two-phase write: a $t{=}1$ spike ($\Delta\cos \!\approx\! 0.28$)
  for initial encoding, then a secondary bump at $t{=}5$--$7$
  ($\approx 0.13$) when the cue is fully revealed and committed.
  After $t{\approx}10$ the cosine settles to $\approx 0.05$ and the
  L2 norm plateaus at $\approx 60$.}
\label{fig:mem_dynamics_remsac}
\vspace{-1em}
\end{figure*}

Figures~\ref{fig:mem_dynamics_rc5_full}--\ref{fig:mem_dynamics_remsac}
collect the full four-panel cards (carrier $K{=}2$, $m{=}64$, one
successful episode per env) for all five training tasks of
\S\ref{sec:exp:diagnostics:01}. Each card has the same four panels:
top + wrist rollout strip, per-token L2 norm line plot,
cosine-distance-to-previous line plot, and per-token norm heatmap.
Pale blue lines are individual mem tokens, dark grey is the mean
over $64$ tokens. Headline interpretation per env is given in
\S\ref{sec:exp:diagnostics:01}.

\section{Per-Environment Attention Rollouts}
\label{app:attn-rollouts}

This appendix collects the attention-rollout overlays referenced in
\S\ref{sec:exp:diagnostics:01}. The memory$\to$vision panel for the
running diagnostic env \texttt{RememberColor5} is summarised in the
main text. The full three-query card for
\texttt{RememberColor5} and the four other training tasks is given
below. All overlays are
produced for \muvla\ at $K{=}2$, $m{=}64$ on a single successful
episode per env. At each step we compute attention rollout across
the $32$ backbone layers using the residual factor $0.5$, average
over heads, and read the row of
the resulting composite matrix at three query positions: the first
action token, a memory token (averaged over the $64$ memory tokens),
and the vision tokens themselves. The vision-token slice is split
into the two cameras (top and wrist) and reshaped to the
$16\times 16$ patch grid, then nearest-resampled to camera resolution
and rendered with the jet colormap (blue $=$ low, red $=$ high) over
the raw frame. Step counts and patch grid sizes are recorded in the
per-env \texttt{summary.json}.

\begin{figure}[t]
\centering
\begin{minipage}{0.98\linewidth}\centering
  \includegraphics[width=\linewidth]{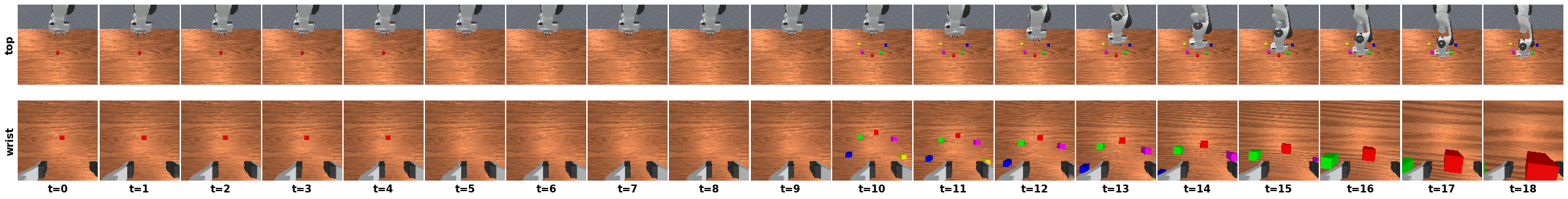}\\[-2pt]
  \footnotesize (a) Plain rollout (top + wrist, 19 steps, success).
\end{minipage}\\[0.3em]
\begin{minipage}{0.98\linewidth}\centering
  \includegraphics[width=\linewidth]{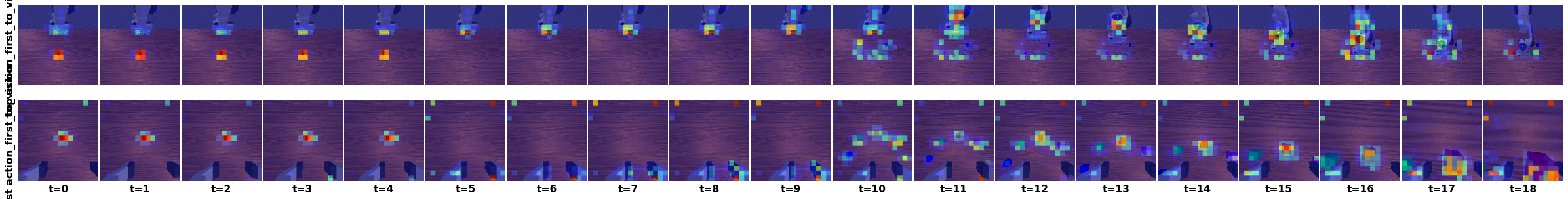}\\[-2pt]
  \footnotesize (b) action$\to$vision (first action token query).
\end{minipage}\\[0.3em]
\begin{minipage}{0.98\linewidth}\centering
  \includegraphics[width=\linewidth]{figures/mikasa_attn/RememberColor5-VLA-v0/memory_to_vision/rollout_attn.png}\\[-2pt]
  \footnotesize (c) memory$\to$vision.
\end{minipage}\\[0.3em]
\begin{minipage}{0.98\linewidth}\centering
  \includegraphics[width=\linewidth]{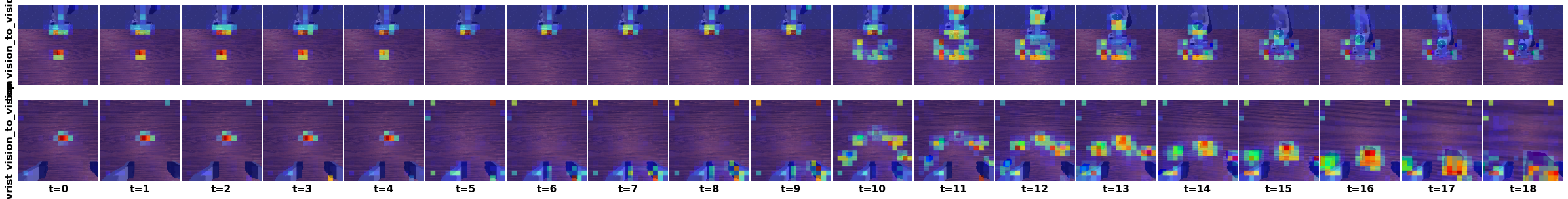}\\[-2pt]
  \footnotesize (d) vision$\to$vision.
\end{minipage}
\caption{\textbf{Attention rollouts on \texttt{RememberColor5}
  ($K{=}2$, $m{=}64$).} Per-step attention rollout over the $32$
  backbone layers, mean over heads, normalised per step. Row (a) is the raw episode (top
  and wrist cameras stacked). Rows (b)--(d) overlay the rolled-out
  attention from three query groups onto the same frames: the first
  action token, the memory tokens, and the vision tokens themselves.
  The action query tracks the gripper and the candidate object, the
  memory query latches onto the lamp patch in the cue frames and then
  diffuses, and the vision baseline highlights object-shaped clusters
  throughout. Companion to the cue-window write event in
  Figure~\ref{fig:cosine_dynamics}.}
\label{fig:attn_rollout_rc5}
\end{figure}

\begin{figure*}[t]
\centering
\begin{minipage}{0.98\textwidth}\centering
  \includegraphics[width=\linewidth]{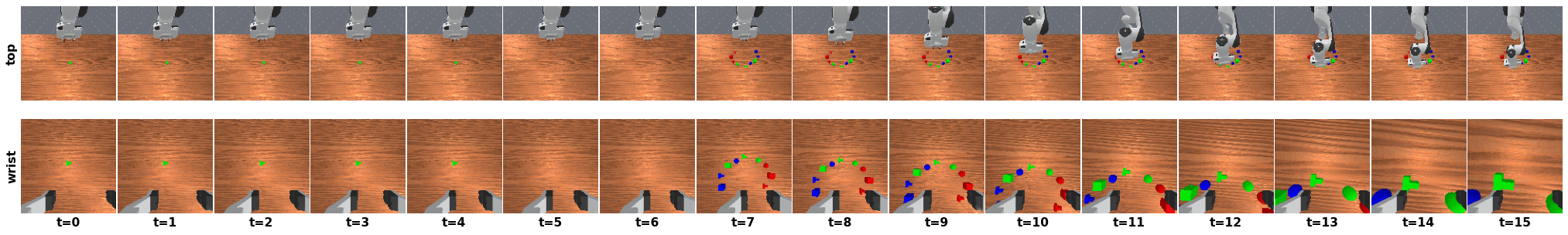}\\[-2pt]
  \footnotesize (a) Plain rollout (top + wrist, 16 steps, success).
\end{minipage}\\[0.3em]
\begin{minipage}{0.98\textwidth}\centering
  \includegraphics[width=\linewidth]{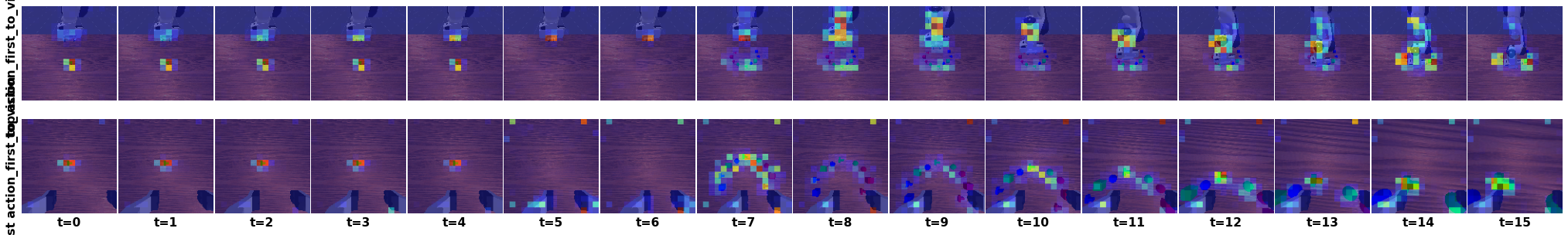}\\[-2pt]
  \footnotesize (b) action$\to$vision (first action token query).
\end{minipage}\\[0.3em]
\begin{minipage}{0.98\textwidth}\centering
  \includegraphics[width=\linewidth]{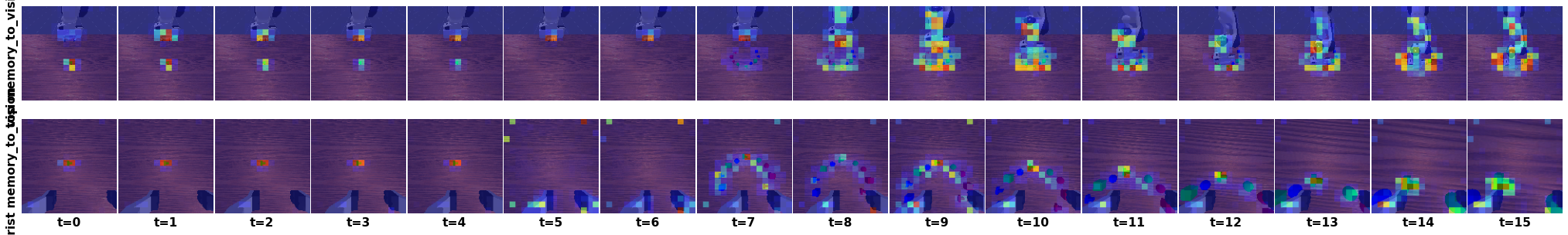}\\[-2pt]
  \footnotesize (c) memory$\to$vision.
\end{minipage}\\[0.3em]
\begin{minipage}{0.98\textwidth}\centering
  \includegraphics[width=\linewidth]{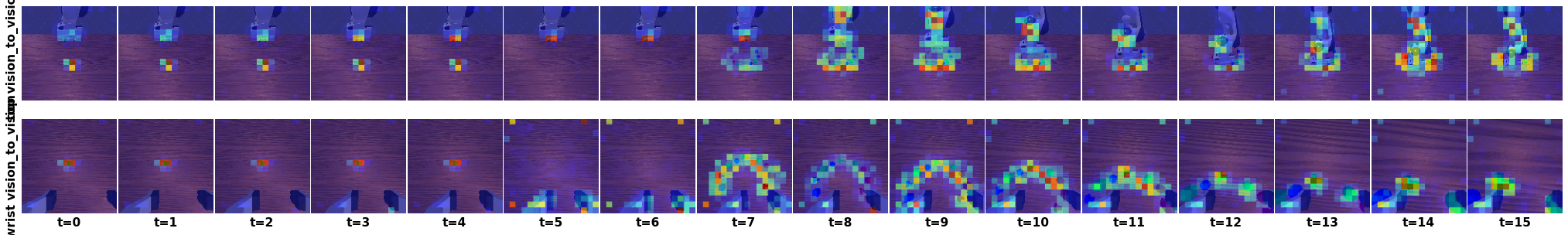}\\[-2pt]
  \footnotesize (d) vision$\to$vision.
\end{minipage}
\caption{\textbf{Attention rollouts on \texttt{RememberShapeAndColor3x3}
  ($K{=}2$, $m{=}64$).} The cue panel is a $3\times 3$ shape-by-colour
  grid that is fully revealed early and then masked. The
  memory$\to$vision row places its peaks on the grid cells while they
  are visible and broadens once the panel is occluded, matching the
  two-phase write event in Figure~\ref{fig:mem_dynamics_remsac}. The
  action$\to$vision row tracks the gripper approaching the matching
  object on the table. The vision$\to$vision baseline lights up
  generic object-shaped clusters at every step and is not cue-specific.}
\label{fig:attn_rollout_remsac}
\end{figure*}

\begin{figure*}[t]
\centering
\begin{minipage}{0.98\textwidth}\centering
  \includegraphics[width=\linewidth]{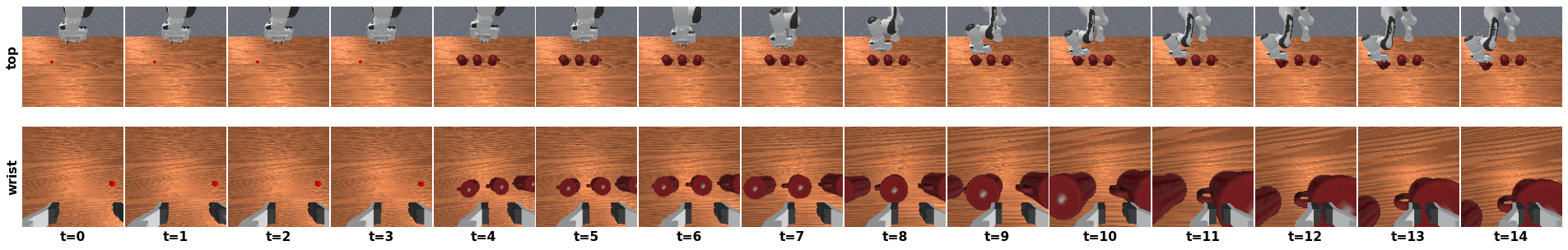}\\[-2pt]
  \footnotesize (a) Plain rollout (top + wrist, 15 steps, success).
\end{minipage}\\[0.3em]
\begin{minipage}{0.98\textwidth}\centering
  \includegraphics[width=\linewidth]{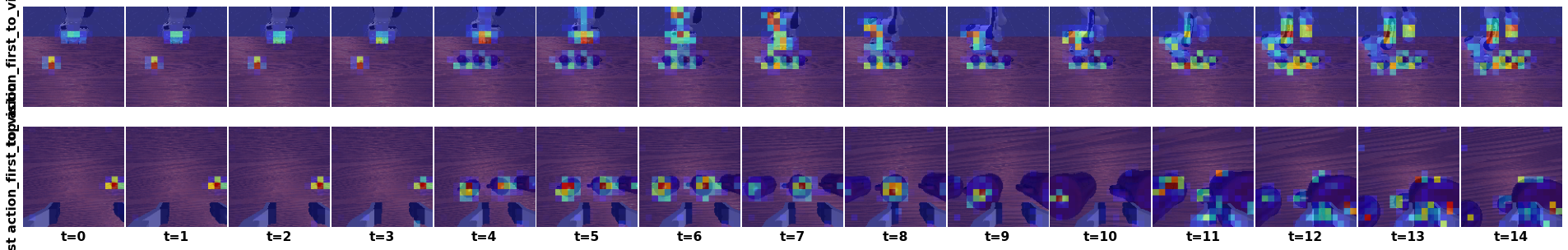}\\[-2pt]
  \footnotesize (b) action$\to$vision (first action token query).
\end{minipage}\\[0.3em]
\begin{minipage}{0.98\textwidth}\centering
  \includegraphics[width=\linewidth]{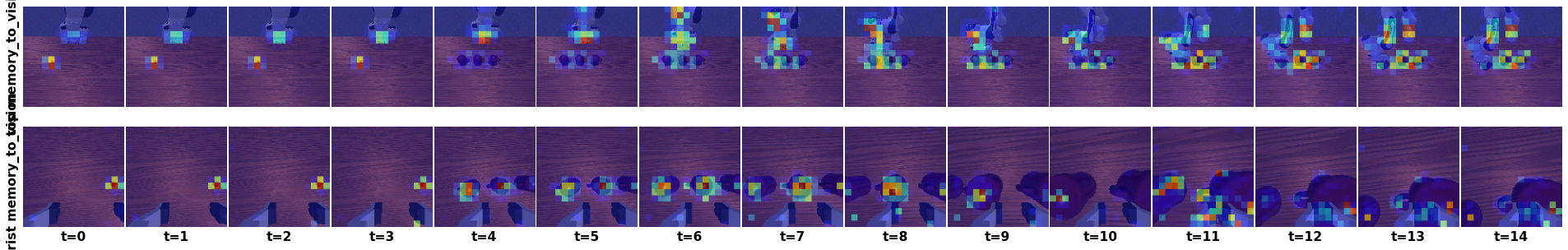}\\[-2pt]
  \footnotesize (c) memory$\to$vision.
\end{minipage}\\[0.3em]
\begin{minipage}{0.98\textwidth}\centering
  \includegraphics[width=\linewidth]{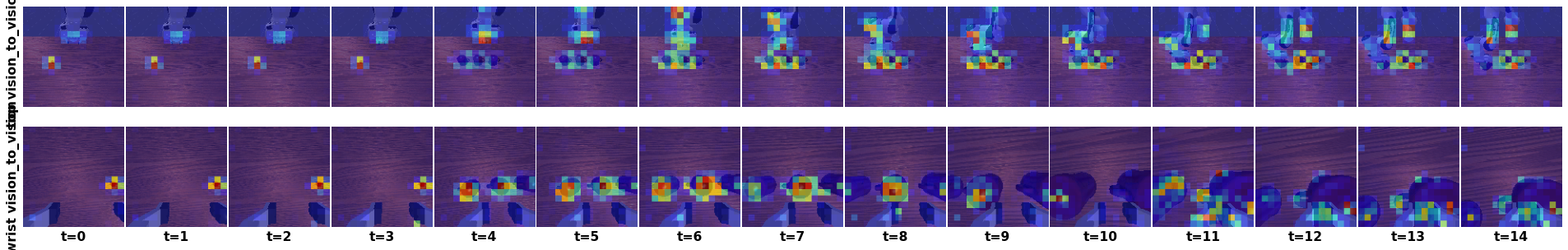}\\[-2pt]
  \footnotesize (d) vision$\to$vision.
\end{minipage}
\caption{\textbf{Attention rollouts on \texttt{ShellGamePush}
  ($K{=}2$, $m{=}64$).} The episode has a single-write cue dynamic:
  the ball is visible only in the first frames before the cup
  occludes it, and the cosine trace
  in Figure~\ref{fig:mem_dynamics_shellgamepush} shows the
  corresponding $t{=}1$ spike. The memory$\to$vision row places its
  peak on the cup that hides the ball in those frames, then carries
  a flat low-magnitude pattern as the cup is pushed. The
  action$\to$vision row tracks the cup being pushed by the gripper.
  The vision$\to$vision baseline remains object-anchored throughout.}
\label{fig:attn_rollout_shellgamepush}
\end{figure*}

\begin{figure*}[t]
\centering
\begin{minipage}{0.98\textwidth}\centering
  \includegraphics[width=\linewidth]{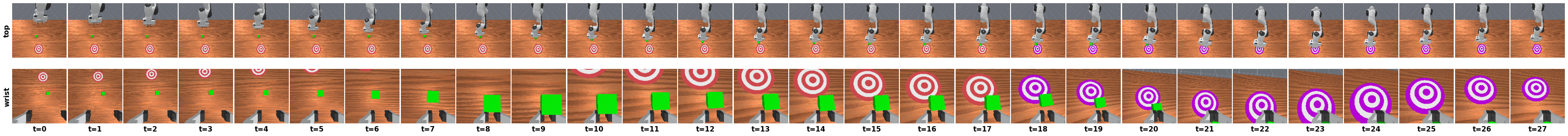}\\[-2pt]
  \footnotesize (a) Plain rollout (top + wrist, 28 steps, success).
\end{minipage}\\[0.3em]
\begin{minipage}{0.98\textwidth}\centering
  \includegraphics[width=\linewidth]{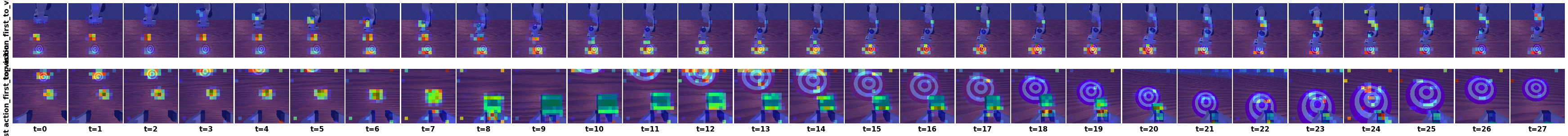}\\[-2pt]
  \footnotesize (b) action$\to$vision (first action token query).
\end{minipage}\\[0.3em]
\begin{minipage}{0.98\textwidth}\centering
  \includegraphics[width=\linewidth]{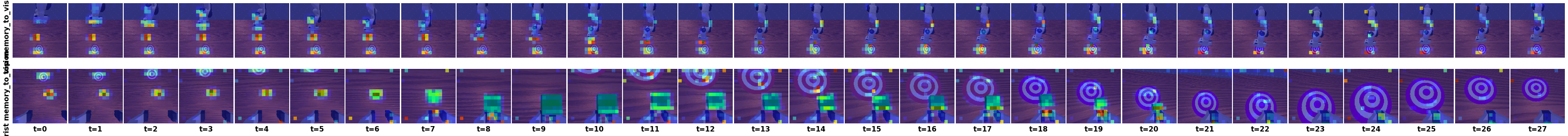}\\[-2pt]
  \footnotesize (c) memory$\to$vision.
\end{minipage}\\[0.3em]
\begin{minipage}{0.98\textwidth}\centering
  \includegraphics[width=\linewidth]{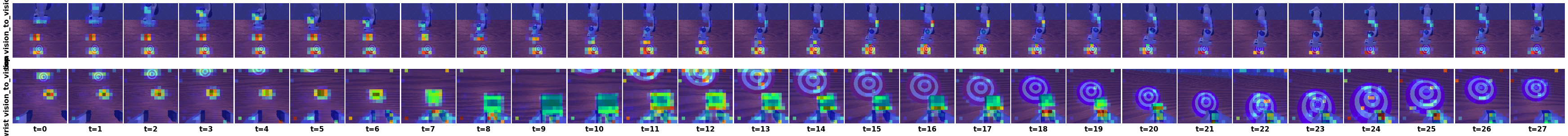}\\[-2pt]
  \footnotesize (d) vision$\to$vision.
\end{minipage}
\caption{\textbf{Attention rollouts on \texttt{TakeItBack}
  ($K{=}2$, $m{=}64$).} The task has two phases (move out, return),
  and the cosine trace in Figure~\ref{fig:mem_dynamics_takeitback}
  shows two write events of comparable magnitude at $t{=}1$ and at
  the phase transition near $t{\approx}18$. The memory$\to$vision row
  flares in both windows: it locks onto the start position in the
  early frames and re-engages with the held object when the policy
  switches to the return phase. The action$\to$vision row tracks the
  object being carried by the gripper across both phases. The
  vision$\to$vision baseline is object-anchored without phase
  selectivity.}
\label{fig:attn_rollout_takeitback}
\end{figure*}

\begin{figure*}[t]
\centering
\begin{minipage}{0.98\textwidth}\centering
  \includegraphics[width=\linewidth]{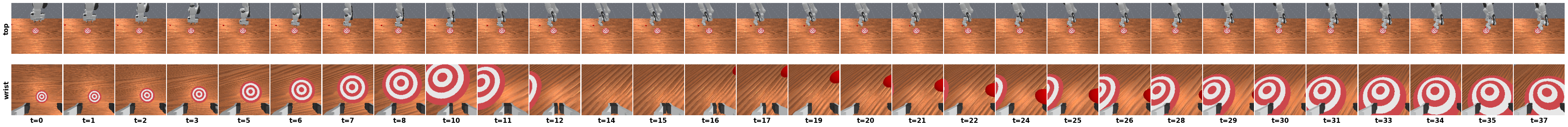}\\[-2pt]
  \footnotesize (a) Plain rollout (top + wrist, 38 steps, success).
\end{minipage}\\[0.3em]
\begin{minipage}{0.98\textwidth}\centering
  \includegraphics[width=\linewidth]{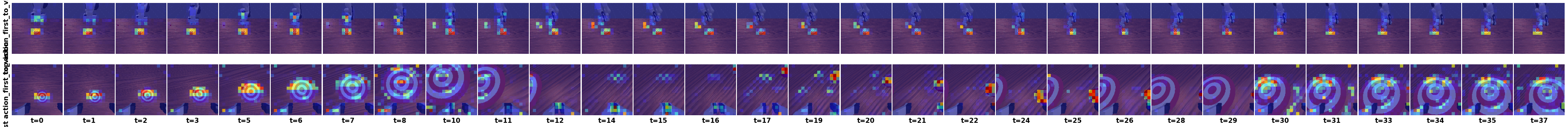}\\[-2pt]
  \footnotesize (b) action$\to$vision (first action token query).
\end{minipage}\\[0.3em]
\begin{minipage}{0.98\textwidth}\centering
  \includegraphics[width=\linewidth]{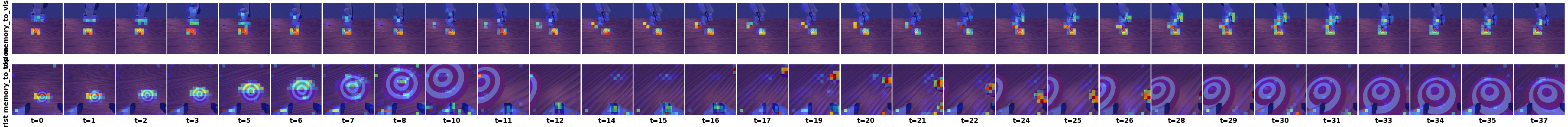}\\[-2pt]
  \footnotesize (c) memory$\to$vision.
\end{minipage}\\[0.3em]
\begin{minipage}{0.98\textwidth}\centering
  \includegraphics[width=\linewidth]{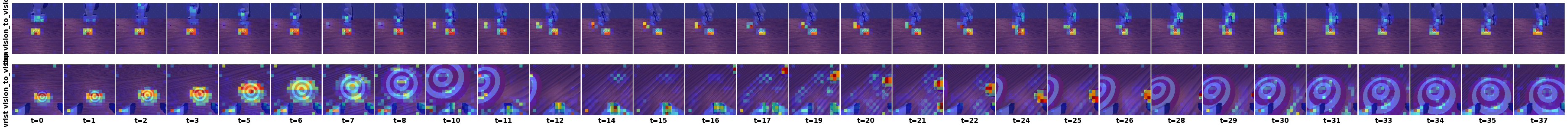}\\[-2pt]
  \footnotesize (d) vision$\to$vision.
\end{minipage}
\caption{\textbf{Attention rollouts on \texttt{InterceptMedium}
  ($K{=}2$, $m{=}64$).} The task requires continuous re-grounding
  rather than a one-shot write, and the cosine trace in
  Figure~\ref{fig:mem_dynamics_intercept} accordingly shows a
  sustained mid-episode plateau and an L2-norm bump at the catch
  event. The memory$\to$vision row tracks the moving object across
  the table at every step, with its highest weight near the catch.
  The action$\to$vision row stays on the gripper and the projected
  interception point. The vision$\to$vision baseline is again
  object-anchored but does not concentrate on the trajectory.}
\label{fig:attn_rollout_intercept}
\end{figure*}

\section{Per-Environment Memory Intervention}
\label{app:intervention-full}

This appendix collects the full memory-intervention sweep referenced
in \S\ref{sec:exp:diagnostics:06}. The representative panel for
\texttt{RememberColor5} appears inline as
Figure~\ref{fig:intervention}; per-environment panels for the four
remaining \mikasa\ training tasks are given below. All bars report
success rate at $100$ episodes per cell. The three conditions are
baseline (clean memory), \texttt{noise} (memory replaced with i.i.d.\
Gaussian noise before every forward), and \texttt{freeze\_first}
(memory locked to its first-step value $\boldsymbol{M}_1$).

\begin{figure*}[t]
\centering
\begin{minipage}[t]{0.47\textwidth}\centering
  \includegraphics[width=\linewidth]{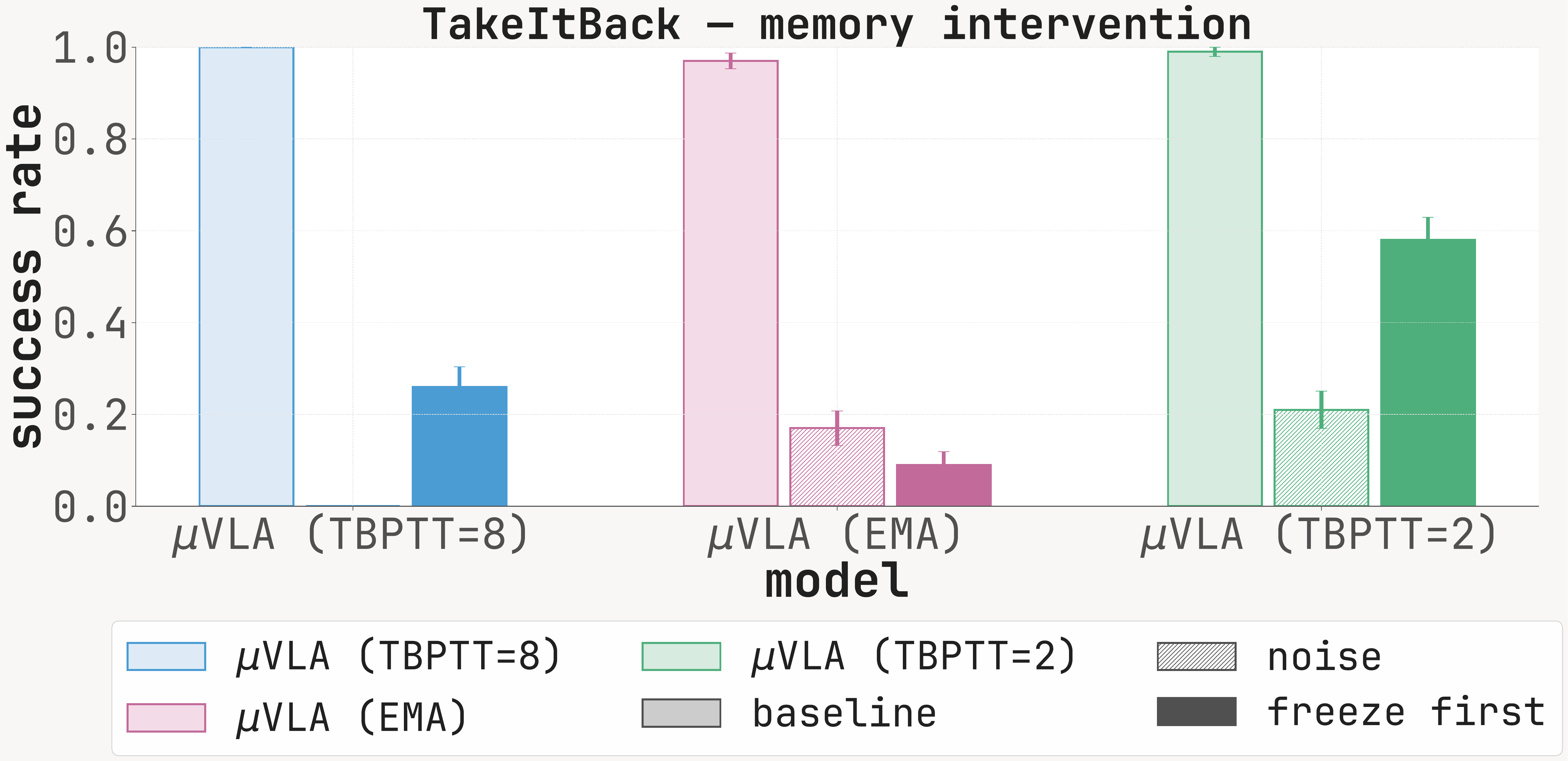}\\[-2pt]
  \footnotesize (a) \texttt{TakeItBack}
\end{minipage}\hfill
\begin{minipage}[t]{0.47\textwidth}\centering
  \includegraphics[width=\linewidth]{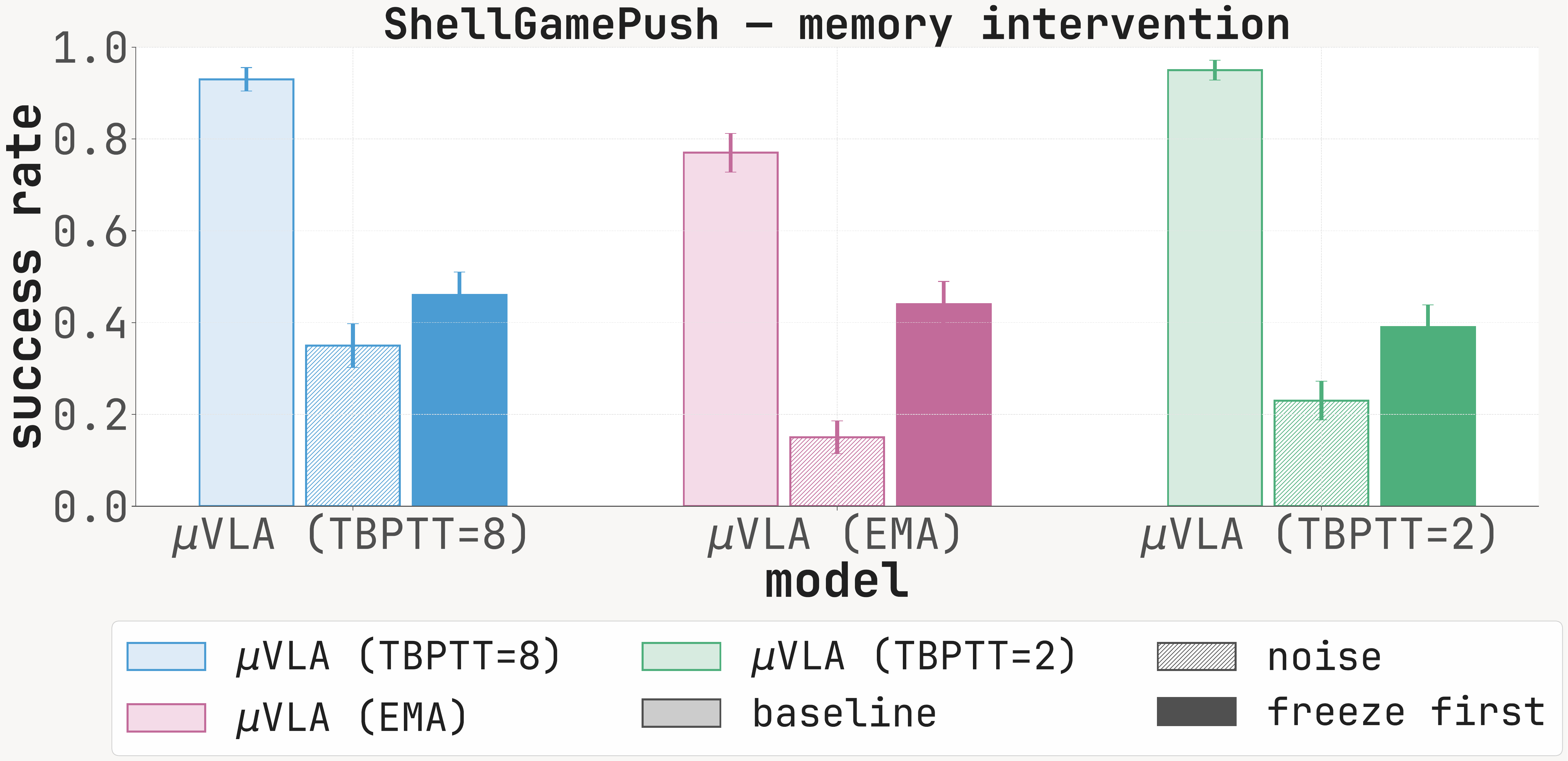}\\[-2pt]
  \footnotesize (b) \texttt{ShellGamePush}
\end{minipage}

\vspace{0.6em}

\begin{minipage}[t]{0.47\textwidth}\centering
  \includegraphics[width=\linewidth]{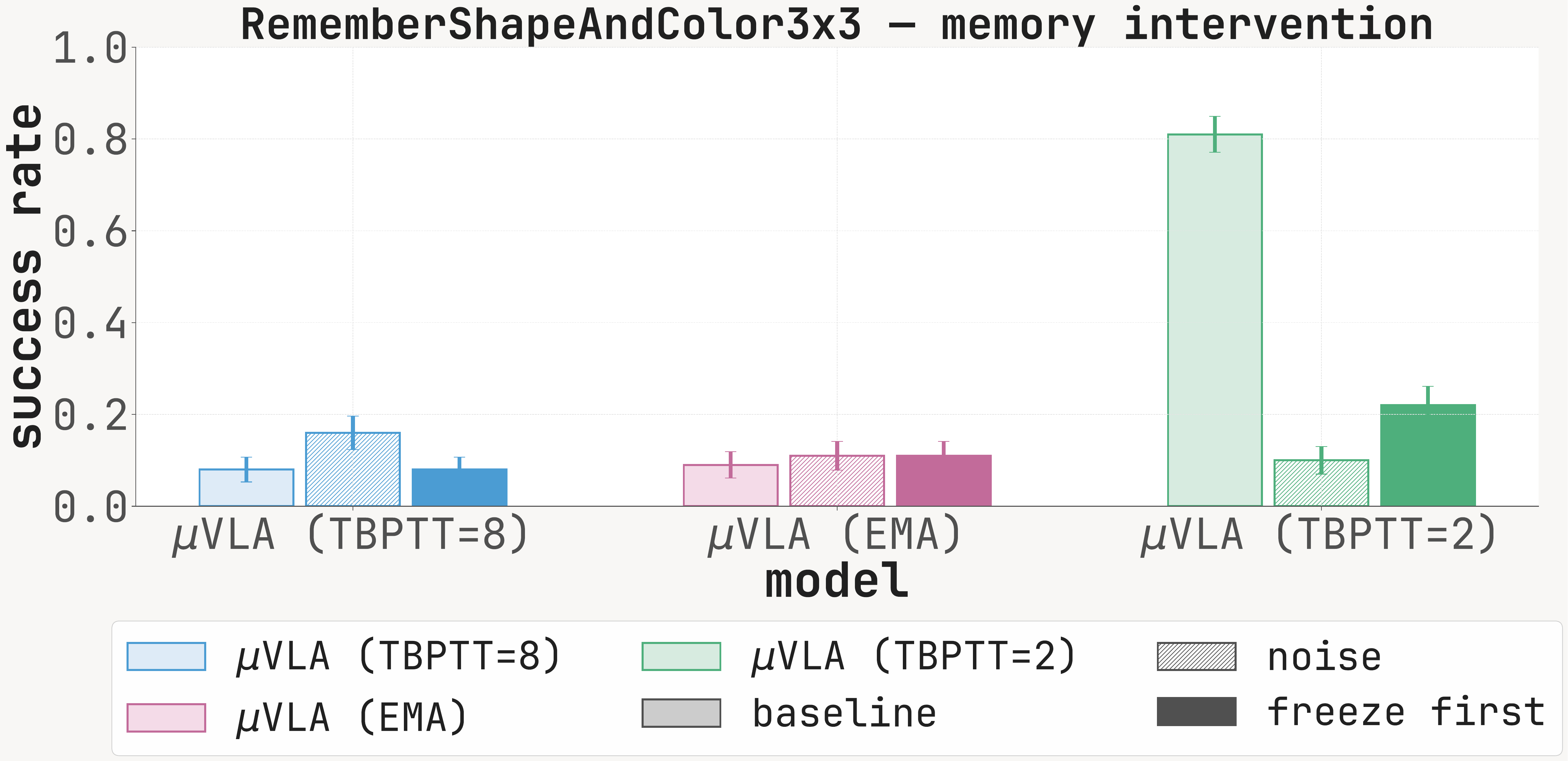}\\[-2pt]
  \footnotesize (c) \texttt{RememberShapeAndColor3x3}
\end{minipage}\hfill
\begin{minipage}[t]{0.47\textwidth}\centering
  \includegraphics[width=\linewidth]{figures/ablations/intervention_combined4.pdf}\\[-2pt]
  \footnotesize (d) \texttt{RememberColor5}
\end{minipage}

\vspace{0.6em}

\makebox[\textwidth][c]{%
\begin{minipage}[t]{0.47\textwidth}\centering
  \includegraphics[width=\linewidth]{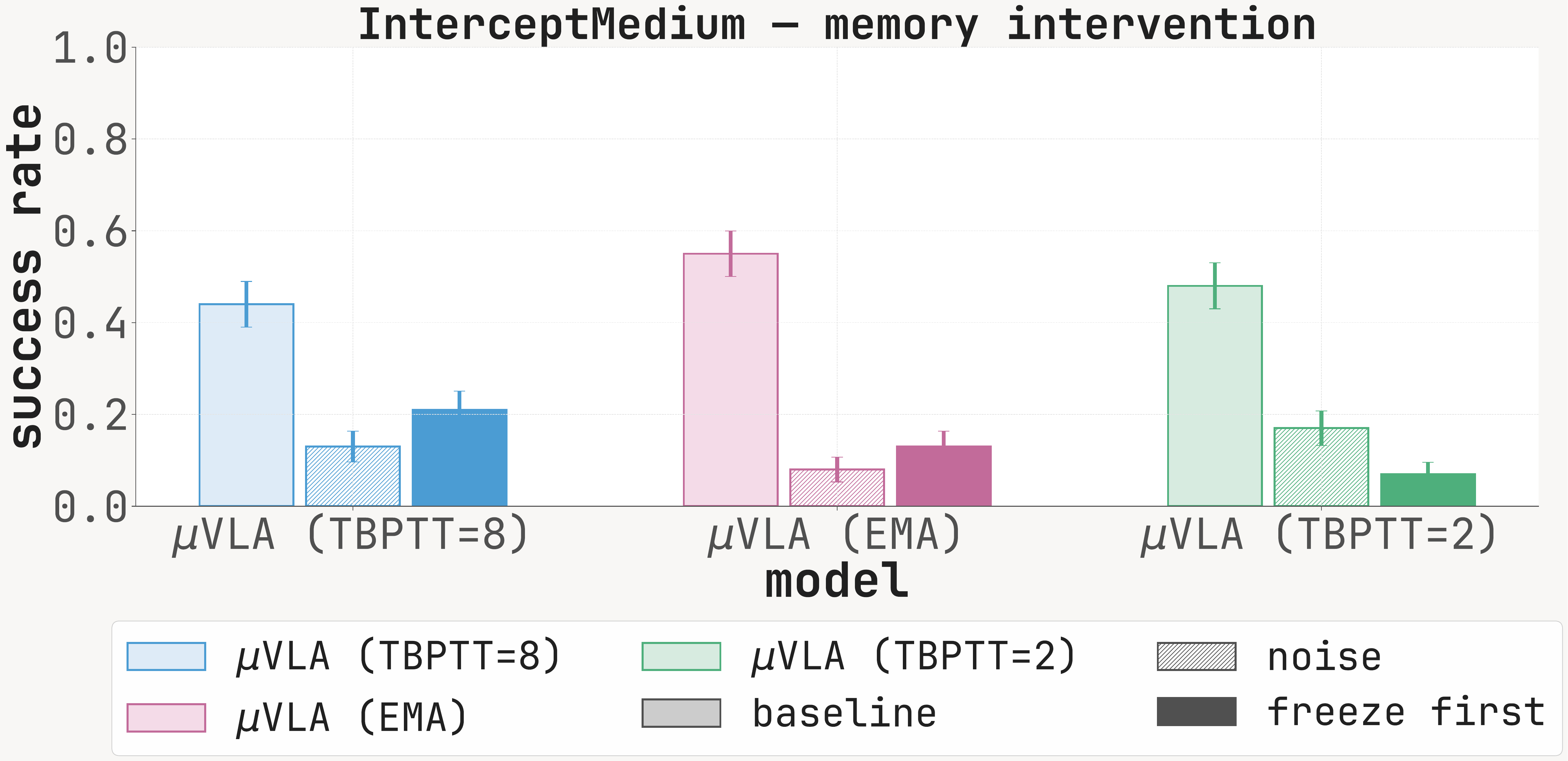}\\[-2pt]
  \footnotesize (e) \texttt{InterceptMedium}
\end{minipage}%
}

\caption{\textbf{Memory intervention on the five \mikasa\ training
tasks.}
The \texttt{noise} bar is below the baseline in every cell with a
non-trivial baseline, confirming that the recurrent channel is
functionally read at inference. \texttt{freeze\_first} retains part
of the SR on cue-recall tasks where the cue is localised in the
first frames (RC5, TakeItBack with $K{=}2$), and collapses on
dynamics-grounded tasks (\texttt{InterceptMedium}). Companion to
Figure~\ref{fig:intervention} in the main text.}
\label{fig:intervention_full}
\end{figure*}

\begin{figure*}[t]
\centering
\begin{minipage}[t]{0.47\textwidth}\centering
  \includegraphics[width=\linewidth]{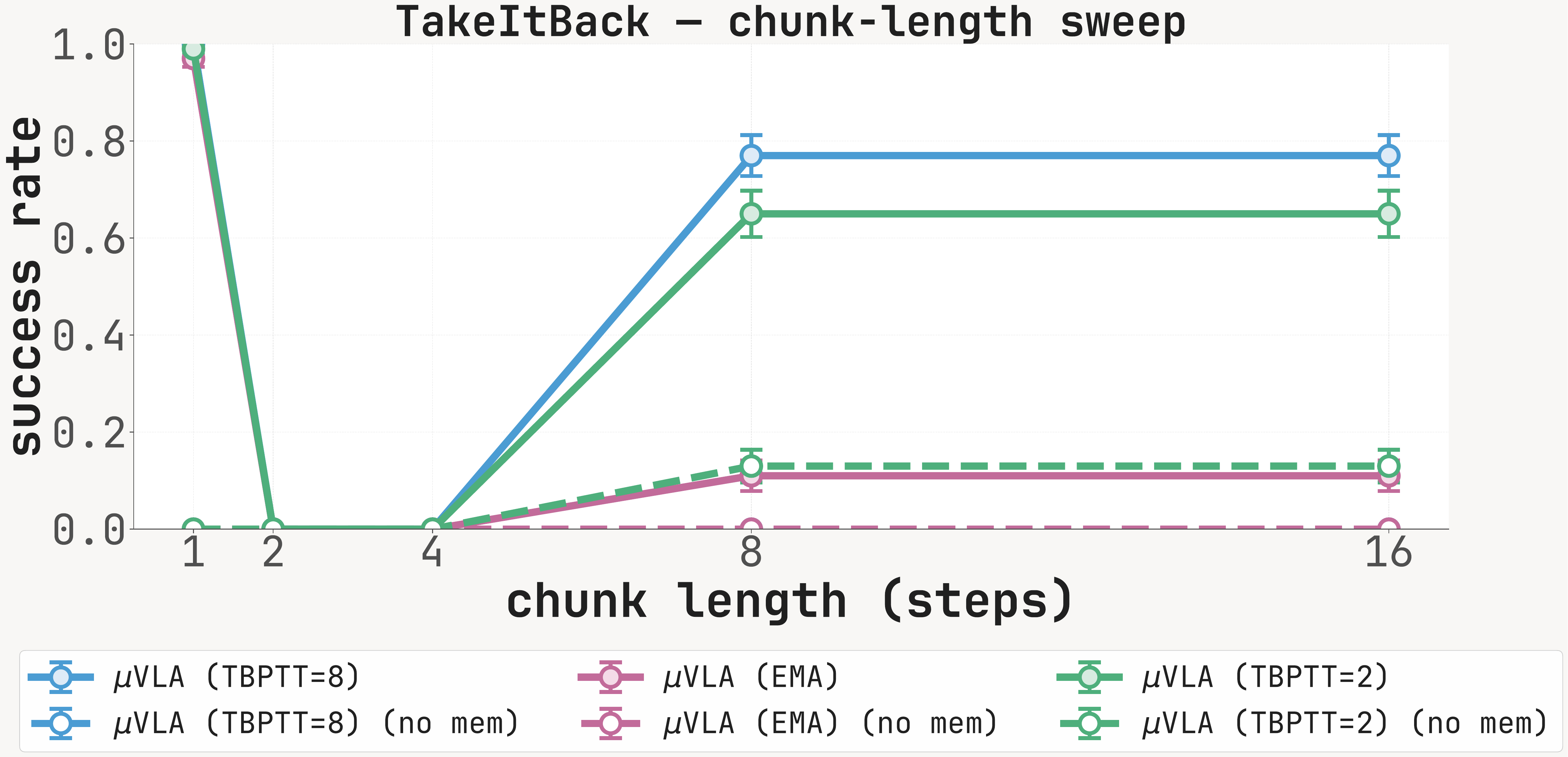}\\[-2pt]
  \footnotesize (a) \texttt{TakeItBack}
\end{minipage}\hfill
\begin{minipage}[t]{0.47\textwidth}\centering
  \includegraphics[width=\linewidth]{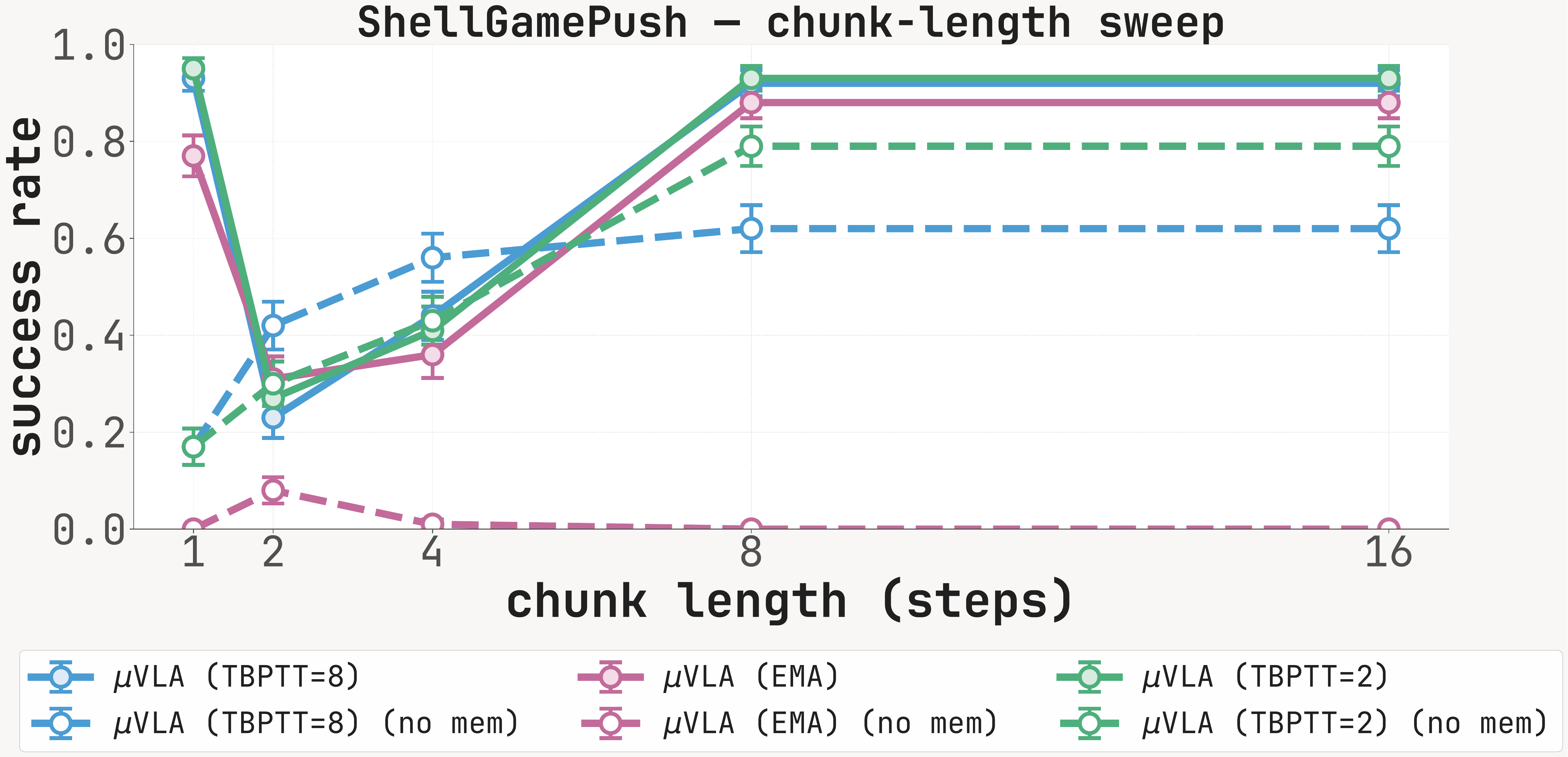}\\[-2pt]
  \footnotesize (b) \texttt{ShellGamePush}
\end{minipage}

\vspace{0.6em}

\begin{minipage}[t]{0.47\textwidth}\centering
  \includegraphics[width=\linewidth]{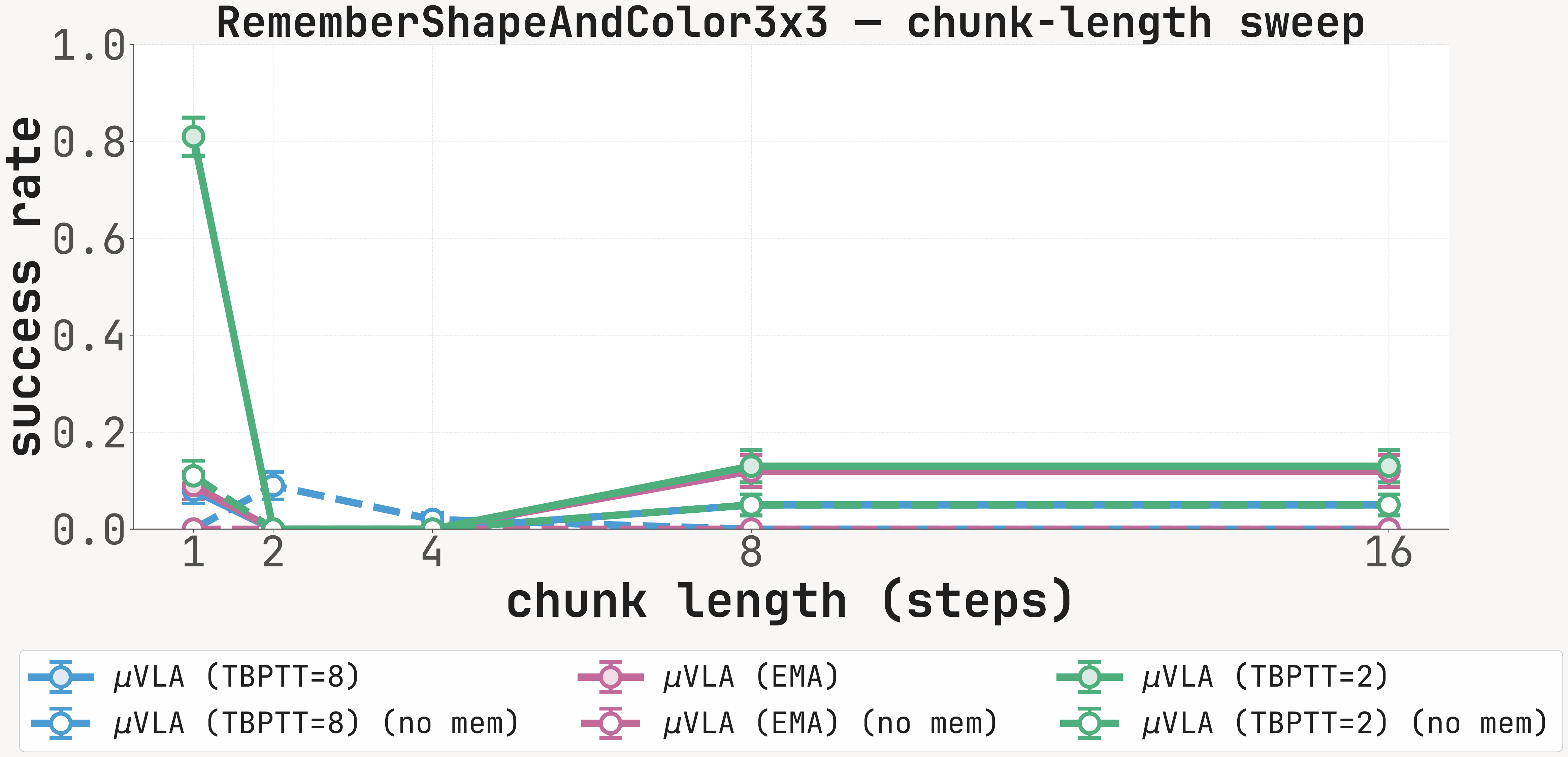}\\[-2pt]
  \footnotesize (c) \texttt{RememberShapeAndColor3x3}
\end{minipage}\hfill
\begin{minipage}[t]{0.47\textwidth}\centering
  \includegraphics[width=\linewidth]{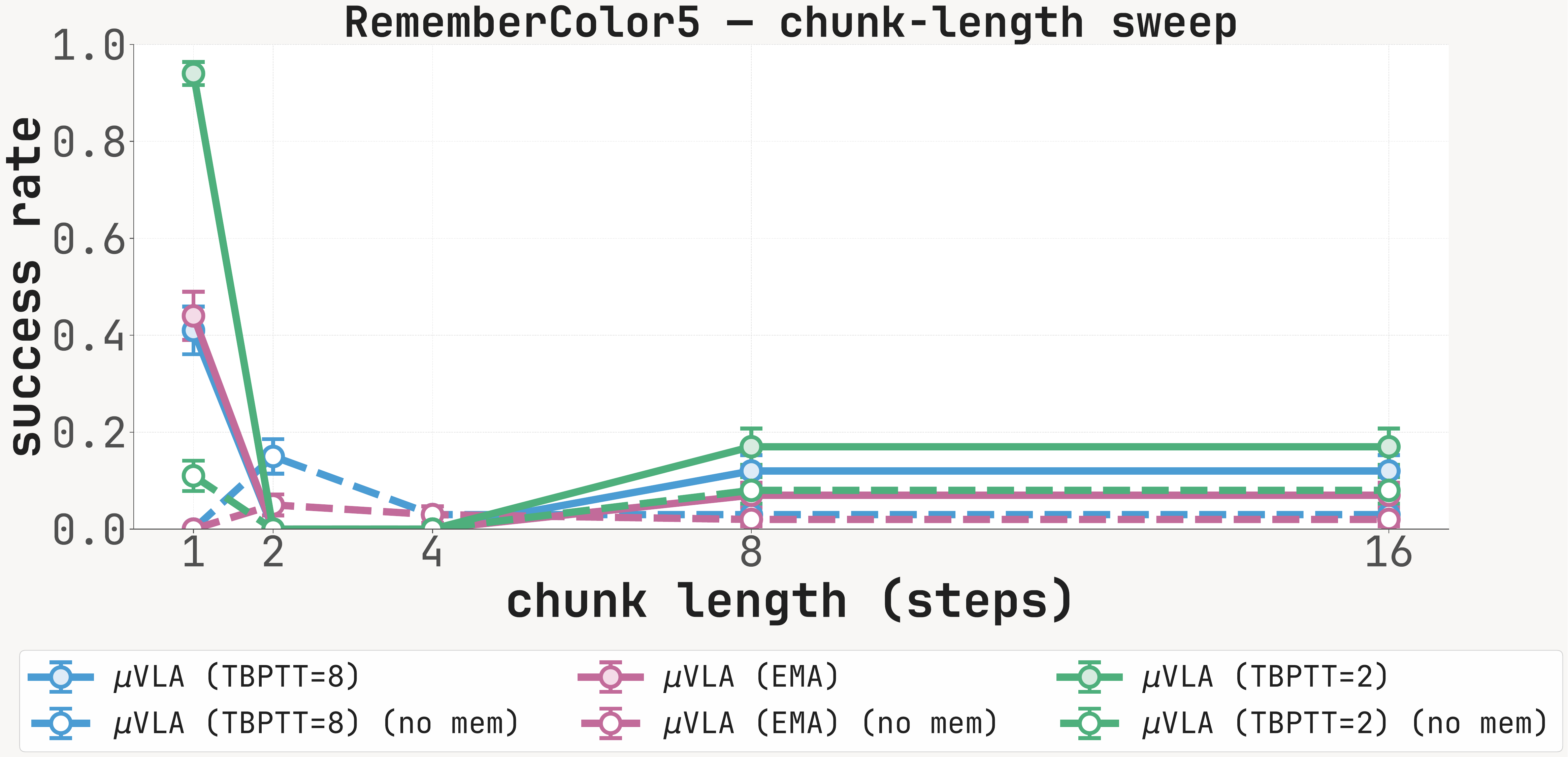}\\[-2pt]
  \footnotesize (d) \texttt{RememberColor5}
\end{minipage}

\vspace{0.6em}

\makebox[\textwidth][c]{%
\begin{minipage}[t]{0.47\textwidth}\centering
  \includegraphics[width=\linewidth]{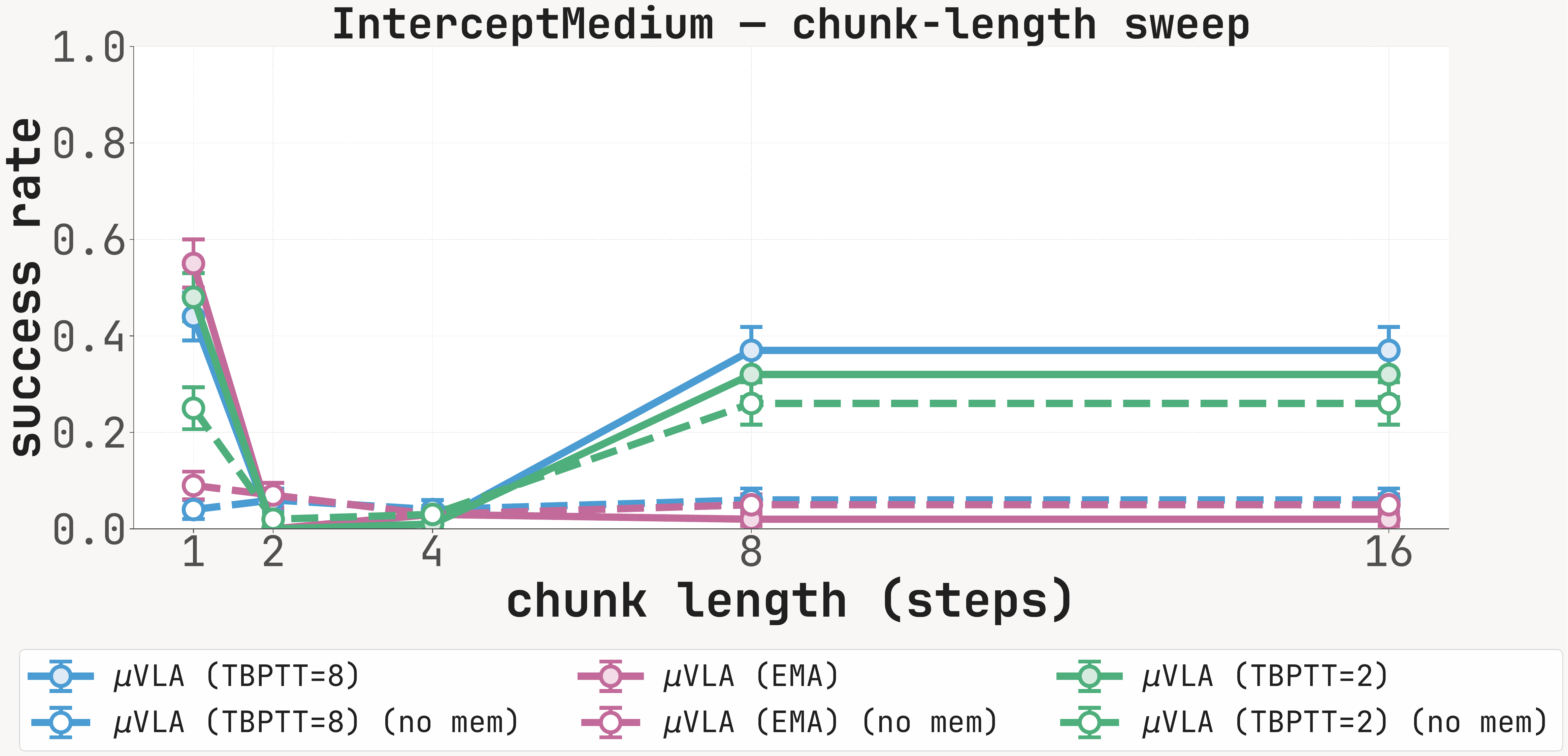}\\[-2pt]
  \footnotesize (e) \texttt{InterceptMedium}
\end{minipage}%
}

\caption{\textbf{Chunk-length sweep on the five \mikasa\ training
tasks.}
For each carrier ($K{=}8$, EMA, $K{=}2$) we report SR at chunk
lengths $\{1, 2, 4, 8, 16\}$ with the memory channel active (solid)
and zeroed at inference (dashed). $\text{chunk}{=}1$ is the
receding-horizon regime used at training. The with-memory minus
no-memory gap is largest at $\text{chunk}{=}1$ on cue-recall tasks
(RC5, RemSAC3x3, TakeItBack), while long-chunk inference partially
bypasses the recurrent channel.}
\label{fig:chunk_sweep}
\end{figure*}

\section{MIKASA-Robo Training Environment Descriptions}
\label{app:envs}

\begin{description}
  \item[\texttt{RememberColor5-VLA-v0}]
    At the beginning of each episode, one of five colored lamps illuminates
    for a brief period, indicating which of five identically shaped objects
    the robot should grasp.
    The lamp turns off before the robot must act.
    Requires: color memory.

  \item[\texttt{RememberShapeAndColor3x3-VLA-v0}]
    A $3 \times 3$ grid of object properties (shape $\times$ color) is
    presented at the start of the episode.
    The robot must retrieve the object matching a specified shape-color pair.
    Requires: joint shape and color memory.

  \item[\texttt{ShellGamePush-VLA-v0}]
    Three cups are shuffled in a sequence visible only at the episode start.
    The robot must push the cup concealing a target ball.
    Requires: tracking through occlusion.

  \item[\texttt{TakeItBack-VLA-v0}]
    An object is placed in a visible goal region at episode start, then
    displaced by an external disturbance.
    The robot must return it to the original goal position, which is no longer
    marked in the observation.
    Requires: spatial memory.

  \item[\texttt{InterceptMedium-VLA-v0}]
    A target object moves along a trajectory observable early in the episode.
    The robot must intercept the object at a future position.
    Requires: predictive memory / velocity estimation.
\end{description}

\section{Full MIKASA-Robo Task List, Descriptions, and Language Instructions}
\label{app:tasks}

Table~\ref{tab:tasks} lists all \mikasa environments together with their
horizons, task descriptions, and the natural-language instructions used
during training and evaluation; tasks used for multi-task training are
bolded. Task descriptions follow the MIKASA-Robo-VLA documentation
(\url{https://mikasarobo.github.io/}).
LIBERO instructions follow the suite's distributed defaults and are not
duplicated here.

\begin{table}[ht!]
\caption{
  \textbf{Full \mikasa task list, task descriptions, and language
  instructions used in this paper.} Tasks in bold are the five used for
  multi-task training; the remaining environments are held out and used
  as a transfer probe. Horizons are the maximum episode lengths. Task
  descriptions follow the MIKASA-Robo-VLA documentation
  (\url{https://mikasarobo.github.io/}).
}
\label{tab:tasks}
\centering
\setlength{\tabcolsep}{4pt}
\begin{adjustbox}{width=\textwidth}
\begin{tabular}{r l c p{6.4cm} p{7.6cm}}
\toprule
\# & \textbf{Environment} & \textbf{Horizon}
   & \textbf{Task description}
   & \textbf{Language instruction} \\
\midrule
0  & \textbf{ShellGameTouch-VLA-v0}  & 30
   & Short hidden object memory: observe which cup hides the ball, wait, and touch that cup.
   & Observe which cup hides the ball, wait, then touch that cup. \\
1  & ShellGamePush-VLA-v0   & 30
   & Hidden object selection: remember which cup hides the ball and push that cup forward.
   & Observe which cup hides the ball, wait, then push that cup forward. \\
2  & ShellGamePick-VLA-v0   & 30
   & Hidden object selection: remember which cup hides the ball and pick that cup up.
   & Observe which cup hides the ball, wait, then pick up that cup and lift it. \\
3  & InterceptSlow-VLA-v0   & 60
   & Predictive interception: infer a rolling ball path and deflect it toward the target.
   & Intercept the rolling ball by moving to its path and deflecting it toward the target. \\
4  & \textbf{InterceptMedium-VLA-v0} & 60
   & Predictive interception: infer a rolling ball path and deflect it toward the target.
   & Intercept the rolling ball by moving to its path and deflecting it toward the target. \\
5  & InterceptFast-VLA-v0   & 60
   & Predictive interception: infer a rolling ball path and deflect it toward the target.
   & Intercept the rolling ball by moving to its path and deflecting it toward the target. \\
6  & InterceptGrabSlow-VLA-v0   & 60
   & Predictive capture: infer a rolling ball path and grasp or stop it.
   & Intercept the rolling ball and grasp it to stop it. \\
7  & InterceptGrabMedium-VLA-v0 & 60
   & Predictive capture: infer a rolling ball path and grasp or stop it.
   & Intercept the rolling ball and grasp it to stop it. \\
8  & InterceptGrabFast-VLA-v0   & 60
   & Predictive capture: infer a rolling ball path and grasp or stop it.
   & Intercept the rolling ball and grasp it to stop it. \\
9  & RotateLenientPos-VLA-v0    & 60
   & Spatial transformation: rotate a peg by a requested angle with a lenient center-position criterion.
   & Rotate the peg by \{angle\_deg\} degrees to match the target angle. \\
10 & RotateLenientPosNeg-VLA-v0 & 60
   & Spatial transformation: rotate a peg by a requested angle with a lenient center-position criterion.
   & Rotate the peg by \{angle\_deg\} degrees to match the target angle. \\
11 & RotateStrictPos-VLA-v0     & 90
   & Spatial transformation: rotate a peg while keeping its center close to the original position.
   & Rotate the peg by \{angle\_deg\} degrees to match the target angle while keeping the center of the peg in place. \\
12 & RotateStrictPosNeg-VLA-v0  & 90
   & Spatial transformation: rotate a peg while keeping its center close to the original position.
   & Rotate the peg by \{angle\_deg\} degrees to match the target angle while keeping the center of the peg in place. \\
13 & \textbf{TakeItBack-VLA-v0} & 60
   & Spatial restoration: push a cube to a target, then return it to its original position after the target changes.
   & Push the cube onto the red target, and when the target changes color, return the cube to its original position. \\
14 & RememberColor3-VLA-v0      & 25
   & Delayed color recall: observe one target color, wait, and touch the cube with the same color.
   & Observe the cube's color, wait, then touch the cube of the same color. \\
15 & \textbf{RememberColor5-VLA-v0} & 25
   & Delayed color recall: observe one target color, wait, and touch the cube with the same color.
   & Observe the cube's color, wait, then touch the cube of the same color. \\
16 & RememberColor9-VLA-v0      & 25
   & Delayed color recall: observe one target color, wait, and touch the cube with the same color.
   & Observe the cube's color, wait, then touch the cube of the same color. \\
17 & RememberShape3-VLA-v0      & 25
   & Delayed shape recall: observe one target shape, wait, and touch the object with the same shape.
   & Observe the object's shape, wait, then touch the object of the same shape. \\
18 & RememberShape5-VLA-v0      & 25
   & Delayed shape recall: observe one target shape, wait, and touch the object with the same shape.
   & Observe the object's shape, wait, then touch the object of the same shape. \\
19 & RememberShape9-VLA-v0      & 25
   & Delayed shape recall: observe one target shape, wait, and touch the object with the same shape.
   & Observe the object's shape, wait, then touch the object of the same shape. \\
20 & RememberShapeAndColor3x2-VLA-v0 & 25
   & Delayed binding recall: remember both shape and color and select the matching object.
   & Observe the object's shape and color, wait, then touch the object of the same shape and color. \\
21 & \textbf{RememberShapeAndColor3x3-VLA-v0} & 25
   & Delayed binding recall: remember both shape and color and select the matching object.
   & Observe the object's shape and color, wait, then touch the object of the same shape and color. \\
22 & RememberShapeAndColor5x3-VLA-v0 & 25
   & Delayed binding recall: remember both shape and color and select the matching object.
   & Observe the object's shape and color, wait, then touch the object of the same shape and color. \\
\bottomrule
\end{tabular}
\end{adjustbox}
\end{table}

\section{Extended Related Work}

\paragraph{Vision-language-action models.}
RT-1~\citep{rt1} introduced a scalable transformer policy that
tokenizes images, language, and robot actions for real-world
control.
RT-2~\citep{rt2} extended this paradigm by co-fine-tuning
pretrained vision-language models on web-scale
vision-language data and robot trajectories, expressing
actions themselves as text tokens.
OpenVLA~\citep{kim2024openvla} released an open-source 7B VLA and
showed that it can be adapted efficiently with low-rank
adaptation (LoRA)~\citep{lora}, while \openvlaoft~\citep{openvlaoft}
demonstrated a stronger adaptation recipe based on parallel 
action-chunk decoding, continuous actions, and an L1
regression objective, while accommodating additional robot
inputs such as proprioception and wrist cameras.
$\pi_0$~\citep{pi0} replaces autoregressive token prediction
with flow-matching action generation, and
$\pi_{0.5}$~\citep{pi05} broadens this recipe through
heterogeneous co-training for open-world generalization.
Octo~\citep{octo} scales generalist policy pretraining across
heterogeneous embodiments; GR00T N1~\citep{gr00t} targets
humanoid control with a dual-system architecture;
X-VLA~\citep{xvla} uses embodiment-specific soft prompts for
cross-embodiment adaptation; SmolVLA~\citep{smolvla}
emphasizes lower-cost deployment; and
F1~\citep{f1}, EgoVLA~\citep{egovla}, and
DexGraspVLA~\citep{dexgraspvla} specialize the recipe with
visual foresight, egocentric human-video pretraining, and
dexterous grasping, respectively.
Despite this diversity, mainstream VLAs still condition
primarily on the current observation or a fixed multi-frame
window, leaving long-horizon partial observability largely
unresolved.

\paragraph{Recurrent latent memory.}
The closest line of work to \muvla learns a compact latent
state that is updated online across control steps.
ReMem-VLA~\citep{rememvla} attaches frame-level and
chunk-level recurrent query tokens and trains them with an
auxiliary past-observation prediction objective.
AVA-VLA~\citep{avavla} models control from a POMDP
perspective and uses a recurrent state as a neural belief
state for active visual attention over current visual tokens.
VPWEM~\citep{vpwem} combines a sliding-window working memory
with a transformer-based episodic compressor that
recursively summarizes older observations into a fixed set
of memory tokens.
MemoryVLA~\citep{memoryvla} augments working memory with a
perceptual--cognitive memory bank that stores, retrieves,
fuses, and consolidates both low-level and high-level
context.
Recursive Belief VLAs~\citep{recursivebelief} learn an
action-conditioned latent belief via self-supervised
world-model objectives, while
Embodied-SlotSSM~\citep{objcentricmem} replaces monolithic
recurrence with persistent object-centric slots and
state-space dynamics.
A nearby but distinct idea is Recurrent-Depth
VLA~\citep{recdepthvla}, which places recurrence inside a
weight-tied action head to scale per-step inference depth
rather than to carry memory across environment steps.
Across this group, the recurrence is consistently paired
with additional machinery --- auxiliary belief or
past-observation losses, dedicated memory modules, or
modified action heads --- and the contribution of the
recurrence \emph{itself} is rarely measured in isolation.
\muvla is deliberately the opposite point in the design
space: a few learnable tokens carried inside the backbone's
own self-attention with no auxiliary loss, no extra
parameters beyond the tokens, and no architectural surgery,
so that the gain attributable to the recurrence primitive
can be read off directly.
The only departure from the host \openvlaoft attention
pattern is to block memory tokens from attending to action
tokens, which we show is necessary to prevent the
recurrence from collapsing to copying the predicted action
into memory.

\paragraph{Compressed or amortized temporal context.}
A second family extends the horizon without maintaining a
learned per-step recurrent state.
ContextVLA~\citep{contextvla} compresses past observations
into a single context token.
CronusVLA~\citep{li2025cronusvla} introduces a two-stage
single-frame pretraining plus multi-frame post-training
recipe with feature-chunk aggregation.
Long-VLA~\citep{longvla} uses phase-aware input masking for
long-horizon subtask structure,
HiF-VLA~\citep{hifvla} replaces stacked frames with
motion-based hindsight and foresight representations, and
LoLA~\citep{wang2025lola} combines current-observation encoding with
downsampled historical motion encoding in an
embodiment-grounded latent action space.
TempoFit~\citep{tempofit} repurposes frozen prefix KV states
as a training-free temporal memory,
KV-Efficient VLA~\citep{kvefficient} chunks and filters the
KV cache with an RNN gate, and
SD-VLA~\citep{sdvla} disentangles static and dynamic tokens
so temporally persistent content can be reused efficiently.
Mixture of Horizons~\citep{mixhorizons} jointly processes
several action-chunk horizons,
VLA Knows Its Limits~\citep{vlaknowslimits} adapts the
execution horizon at test time, and
EvoVLA~\citep{evovla} combines selective long-horizon memory
with a stage-aligned reward to reduce stage hallucination.
These methods broaden usable context substantially, but they
still rely on finite windows, cached summaries, or
execution-horizon heuristics rather than an end-to-end
recurrent latent state.

\paragraph{External memory, retrieval, and scratchpads.}
A third strategy stores history outside the backbone
dynamics in an explicit memory system.
HAMLET~\citep{hamlet} retrofits a pretrained VLA with
time-contrastively initialized moment tokens and a
lightweight temporal memory module.
MEM~\citep{torne2026mem} combines video-based short-term memory with
text-based long-term memory at different temporal scales.
MAP-VLA~\citep{mapvla} constructs a library of
stage-specific soft prompts from demonstrations and
retrieves them by trajectory similarity at test time.
MemER~\citep{memer} trains a high-level policy to select
task-relevant keyframes from prior experience and translate
them into language instructions for a lower-level executor.
Chameleon~\citep{chameleon} writes geometry-grounded
multimodal tokens into a differentiable episodic memory
stack.
EchoVLA~\citep{echovla} couples scene memory with episodic
task memory for mobile manipulation,
RoboMemory~\citep{robomemory} unifies spatial, temporal,
episodic, and semantic memory in a broader embodied-agent
framework, and Notes-to-Self~\citep{noteself} stores memory
as an explicit natural-language scratchpad.
TacMamba~\citep{tacmamba} extends this template to
high-frequency tactile streams with a Mamba-based history
compressor, and Affordance Field
Intervention~\citep{affordfield} addresses the complementary
failure mode of ``memory traps'', where a VLA overcommits to
a memorized trajectory after the scene changes.
Relative to latent recurrent-state methods, these
approaches expose memory through retrieved artifacts,
prompts, or external stores, which can be highly
interpretable but require an additional retrieval or write
mechanism.

\paragraph{Sparse history selection and structured state.}
A related line argues that useful history is sparse or
should be represented structurally instead of densely.
BPP~\citep{bpp} uses a VLM to project trajectories onto a
small set of meaningful keyframes.
Non-Markovian Keyframe Chaining~\citep{keyframechain}
learns a discriminative keyframe selector and retrieves
progress-relevant frames as interleaved visual tokens.
History-Aware Visuomotor Policy Learning via Point
Tracking~\citep{historytracking} replaces raw image history
with compact object-centric point tracks,
VQ-Memory~\citep{vqmemory} discretizes past proprioception
into VQ-VAE tokens, and
Spatial Traces~\citep{spatialtraces} overlays projected
keypoint traces on depth maps to inject spatial-temporal
context.
Gated Memory Policy~\citep{gatedmem} learns both when to
recall and what to recall through a memory gate and a
lightweight cross-attention module, while
Global Prior Meets Local Consistency~\citep{globalprior}
uses a global prior memory for diffusion initialization and
a local consistency memory for progress-aware action
generation.
CodeGraphVLP~\citep{codegraphvlp} goes further by
maintaining a persistent semantic graph under partial
observability and using executable code plus progress-guided
prompting to drive the VLA.
These ideas are largely orthogonal to recurrent backbone
memory and could be combined with it.

\paragraph{Hierarchical planners.}
Another line delegates long-horizon reasoning to a planner
above a shorter-horizon executor.
LoHoVLA~\citep{lohovla} jointly generates language subgoals
and action tokens with a shared VLM backbone.
LiLo-VLA~\citep{lilovla} decomposes tasks into linked
object-centric modules for reaching and interaction,
enabling zero-shot composition and failure recovery.
Goal2Skill~\citep{goal2skill} uses a VLM planner with
structured task memory, verification, and reflection above a
diffusion-based executor.
Trace-Conditioned VLA Planning~\citep{traceplanning}
predicts a progress-aware remaining plan consisting of a
subtask sequence and a 2D visual trace, which the executor
follows in receding-horizon fashion.
HELM~\citep{helm} combines episodic keyframe retrieval with
a learned state verifier and a rollback-and-replan
controller, and Action-Sketcher~\citep{actionsketcher}
externalizes spatial intent as editable visual sketches in a
See-Think-Sketch-Act loop.
Hierarchical planners address task decomposition and
recovery, but the underlying executor can still remain
non-Markovian within each subtask.

\paragraph{Recurrent memory in transformers.}
Outside robotics, the Recurrent Memory
Transformer~\citep{rmt} prepends memory tokens to each
segment and carries the hidden states at those positions
forward as the memory input for the next segment.
This yields recurrence over segments without changing the
transformer's basic attention machinery.
\muvla applies the same primitive one level down --- per
environment step rather than per text segment --- to the
multimodal token sequence of a VLA backbone, while
additionally blocking action-to-memory attention and using
the receding-horizon inference protocol described in
Section~\ref{sec:method}.

\paragraph{Benchmarks for memory-dependent manipulation.}
CALVIN~\citep{calvin} established long-horizon
language-conditioned manipulation and remains a standard
downstream benchmark, but it was not designed specifically
to isolate memory-dependent partial observability.
\mikasa~\citep{mikasarobo}, which we use, closes more of
that gap with a 32-task tabletop suite for memory-intensive
manipulation under partial observability, including
occlusion and past-configuration recall.
RMBench~\citep{rmbench} contributes nine simulation tasks
spanning multiple levels of memory complexity together with
the Mem-0 analysis policy, and
RoboMME~\citep{robomme} adds a standardized 16-task
benchmark plus 14 memory-augmented $\pi_{0.5}$ variants for
controlled representation studies.
LongBench~\citep{longbench} moves evaluation to more than
1,000 real-world episodes split into context-independent and
context-dependent regimes.
RoboCerebra~\citep{robocerebra} and
RoboHiMan~\citep{robohiman} emphasize high-level reasoning,
hierarchical composition, and perturbation robustness.
ReMemBench in PRISM~\citep{prism} stresses short-term memory
across eight household manipulation tasks with horizons up
to roughly two minutes, MemMimic from Gated Memory
Policy~\citep{gatedmem} adds additional non-Markovian
manipulation tasks, and PhysMem~\citep{physmem} studies
cross-episode physical memory through verified hypothesis
formation.
Taken together, these benchmarks show that long-horizon
failure is not a single phenomenon: it can come from missing
within-episode memory, poor task decomposition, weak
recovery, or lack of cross-episode adaptation.

\end{document}